\documentclass[10pt,journal,compsoc]{IEEEtran}
\usepackage[pagebackref,breaklinks,colorlinks]{hyperref}

\usepackage{times}
\usepackage{latexsym}
\usepackage{graphicx}
\usepackage{epstopdf}
\usepackage{amsmath}
\usepackage{booktabs}
\usepackage{array}
\usepackage{amsfonts}
\usepackage{bbm}
\usepackage{enumerate}
\usepackage{multirow}
\usepackage{color}
\usepackage{threeparttable}

\usepackage[normalem]{ulem}
\usepackage{mathrsfs}
\usepackage{amsthm}
\usepackage{amssymb, bm}
\usepackage{booktabs}
\usepackage{enumitem}
\usepackage{makecell}
\usepackage{microtype}
\usepackage[dvipsnames]{xcolor}
\usepackage{subfigure}

\newcommand{\newmodule}[0]{edge-reasoning mechanism }
\newcommand{\newmoduleNoSpace}[0]{edge-reasoning mechanism}

\newcommand{\robertalarge}[0]{RoBERTa$_\text{Large}$}

\newcommand{\electralarge}[0]{ELECTRA$_\text{Large}$}

\newcommand{\tqt}[1]{``\textit{#1}''}

\theoremstyle{definition}

\newtheorem{example}{Example}[section]

\usepackage{colortbl}
\definecolor{mygray}{gray}{.9}
\definecolor{mypink}{rgb}{.99,.91,.95}

\newcommand{\hl}{}
\newcommand{\boxyl}[1]{{#1}}
\ifCLASSOPTIONcompsoc
  \usepackage[nocompress]{cite}
\else
  \usepackage{cite}
\fi
\ifCLASSINFOpdf
\else
\fi
\usepackage[linesnumbered,ruled]{algorithm2e}
\usepackage{algorithmic}
\begin{document}
\title{
% Long Document Comprhension with Logical Reasoning
% DAGN: Discourse-Aware Graph Network for Logical Relation Representations
% Discourse-Aware Dynamic Graph Neural Network for Logical Reasoning
% Discourse-Aware Graph Neural Networks via Meta-Path Generation for Logical Reasoning
% \dagn{DAGN: Discourse-Aware Graph Network for Logical Reasoning}
% \angleone{Bipartite Logic Graph Reasoning for Logical Reasoning}
% \angleone{Equipping Pre-trained LMs with Logical Knowledge via Bipartite Logic Graphs}
% \angleone{DAGN+: Fine-grained Discourse-Aware Graph Learning for Textual logic representations}
% \angleone{DAGN+: Bipartite Discourse-Aware Graph Learning for Textual logic representations}
% Discourse-Aware Graph Learning for High- and Low-Resource Textual logic representations 
% Discourse-Aware Graph for Supervised and Zero-shot Textual Logical Reasoning
% Discourse-Aware Graph Neural Networks for Textual Logical Reasoning
% Discourse-Aware Graph Networks for Supervised and Zero-shot Textual Logical Reasoning
Discourse-Aware Graph Networks for Textual Logical Reasoning
}

\author{Yinya Huang, Lemao Liu, Kun Xu, Meng Fang, Liang Lin,~\IEEEmembership{Senior Member,~IEEE,} \\Xiaodan Liang,~\IEEEmembership{Senior Member,~IEEE}
        % Michael~Shell,~\IEEEmembership{Member,~IEEE,}
        % John~Doe,~\IEEEmembership{Fellow,~OSA,}
        % and~Jane~Doe,~\IEEEmembership{Life~Fellow,~IEEE}% <-this % stops a space
\IEEEcompsocitemizethanks{
% \IEEEcompsocthanksitem M. Shell was with the Department
% of Electrical and Computer Engineering, Georgia Institute of Technology, Atlanta,
% GA, 30332.\protect\\
% E-mail: see http://www.michaelshell.org/contact.html
% \IEEEcompsocthanksitem J. Doe and J. Doe are with Anonymous University.
\IEEEcompsocthanksitem X. Liang is the corresponding author. \protect\\ Email address: xdliang328@gmail.com\protect\\
% \IEEEcompsocthanksitem Y. H., Q. C., L. Lin, and X. L. are with Sun Yat-sen University, China.  
\IEEEcompsocthanksitem 
Y. H. and X. L. are with the Shenzhen Campus of Sun Yat-sen University, China.
L. Lin is with Sun Yat-sen University, China.
L. Liu is with Tencent AI Lab.
K. X. is with Huawei, and M. F. is with the University of Liverpool.
%L. Lin is with the School of Computer Science and Engineering, Sun Yat-sen University, Key Laboratory of Machine Intelligence and Advanced Computing, Ministry of Education Engineering Research Center for Advanced Computing Engineering Software of Ministry of Education, China. 
\IEEEcompsocthanksitem This study was done during Y. Huang's internship at Tencent AI Lab.
\IEEEcompsocthanksitem Part of this study has been accepted as ``DAGN: Discourse-Aware Graph Network for Logical Reasoning'' \cite{huang-etal-2021-dagn} in the Proceedings of the 2021 Conference of the North American Chapter of the Association for Computational Linguistics (NAACL 2021). 
This paper extends the previous work in the following aspects. 
First, we add an \newmodule to evolve the constructed logic graphs for adaptive representations,
% in the graph reasoning to evolve the constructed logic graphs, which is learned in the end-to-end training. 
% Second, w
and we conduct experiments in zero-shot scenarios to verify the generalization ability of learned logic representations. 
We further conduct experiments on dialogue-understanding tasks to investigate the adaptation from formal text to informal language.
}% <-this % stops an unwanted space
% \thanks{Manuscript received April 19, 2005; revised August 26, 2015.}
}


\markboth{Journal of \LaTeX\ Class Files,~Vol.X, No.X}%
{Shell \MakeLowercase{\textit{et al.}}: Bare Demo of IEEEtran.cls for Computer Society Journals}



\IEEEtitleabstractindextext{%
\begin{abstract}
Textual logical reasoning, especially question-answering (QA) tasks with logical reasoning, requires awareness of particular logical structures. The passage-level logical relations represent entailment or contradiction between propositional units (e.g., a concluding sentence). However, such structures are unexplored as current QA systems focus on entity-based relations.
In this work, we propose logic structural-constraint modeling to solve the logical reasoning QA and introduce discourse-aware graph networks (DAGNs). 
{The networks first construct logic graphs leveraging in-line discourse connectives and generic logic theories, then learn logic representations by end-to-end evolving the logic relations with an \newmodule and updating the graph features.}
This pipeline is applied to a general encoder, whose fundamental features are joined with the high-level logic features for answer prediction.
Experiments on three textual logical reasoning datasets demonstrate the reasonability of the logical structures built in DAGNs and the effectiveness of the learned logic features. Moreover, zero-shot transfer results show the features' generality to unseen logical texts.
% \lm{The abstract is too long. It would be better to trim it.}

\end{abstract}

\begin{IEEEkeywords}
% Computer Society, IEEE, IEEEtran, journal, \LaTeX, paper, template.
Natural Language Processing, Logical Reasoning, Question Answering, Multi-Turn Dialogue Reasoning, Graph Neural Networks, Supervised Learning, Zero-shot Learning.
\end{IEEEkeywords}}

\maketitle

\IEEEdisplaynontitleabstractindextext


% \fi
\IEEEpeerreviewmaketitle

\section{Introduction} % section 1
\label{sec:intro}

% intro para 1
% The task of textual logical reasoning is to predict the answer that is logically consistent with the given context, sometimes in the guidance of a query. 
Natural language understanding in progress is introducing investigation of machines' reasoning capabilities. 
The recent anticipated logical reasoning requires advanced comprehension of uncovering hidden logical structures.
% Textual logical reasoning requires advanced natural language comprehension of uncovering the logical structures between the lines.
A representative task is logical reasoning QA \cite{yu2020reclor,liu2020logiqa}. It collects questions from standardized exams such as GMAT and LSAT. Each question provides a passage, several answer options, and a question sentence about logical relations, structures, or fallacies. To predict the correct answer, machines need to identify the conclusion and premises in the text and understand how they support or contradict each other. 
% which are solved by understanding the logic in the passages and choosing the corresponding logical component (e.g., conclusion, sufficient assumption) from the options, following the instruction of the question. 
Another representative is multi-turn dialogue reasoning \cite{cui2020mutual}, which requires the machine to predict the next utterance that is logically consistent with the conversation.
% needs a system to comprehend conversations and choose the next utterance avoiding logical mistakes. 
% Solving this problem requires logical reasoning capability.

\begin{figure}[!t]
    \centering
    \includegraphics[width=0.485\textwidth]{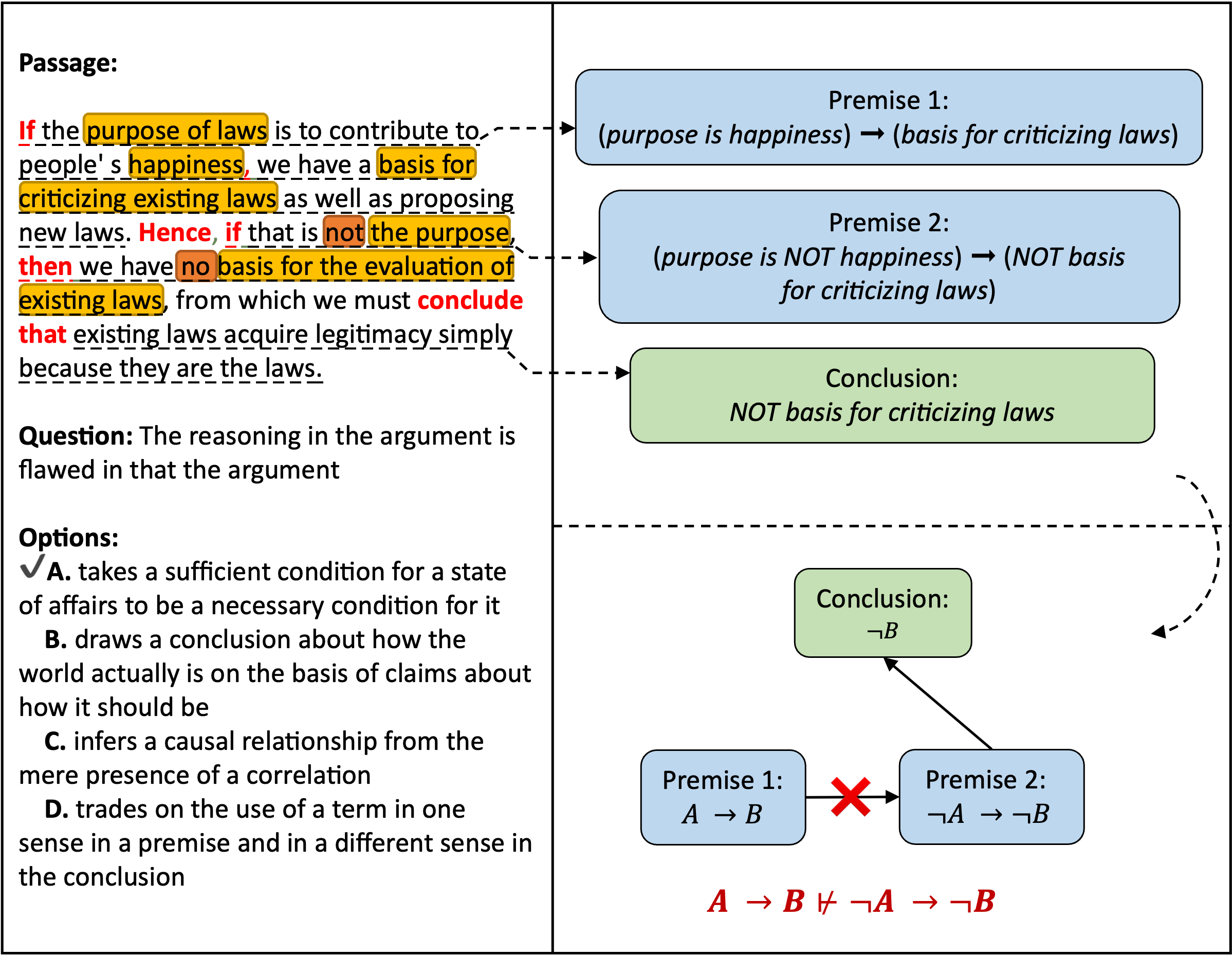}
    \caption{
        An example of logical reasoning QA (left) and the logical structure-based solution (right). Inference patterns are found by linguistic clues. The logical units are the conclusion or premises, which are the sentential text spans. The highlighted key terms indicate the logical variables in logical reasoning.      
    }
    \label{fig:intuition}
\end{figure}

% intro para 2
In principle, logical structures consist of two critical factors, logical components, and logical relations. The core logical components include conclusion and premises, usually complete sentences or subordinate clauses. The logical relations, on the other hand, are mainly entailment, refutation, or contradiction between these sentences. {Moreover, the key phrases in the statements indicate inference patterns.}
Practically, an example is illustrated in Figure~\ref{fig:intuition}. 
{To find the flaw in the argument, one first needs to identify the conclusion and premises. Indicated by the clue words such as ``conclude'', ``if'', and ``then'', the third sentence is the conclusion, whereas the first two sentences provide supporting premises. Indicated by the connectives and the key terms as highlighted, the premises are further decomposed into two entailing structures. From the repeating key terms, one can find the inference patterns $A\to B$ and $\neg A \to \neg B$ in the two premises, respectively. According to the context, Premise 2 is derived from Premise 1, which then derives the conclusion of $\neg B$. However, the reasoning in this argument contradicts the law of contraposition, which is $A\to B \vdash \neg B \to \neg A$. This leads to the correct option A. In contrast, one can hardly answer this question regardless of the logical structure.}

% intro para 3
However, many existing deep models often neglect how to mine such appropriate logical structures, and consequently, is hard to learn logic features to handle complex reasoning.
For example, traditional deep QA systems \cite{chen2016thorough,dhingra2017gated,wang2018co} and retrieval-based dialogue systems \cite{wu-etal-2017-sequential,zhou-etal-2018-multi} learn to match key entities between the passage and the question.
Though mastering previous tasks, they only perform slightly better than random in logical reasoning.
More recent QA systems \cite{de2019question,qiu-etal-2019-dynamically,fang2020hierarchical,zheng2020srlgrn} construct discrete structures according to co-occurrence and coreference of named entities and simulate multi-hop reasoning \cite{welbl2018constructing,yang2018hotpotqa} with graph neural networks \cite{kipf2016semi}. 
Similarly, numerical reasoning systems \cite{ran2019numnet} encode numerical relations between numbers with the topology of graphs.
{Moreover, current Fact-Checking models \cite{zhou-etal-2019-gear,liu2020fine} and NLI models \cite{kim2019semantic,zhang2020semantics,zhang2020sg} focus on semantic matching for better knowledge retrieval or estimating the inference type between sentence pairs. In contrast, solving logical reasoning requires awareness of inference patterns beyond knowledge.}
Therefore, current structures and reasoning processes are insufficient for solving textual logical reasoning, as the core logical structure includes passage-level relations over clause-like units.

% intro para 4
On the other hand, recent advances in transformer-based pre-trained language models (PLMs) \cite{radford2018improving,devlin2019bert,liu2019roberta,yang2019xlnet,lan2019albert} have witnessed great success in extensive natural language tasks \cite{wang2018glue,lai2017race,rajpurkar2016squad}, but fail logical reasoning \cite{yu2020reclor,liu2020logiqa,cui2020mutual}. 
The PLMs are trained on large numbers of unlabeled corpora, and the transformer-based architecture with multiple self-attention layers facilitates the encoding of contextualized representations.
They learn syntactic and semantic structures in an implicit manner \cite{reif2019visualizing}. 
Besides, several works \cite{zhang2020semantics,ye-etal-2020-coreferential,zhang2020sg} incorporate explicit syntactic or semantic structural constraints into PLMs and further improve the representations.
However, the highlighted token correlations do not guarantee appropriate logical components and relations.
Moreover, although the community further observes some reasoning capability \cite{clark2020transformers} from these pure transformer-based models, it is not sufficient for advanced reasoning.

Therefore, several questions are remained open: \textit{
How to construct logical structures to benefit the systems for textual logical reasoning? And how to better learn logic representations?}

% fig 1: intro

To this end, we propose discourse-aware graph networks (DAGNs) {to focus on inference patterns and learn general logic representations. To do this, DAGNs construct logical structure from the plain text as structural constraints, then learns logic representations by end-to-end evolving the logic relations in the graphs and updating the graph features.} 
Generally speaking, the logic graphs are built via linguistic clues and logic theories so that are easily applied to new text. 
{The logic representation learning applies an \newmodule over the constructed graphs, then conducts graph reasoning to update the logic graph features,}
which leverages fundamental embeddings from a general encoder such as PLM.
Specifically, the logic graph construction uses discourse connectives such as \tqt{because} and \tqt{if} \cite{prasad2008penn} as text span delimiters.
They indicate the logical relations and delimit the texts into clause-like logical units, which is in line with the intuition in informal logic theories \cite{toulmin2003uses,Freeman2011ArgumentSR}.
The delimited text spans are regarded as logical reasoning units.
The logic graphs are formed with text spans as nodes, connected by linguistic and logical edges.

Logic representation learning is a graph reasoning process.
% that 
{
It first discovers advanced logical relations from the constructed logic graphs, for instance, 
multi-hop relations with different edge types.
% the multi-hop relation from premise 1 to conclusion in Figure~\ref{fig:intuition}. 
% a joint relation of premise-conclusion and 
% first reasons over the logic graphs to identify advanced logical relations, then} transforms fundamental embeddings into structural features according to the constructed logic graphs. 
The relation discovery is an iterative edge selection and propagation procedure inspired by the previous meta-path generation model \cite{yun2019graph}.
Given the updated logical relations, it then initializes the graph features with token embeddings, then performs graph reasoning to aggregate the node embeddings by a node-weighted graph convolutional network.
} 
% It takes embeddings from a general encoder as input, and the embeddings are further merged by the logical units as initialization of graph nodes. The node features are then aggregated by a node-weighted graph convolutional network via multiple edge types and in multiple steps. 
The output multi-hop logic features are further fused with the fundamental embeddings to provide hierarchical features for downstream prediction. The learning process leverages underlying features such as pre-trained contextual embeddings and merely needs a few rounds of fine-tuning, and is therefore efficient.

We conduct comprehensive experiments on three datasets, including two logical reasoning QA datasets \cite{yu2020reclor,liu2020logiqa} and one multi-turn dialogue understanding dataset \cite{cui2020mutual} in both supervised and zero-shot scenarios. 
% , and in two scenarios, supervised learning and zero-shot learning. 
% \xd{first explain our DAGN+ is superior to existing works[1][2][3]}
In general, DAGNs outperform the state-of-the-art models in supervised settings, showing strong generality in zero-shot transfer. 
% Overall, DAGNs boost the systems in logical reasoning performances. 
{The results show that the \newmodule leads to logical feature generality and model stability.} The logic graphs are proved effective for learning general and transferrable logic representations. This indicates the importance of focusing on inference patterns beyond knowledge in logical reasoning tasks.
% \footnote{
% A preliminary version of this paper is published in NAACL 2021. In this paper, we inherit the idea. This work is extended in four respects. 
% First, we add an \newmodule in the graph reasoning to evolve the constructed logic graphs, which is learned in the end-to-end training. 
% Second, we evaluate the model in zero-shot scenarios to inspect the generality of learned logic representations. We find that the features are more general for logical reasoning when learned with the constraint of the built logic graphs.
% Third, more refined logic graph construction is defined, which contains richer logical knowledge including (1) binary node types, (2) topic-related terms, and (3) variable edges,
% and we demonstrate their effectiveness with extended experiments.
% Fourth, we extend the model evaluation to include a dialogue understanding dataset to see its generalization from written language in standardized exams to ordinary language.
% }

The contributions of this paper are summarized as follows:
\begin{itemize}
    \item {We explore effective discourse-aware graph networks (DAGNs) for textual logical reasoning. The model constructs logic graphs as structural constraints then learns to identify advanced logical relations and learn logic representations by the graphs.}
    
    % We explore effective logic graph construction and representation learning for textual logical reasoning and propose discourse-aware graph networks (DAGNs). The logic graphs are built according to discourse-based indicators and generic logic theories, whereas the logic representation learning is via graph reasoning.

    \item {The \newmodule evolves the logical relations to adapt the logic representation learning, which results in feature generality and model stability.}
    
    \item The proposed logic graph construction uses generic textual clues and logic theories and is easily applied to new texts. Meanwhile, graph-based representation learning leverages fundamental encoding techniques; hence is handy for fine-tuning and is widely applicable.
    
    \item Experiments on three datasets indicate that DAGNs are superior in textual logical reasoning and provide beneficial logical information. Besides, DAGNs show strong generality to unseen logical questions.
    
\end{itemize}

\section{Preliminaries} % section 2
\label{sec:preliminaries}

\subsection{Task: Logical Reasoning QA} % section 2.1
\label{sec:task_definition}
% \xd{first describe the logical reasoning QA task, and give notations}
Logical reasoning QA requires a machine to understand the logic behind the text, for example, identifying the logical components, logical relations, or fallacies.

For multiple-choice logical QA, given a logical passage, a question, and several candidate answer options, a machine needs to predict the answer by understanding the logic of the passage. 
We give notations for convenient discussion. For a logical reasoning question \texttt{(passage, question, options)}, we denote the sequences \texttt{passage}, \texttt{question}, and \texttt{option} as $S_p$, $S_q$, $S_o^c$, respectively, where $c\in C$, $c$ is the candidate index and $C$ is the overall number of candidates. Then a machine's inputs are $S^c = [S_p; S_q; S_o^c]$, $c\in C$, where ``$;$'' denotes sequence concatenation.

Similarly, for multi-turn dialogue reasoning, a machine is given dialogue context and multiple candidate responses and is required to give the logically correct response according to the dialogue context.
For a single dialogue \texttt{(dialogue context, candidate responses)}, 
we denote the sequences \texttt{dialogue context} and \texttt{candidate response} as $S_d$ and $S_r$, respectively. 
The machine's inputs are $S^c = [S_d; S_r^c]$ for each $c\in C$.
Predicting the answer from $C$ options needs to give ranking scores $p^c$ for all $c\in C$.

% The questions and options explore comprehension of the logical passage, including detecting logical components or relationships, understanding the strength of the arguments, evaluating the arguments such as fallacy detection. 
% A similar task is multi-turn dialogue reasoning with logic. Given a dialogue context, a machine needs to pick a response from candidates to make the dialogue logically sound, which also needs logical passage comprehension.
% The overall task setting is text ranking among the options.

% \subsection{Difference among Linguistic Knowledge, Factual Knowledge and Logical Knowledge.}

\subsection{Logic Theories for Logical Reasoning QA} % section 2.2 
% \subsection{Informal Logic and Relevance to Logical Reasoning QA}
% \subsection{Informal Logic} 
% \subsection{Logical Relations in Informal Logic}
% \xd{explain the relations between informal logic and logic reasoning qa task}
% \todo{\textbf{We are the first introducing informal logic to machine logical reasoning.}}
% \todo{(A group of relations in informal logic?)}

Logic theories study symbolic reasoning processes in daily language use. It can be generally grouped into informal logic \cite{toulmin2003uses, Freeman2011ArgumentSR} and formal logic \cite{walicki2016introduction}. 
% The informal logic emphasizes natural language context. 
The informal logic uncovers reasoning structure in context.
In contrast, formal logic extracts the language into symbolic axiomatic systems to evaluate its validity. Both inspire the modeling for logical reasoning QA.
% and constraints logical validity with rules.

\subsubsection{Informal Logic} % section 2.2.1
\textbf{Logical Components in Arguments.} Informal logic \cite{toulmin2003uses,Freeman2011ArgumentSR} studies the structural reasoning processes in argumentation. The structure is named argument \cite{sep-argument}. An example argument is:
\begin{center}
    \textit{A and B; therefore C.}
\end{center}
Here, \tqt{A}, \tqt{B} and \tqt{C} are propositions, and\tqt{C} is a conclusion drawn from the two premises \tqt{A} and \tqt{B}. Hence in this discrete structure, conclusion and premise are two fundamental logical components, which are usually complete sentences or sub-sentences \cite{sep-logic-informal}.

\noindent\textbf{Inference Indicators.}
To uncover the logical components from text and reconstruct the structure, informal logic has organized frequently encountered indicators that prompt the premise or conclusion. Representative premise indicators involve \tqt{since}, \tqt{because}, \tqt{for}, \tqt{given that} and so forth. Meanwhile, conclusion indicators include \tqt{therefore}, \tqt{so}, \tqt{consequently} and others.

% \begin{remark}[Inference Indicators]
%     In informal logic study, textual indicators help identifying the logical components. Representative premise indicators involve \tqt{since}, \tqt{because}, \tqt{for}, \tqt{given that} and so forth. Conclusion indicators include \tqt{therefore}, \tqt{so}, \tqt{consequently} and others.
% \end{remark}

Inspired by these, we reconstruct logical structures for logical reasoning QA by leveraging such inference indicators as text delimiters, which segment the passage into multiple sentences or clauses that properly are the basic reasoning units. The indicators themselves then signify corresponding logical relations between the units. 

% A comprehensive example is as follows:
% % \begin{center}

% \textit{When workers do not find their assignments challenging, they become bored and so achieve less than their abilities would allow. On the other hand, when workers find their assignments too difficult, they give up and so again achieve less than what they are capable of achieving. It is, therefore, clear that no worker' s full potential will ever be realized.}
% % \end{center}

% \begin{example}
% In the sentence \tqt{Digital systems are the best information systems because error cannot occur in the emission of digital signals.}, \tqt{digital systems are the best information systems} is a \underline{conclusion}, and \tqt{error cannot occur in the emission of digital signals} is a corresponding \underline{premise}.
% \end{example}

\subsubsection{Formal Logic} % section 2.2.2
% \xd{connection between this section with previous ones}

% Formal logic systems, for instance first-order logic (FOL), build axiomatic systems to formulate language usage. Given basic constants, variables, and core axioms, with transformation rules, more complicated axioms are generated with logical validity and are included in the system. The rule of substitution is one of the most basic rules.

% \begin{definition}[Derivation of Logical Expressions]
%      xxx
% \end{definition}

\textbf{Deviation of Logical Expressions. }
In formal logic system such as first-order logic (FOL), extensive well-formed formulae (i.e., logical expressions) are derived from a few axioms and rules. The soundness of derivation guarantees that the derived expressions are true if only the axioms are true \cite{walicki2016introduction}. 

For example, in first-order propositional logic, the modus ponens rule is as follows: 
\begin{equation}
    \label{eq:mp}
    P\to Q, P \vdash Q.
\end{equation}
Thus, if $\alpha\land\beta\to\gamma$ is an axiom and is true, and $\alpha\land\beta$ is true, then it is derived that $\gamma$ is true.

Another example is that given that we have the rule of addition: 
\begin{equation}
    P \vdash P\lor Q,
\end{equation}
then say $\alpha\to\beta$ is an axiom and is true, then $(\alpha\to\beta) \lor \gamma$ as a derived expression is true.

Therefore, it is observed that in the logical expression derivation, the expressions that are derived from each other are correlated only if they have shared variables,
% by the recurring variables, 
such as the $\alpha\land\beta$ in the first example and the $\alpha\to\beta$ in the second one. 
This motivates us to build the variable edges in the logic graph construction.

\noindent\textbf{Validity of Expressions and Instantiation. }
If a logical expression is valid, its multiple instantiations are true as they follow the same valid reasoning process. For instance, two instantiations of the modus ponens rule in eq.~(\ref{eq:mp}) are as follows:

% \begin{definition}[Instantiation and Abstraction]
%      xxx
% \end{definition}

% \begin{definition}[Rule of Substitution in FOL] 
%     % In a given term or formula in FOL, the free variable within the term or formula can be substituted by another term or formula.
%     In first-order logic, the substitution is a mapping from variables to terms. It is applied by replacing each occurrence of a variable with a term as a specific instance.
% \end{definition}

% \begin{example}[Substitution in modus ponens]
% In first-order predicate logic, the notation of modus ponens is: 
% % $P\to Q, P \vdash Q$. 
% $\forall x(P(x) \to Q(x)), P(a) \vdash Q(a)$. Therein, the term $a$ substitutes the variable $x$, so that entails the conclusion $Q(a)$.
% \end{example}

% \begin{definition}[Instantiation] 
%      Instantiation of a formal logic rule is replacing the abstract variables and terms with specific instances.
% \end{definition}

\begin{example}[Instantiation of modus ponens]
% And one of its instantiation is the following renowned syllogism: 
\label{example:s}
\tqt{All men are mortal. Socrates is a man. Therefore, Socrates is mortal.} It is obtained by grounding $P$ to \tqt{be\_men}, and $Q$ to \tqt{be\_mortal}.
\end{example}

\begin{example}[Instantiation of modus ponens]
% Similarly, another instantiation is: \tqt{}
\label{example:e}
\tqt{All birds can fly. Eagles are birds. Therefore, eagles can fly.}. It is obtained by grounding $P$ to \tqt{be\_bird}, $Q$ to \tqt{can\_fly}.
\end{example}

We can tell that the statements in Example~\ref{example:s} and Example~\ref{example:e} are true. Albeit they are in diverse topics, as we know that their shared reasoning skeleton, i.e., the modus ponens rule, is valid.  

Furthermore, in logical texts, the logical reasoning processes are performed in a natural language format. The logical variables are embedded. One of the hints for such logical variables is the topic-related terms, which are mainly the recurring topic words or phrases, such as the \tqt{men} and \tqt{mortal} in Example~\ref{example:s} and the \tqt{birds} and \tqt{fly} in Example~\ref{example:e}. 
Accordingly, we provide topic-related terms detection in our graph node construction.

% In textual logical reasoning, the logical variables and terms are often constituents such as noun phrases. The rule of substitution and instantiation provides connections among these key phrases to uncover the logical structure of the text.

% Moreover, the instantiation guarantees the invariance of the truth in the same logical structure, regardless of various domains or topics. 
% For instance, Example~\ref{example:s} and Example~\ref{example:e} share the logical structure. Therefore, though on different topics, it is derived that Example~\ref{example:e} is true once Example~\ref{example:s} is true.

% Therefore, the logic theories provide inspiration and methodology for understanding logical texts for logical reasoning QA.

% fig 2: link construction and learning
\begin{figure*}[!t]
  \centering
  \includegraphics[width=0.9\textwidth]{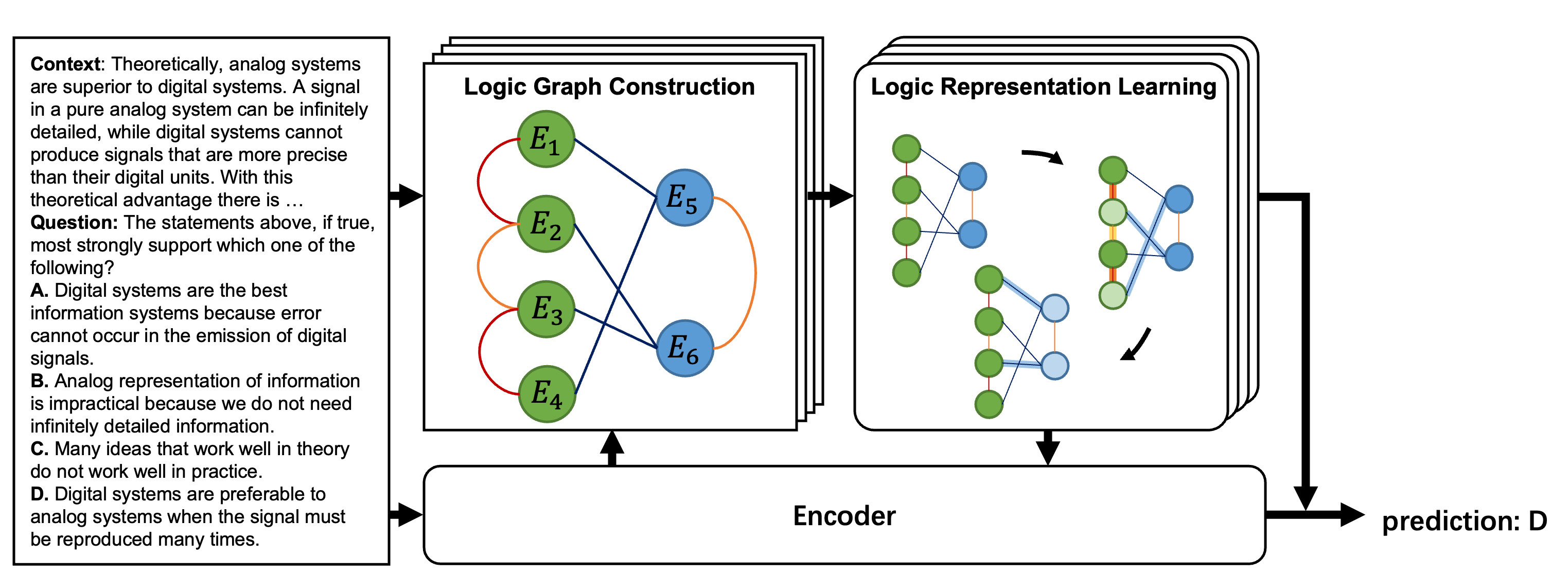}
  \caption{
  The discourse-aware graph networks (DAGNs) pipeline mainly consists of (1) logic graph construction (2) logic representation learning.  
%   \xd{explain in detail}
  The logic graph construction module takes a logical QA data point as input and constructs logic graphs. The logic representation learning module then performs graph reasoning upon the constructed logic graphs. Besides, the encoder provides fundamental embeddings for the pipeline.
  % The overlook of DAGNs. Given a logical reasoning question, it first construct logic graphs, then perform logic representation learning via graph reasoning based on the text features and graph structures.·
  }
  \label{fig:link}
\end{figure*}

\section{Discourse-Aware Graph Networks} % section 3  
% \section{Methodology: DAGNs}

% \subsection{Overview}
% \label{sec:overview}
% \todo{(Question: Do we need a figure in this section?)}\xd{yes, we need a figure to link logic graph construction and representation learning, and get the QA results}

The proposed discourse-aware graph networks (DAGNs) have two main components: logic graph construction and logic representation learning.
The logic graph construction contains strategies of logical unit delimitation, topic-related term detection, graph node arrangement, and graph edge definition.
Meanwhile, logic representation learning is a graph reasoning process that takes contextual encoding as input, updates features with the logic graph constraints, merges multiple features, and is trained end-to-end for logical QA prediction.

\S\ref{sec:graph_construction} introduces the overall strategy of logic graph construction. \S\ref{sec:representation_and_learning} describes the logic representation learning process. The overlook of DAGNs is demonstrated in Figure~\ref{fig:link}.

\subsection{Logic Graph Construction} % section 3.1 
\label{sec:graph_construction}
% To uncover the logical structure behind the texts precisely, we consider a priori logical knowledge in first-order logic (FOL). Specifically, the substitution rule is the core basis of any formal logic systems.
% To maximize the use of text information so that easily adjusting to new unseen logical texts, we mine the logical indicators from the texts. Specifically, we found that discourse connectives are strong logical indicators.
% Therefore in the construction, 

% We design the graph nodes and edges considering strong logical indicators that are found in the texts.
% , so that easily adjust to any new logical texts. 
% Besides, we refer to the concepts of first-order logic (FOL) with our construction, including the character of variable substitution. The construction is demonstrated in Figure~\ref{fig:logic_graph_construction}.

% The construction is of three phases, logical units delimitation, discourse-aware edge linking and substitution edge linking, \todo{which is demonstrated in Figure~\ref{fig:logic_graph_construction}.}

% The graph nodes are built via discourse unit delimitation, and the graph edges are defined with discourse connectives and key-term substitution. The overall construction is demonstrated in Figure~\ref{fig:logic_graph_construction}.

% The logic graph construction is based on discourse information from the text and logic theories. 
Given a logical reasoning question \texttt{(passage, question, options)} or \texttt{(dialogue context, candidate responses)}, which is formalized as $S^c, c\in C$ as described in \S\ref{sec:task_definition}, we construct logic graphs $\mathcal{G}^c = \{\mathcal{V}^c, \mathcal{E}^c\}, c\in C$. 

% The graphs are constructed mainly via two steps: (1) Obtaining graph nodes via delimiting text into elementary discourse units (EDUs). The EDUs are grouped into two disjoint sets, and topic-related terms inside are detected. (2) Forming the graph using discourse-connective edges and variable edges. 
% The following discussion is based on single option $c$, and we omit the superscript.
% The construction is illustrated in Figure~\ref{fig:logic_graph_construction}.

We describe the graph node and edge definition separately. The graph nodes are text's segmented sentences or sub-sentences, indicated by discourse-aware connectives. Each node is further attached with topic-related terms and is assigned a node type. As for the graph edges, discourse-connective edges and variable edges link the nodes differently. The overall construction is illustrated in Figure~\ref{fig:logic_graph_construction}.

\subsubsection{Nodes via Discourse Unit Delimitation} % section 3.1.1
% \subsection{Nodes via Logical Unit Delimitation \todo{(via Discourse Unit Delimitation)}}
\label{sec:node_delimitation}
% \subsubsection{Discourse Units \dagn{Delimitation}}

% The logical reasoning units are usually span of texts describing a conclusion, an assumption, a condition to 
% Logical reasoning usually contains relations such as causation, entailment, contradiction and so forth. Correspondingly, the logical units that are connected by the logical relations are concepts such as conclusion, assumption, evidence, which usually present as spans of texts. 
% Discourse relations such as ``because'', ``if ... then ...'' show strong indication for the logical units. For example, in a sentence with the subordinating conjunction ``because'', the clause following the word ``because'' is usually a reason to the other clause. 

%% graph node construction para 1
It is studied that clause-like text spans delimited by discourse relations can be discourse units that reveal the rhetorical structure of texts \cite{mann1988rhetorical,prasad2008penn}.
We further observe that such discourse units are essential logical propositions in logical reasoning, such as premise or conclusion. As the example shown in Figure~\ref{fig:logic_graph_construction}, the \tqt{while} in the passage indicates a comparison between the attributes of the \tqt{analog system} and that of the \tqt{digital system}. The \tqt{because} in the option uncovers that \tqt{error cannot occur in the emission of digital signals} as a premise to the conclusion \tqt{digital systems are the best information systems}.

%% graph node construction para 2
This observation is agreed with informal logic theories \cite{toulmin2003uses,Freeman2011ArgumentSR}, which study uncovering logical structure from the texts and have conventional in-line logical indicators. For example, acknowledged premise indicators include \tqt{since}, \tqt{because}, \tqt{given that}. Conclusion indicators include \tqt{therefore}, \tqt{so}, \tqt{consequently}, and so forth. Most of these indicators are discourse connectives.

Some discourse parsers \cite{feng2014two,li2018segbot} perform discourse unit segmentation. However, discourse parsing is still challenging, and the parsers are not general to new data, such as logical reasoning questions. For example, SegBot \cite{li2018segbot} is good on the RST-DT dataset but does not work well on the standardized exam texts as in the ReClor dataset.
% \todo{Thus, we turn to develop a general method for parsing discourse for the logical reasoning task.
Thus, we customize discourse unit delimitation strategy for logical texts.

We use the Penn Discourse TreeBank (PDTB 2.0) \cite{prasad2008penn} to help draw discourse connectives. 
PDTB 2.0 contains discourse relations that are manually annotated on the 1 million Wall Street Journal (WSJ) corpus and are broadly characterized into ``Explicit'' and ``Implicit'' connectives. 
The former ones are explicitly present in sentences such as discourse adverbial \tqt{instead} or subordinating conjunction \tqt{because}, whereas the latter ones are inferred by PDTB annotators between successive pairs of text spans split by punctuation marks such as ``.'' or ``;''. 
We take all the ``Explicit'' connectives as well as common punctuation marks to form our discourse-aware delimiter library, presented in Table~\ref{tab:delimeter}. 
Each logical text is split into elementary discourse units (EDUs) by all the delimiters in the library. The EDUs are taken as graph nodes $\mathcal{V}$.

\textbf{Nodes with Topic-Related Terms. }
% \textbf{Domain Phrases Detection.}
% The logical reasoning questions contain massive technical terms which may be unseen by the pre-trained language model before. More over, to answer a logical reasoning question, the key information is the logical structure rather than the explicit semantics of such technical terms. It is related to the idea of variable substitution in first-order logic or numerical reasoning. 
% We detect the ``domain words'' in the text, and link the nodes with the same ``domain words''. Specifically, the ``domain words'' are those longest token spans appearing more than once in the text excluding stop words. 
% We define substitutable ``domain phrase'' as those longest token spans appearing more than once in the text, excluding stop words. We name them as ``domain phrase'' because they are often related to the content of the texts. To detect the them, we use n-gram windows to slide across the text and save the longest repeat n-gram words, after which removing the stop words from them. 
% Considering that the logical texts present great variety of domains such as law, medical science, finance, each with their own domain-specific key terms, we extract all of them out of the texts. 
The desired key terms are those real nouns or phrases that repeatedly appear in the text. Such nouns or phrases are instantiations of logical variables in propositions. As a result, replacing such terms with abstract variables or terms in other topics does not change the process of reasoning. 
% Such nouns or phrases are essentially not related to the logic of the text and can therefore be replaced by other phrases or abstract symbols, according to the rule of substitution. 
For example, in Figure~\ref{fig:logic_graph_construction}, 
the first two sentences indicate a comparison of \tqt{signal} between \tqt{analog system(s)} and \tqt{digital system(s)}. Performing abstraction by replacing \tqt{signal} with variable $\gamma$, \tqt{analog system(s)} with variable $\alpha$, and \tqt{digital system(s)} with variable $\beta$, the propositions are free from the topic of electronics, but the comparison relation is retained.
% the \tqt{signal(s)} and \tqt{digital system(s)} can be replaced by two variables $\alpha$ and $\beta$, whereas the core logical relations, such as comparison denoted by \tqt{while}, causal relation denoted by \tqt{because}, are unaffected.

We use a sliding window to collect the recurring phrases. Given the input logical text, stemming is first applied to handle morphological diversity. Then, the sliding window loops over n-grams and records the reoccurrence. Next, all the stop words and overlapped substrings are filtered. The resulting topic-related terms are attached to the nodes according to which text segment they belong.
% The longest repeated phrases are kept as key terms. 
% According to the discourse unit delimitation, the key terms are labeled for each graph node.

% To extract the key terms, we first normalize the tokens, then use n-gram windows from large to small sizes sliding over the texts and record those longest token spans appearing more than once, excluding stop words. Each node keeps its key terms for edge connection.

\textbf{Binary Node Types.}  
The text of logical reasoning QA consists of two possible structures: \texttt{(passage, question, options)} or \texttt{(dialogue context, candidate responses)}. We regard \texttt{passage} or \texttt{dialogue context} as context texts that carry the main logical reasoning structure, whereas regard \texttt{(question, options)} or \texttt{candidate responses} as candidate texts that are added to the context texts and should remain their logical consistency.

According to the discourse unit delimitation, the graph nodes are naturally from the context texts or the candidate texts.
Therefore, we define two disjoint and independent node sets: context node set $\mathcal{V}_u$ and candidate node set $\mathcal{V}_v$. $\mathcal{V}_u \cup \mathcal{V}_v = \mathcal{V}$ and $\mathcal{V}_u \cap \mathcal{V}_v = \emptyset$.
The interplay between the two node sets formulates logical consistency between the context and the candidate texts.

\subsubsection{Edge Definition} % section 3.1.2
\label{sec:edge}
\textbf{Discourse-Connective Edges.}
%% discourse-connective edges para 1
% The discourse-connective edges definition follows the logic theories \cite{Freeman2011ArgumentSR,toulmin2003uses}.
We directly use the discourse-aware delimiters to build the discourse-connective edges. 
The intuition is that the delimiters indicate the in-line logical relations, as demonstrated in informal logic theories \cite{toulmin2003uses,Freeman2011ArgumentSR}. 
Therefore, the ``Explicit'' connectives and the punctuation marks are taken as two types of edges, and we name them \textit{explicit-connective edges} and \textit{implicit-connective edges}, respectively. 
% Following the unit delimitation, f
One edge is added between the EDUs before and after each delimiter, with the edge type corresponding to the delimiter.
% The discourse-aware edges follow the discourse relation definition in PDTB, which are indicated by the ``Explicit'' and ``Implicit'' discourse connectives. 
% According to PDTB \cite{prasad2008penn}, the edges are classified into explicit-connective edges and implicit-connective edges. 
% Considering the semantic meaning of the connectives, we refer to their original positions in the texts and connect each sequential node pair with the connective between them. 
If ``Explicit'' and ``Implicit'' connectives are present simultaneously, we choose only to use the ``Explicit'' connectives.
Besides, considering the disjoint node sets $\mathcal{V}_u$ and $\mathcal{V}_v$, the discourse-connective edges only connect nodes within the same node-set. 
The edges are undirected.

% table 1
\begin{table*}[!t]
    \small
    \caption{The discourse-aware delimiter library.}
    \label{tab:delimeter}
  \begin{tabular}{
    p{0.22\textwidth}<\centering
    p{0.72\textwidth}%<\centering
    }
    \toprule
    \textbf{Explicit Connectives} & once, although, though, but, because, nevertheless, before, for example, until, if, previously, when, and, so, then, while, as long as, however, also, after, separately, still, so that, or, moreover, in addition, instead, on the other hand, as, for instance, nonetheless, unless, meanwhile, yet, since, rather, in fact, indeed, later, ultimately, as a result, either or, therefore, in turn, thus, in particular, further, afterward, next, similarly, besides, if and when, nor, alternatively, whereas, overall, by comparison, till, in contrast, finally, otherwise, as if, thereby, now that, before and after, additionally, meantime, by contrast, if then, likewise, in the end, regardless, thereafter, earlier, in other words, as soon as, except, in short, neither nor, furthermore, lest, as though, specifically, conversely, consequently, as well, much as, plus, and, hence, by then, accordingly, on the contrary, simultaneously, for, in sum, when and if, insofar as, else, as an alternative, on the one hand on the other hand \\
    \midrule
    \makecell[c]{\textbf{Punctuation Marks} \\\textbf{(Implicit Connectives)}} & . , ; : ? ! \texttt{<s>} \texttt{</s>} \\
    \bottomrule
    \end{tabular}
\end{table*}

% fig 3: logic graph construction
\begin{figure}[t]
	\centering
	\includegraphics[width=0.5\textwidth]{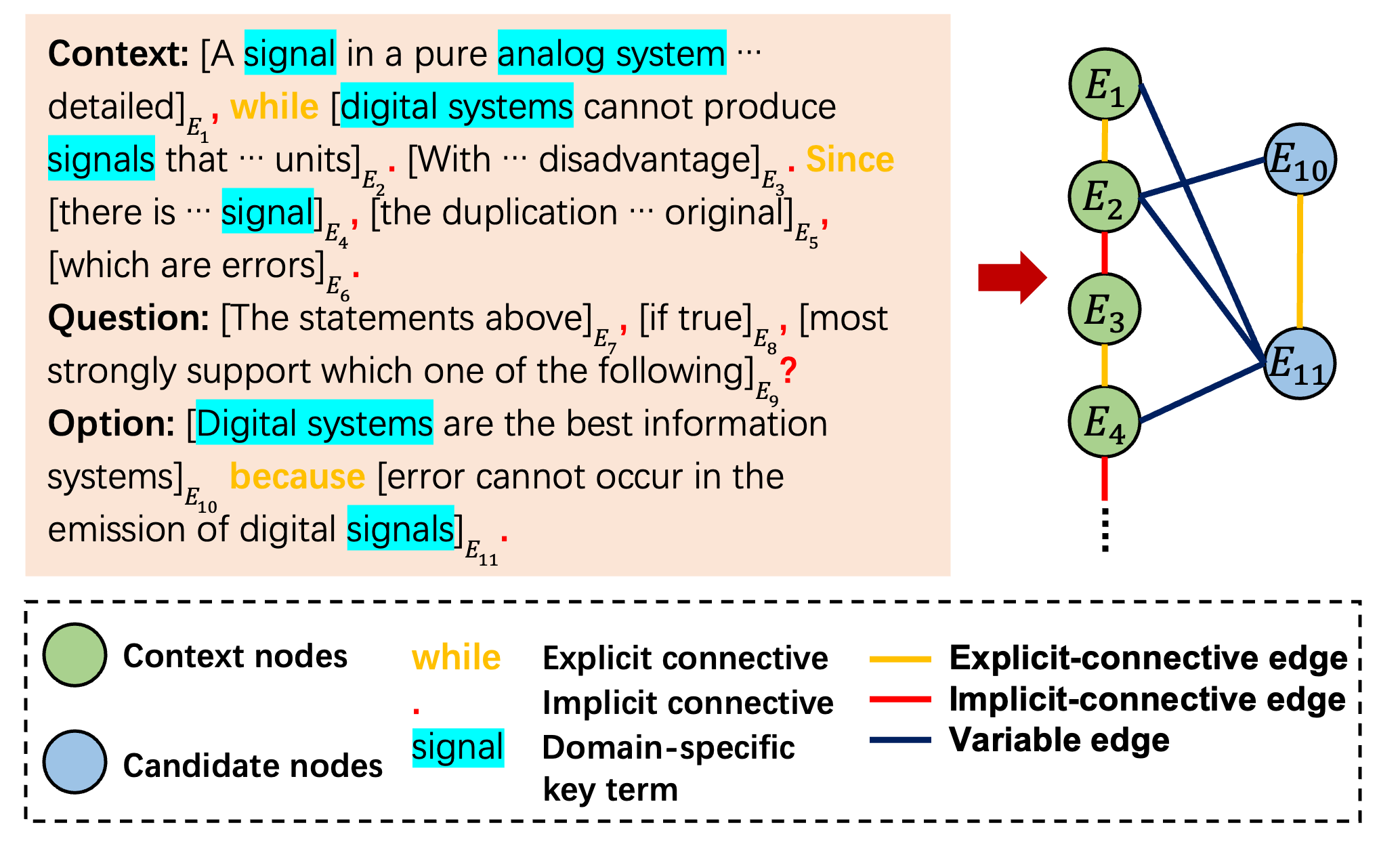}
	\caption{
		The logic graph construction is based on in-line discourse connectives which split the text into segments as logical units and form the graph nodes. 
	}
	\label{fig:logic_graph_construction}
\end{figure}

%% discourse-connective edges para 2
As shown in Figure~\ref{fig:logic_graph_construction}, the two nodes $\text{EDU}_2=$ ``\textit{digital systems cannot produce signals that ... units}'' and $\text{EDU}_3=$ ``\textit{With ... disadvantage}'' are connected with an implicit-connective edge. 
The nodes $\text{EDU}_1=$ ``\textit{A signal in a pure analog system ... detailed}'' and $\text{EDU}_2=$ ``\textit{digital systems cannot produce signals that ... units}'' are joint with the explicit-connective edge when both \tqt{,} and \tqt{while} are between them.
Besides, the nodes $\text{EDU}_6$ and $\text{EDU}_7$ are adjacent in the input text, but there is no discourse-connective edge between them because they are from different node sets.

%% discourse-connective edges para 3
As a comparison, we also try different edge linking strategies for the discourse-connective edges, including random edge linking, full-connection, and single edge type. We further discuss these strategies and their benefits to logical reasoning in Section~\ref{sec:exp_graph_components}.

%% discourse-connective edges para 4
Given the binary node sets $\mathcal{V}_u$ and $\mathcal{V}_v$, we denote the adjacency matrices of explicit-connective and implicit-connective edges as:
\begin{displaymath}
A^E = 
\left(
    \begin{array}{cc}
        A^E_u & 0_{u,v} \\
        0_{v,u} & A^E_v
    \end{array}
\right)
\quad\text{and}\quad
A^I = 
\left(
    \begin{array}{cc}
        A^I_u & 0_{u,v} \\
        0_{v,u} & A^I_v
    \end{array}
\right),
\end{displaymath}
\noindent where $A^E_*$ and $A^I_*$ denote the inner-set edge linkings.
% $u$ and $v$ are the context and candidate node set, respectively, 
 % are the adjacency submatrices for ``Explicit''-connective and ``Implicit''-connective edges.

\textbf{Variable Edges.}
% To follow the basic rule of substitution in logic, we add substitution edges to the logic graphs. The intuition is that, in logic systems such as FOL, the repeated domain-specific key terms can be regarded as variables, and with the relaxation of constraint, free variables, and hence can be substituted by any new term or well-formed formula.
%% sub edges para 1
Variable edges connect the disjoint context nodes $\mathcal{V}_u$ and candidate nodes $\mathcal{V}_v$, representing the derivations between logical propositions.
% thus representing the interaction between the two.
The intuition is that when the candidate nodes from the correct option are joined with the context nodes, the logical consistency is retained, while the intervention of the candidate nodes from the distracting options will disturb the logic graphs.

For simulating such logical consistency as in logical expression derivation, edges are added to those EDU nodes that carry at least one shared variable.
% the EDU nodes, the edges connect the nodes that carry the same variable. 
Practically, the variables are regarded as the tagged topic-related terms.
Thus, given the disjoint node sets, if a node pair shares a topic-related term, an edge is added between them.

%% sub edges para 2
As illustrated in Figure~\ref{fig:logic_graph_construction}, $\text{EDU}_2=$\tqt{digital systems cannot produce signals that ... units} and $\text{EDU}_{10}=$\tqt{digital systems are the best information systems} represent two propositions, and they share the key term \tqt{digital systems}, 
% so these two units have logical derivation relation, 
therefore they are connected with a variable edge. Similarly, $\text{EDU}_1=$\tqt{a signal ... detailed} and $\text{EDU}_{11}=$\tqt{error cannt occur in the emission of digital signals} share the key term \tqt{signal} and are connected with a variable edge. 
The edges are undirected. 

%% sub edges para 3
% For each node with domain-specific key terms, we connect each pair of nodes with the same domain-specific key terms. 
Formally, given the binary node sets $\mathcal{V}_u$ and $\mathcal{V}_v$, for each node pair $(\text{v}_u, \text{v}_v)$, where $\text{v}_u \in \mathcal{V}_u$ and $\text{v}_v \in \mathcal{V}_v$, when there is a key term $\kappa$ that $\kappa \in \text{v}_u$ and $\kappa \in \text{v}_v$, a variable edge is added between them.
% a variable edge is added if $\text{v}_u$ and $\text{v}_v$ share domain-specific key terms.
% As a result, each logic graph is a bipartite graph. 
As a result, the adjacency matrix of the variable edges is:
\begin{displaymath}
A^S = 
\left(
    \begin{array}{cc}
        0_{u} & B^S_{u,v} \\
        B^S_{v,u} & 0_{v}
    \end{array}
\right),
\end{displaymath}
\noindent where $B^S_{u,v}$ and $B^S_{v,u}$ are incidence matrices between $\mathcal{V}_u$ and $\mathcal{V}_v$.

% fig 4: logic representation learning
\begin{figure*}[!t]
	\centering
	\includegraphics[width=\textwidth]{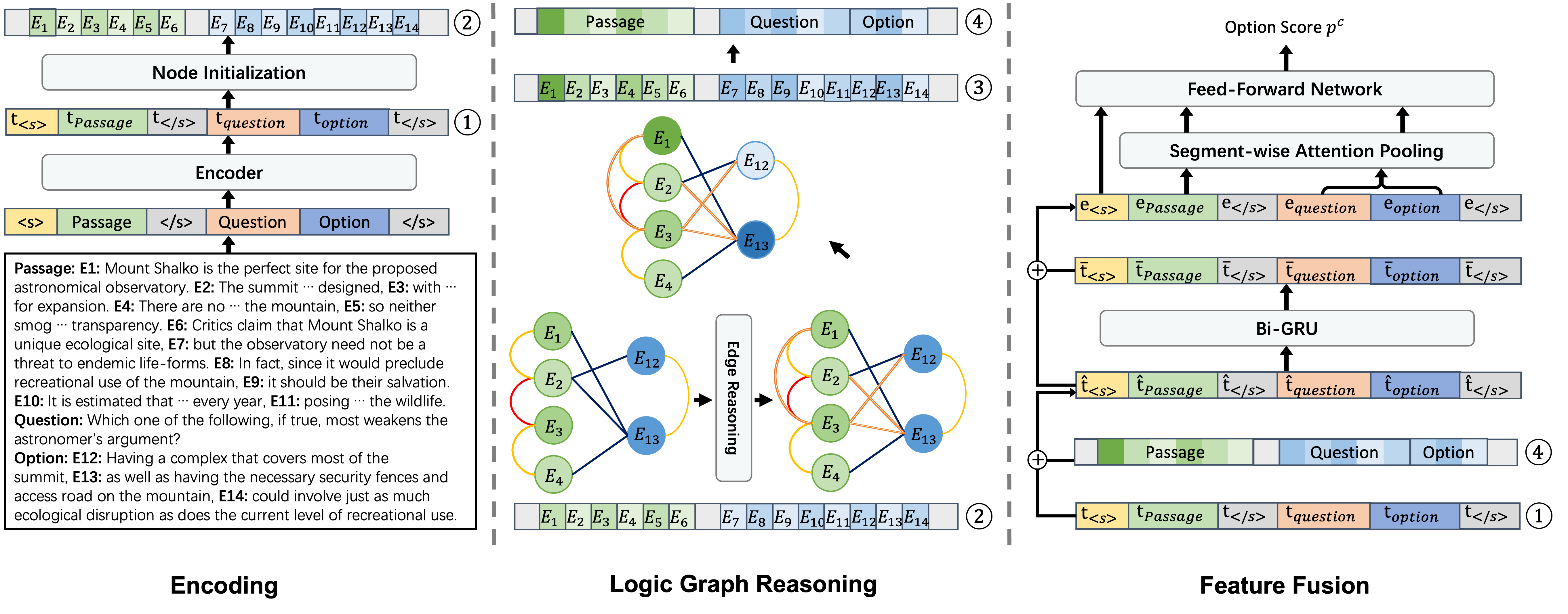}
	\caption{
    The logic representation learning process. Logic graph reasoning starts with node initialization from an encoder and produces logic representations. The initial token embeddings and the high-level logic embeddings are fused for downstream prediction.
	}
	\label{fig:representation_learning}
\end{figure*}

\subsection{Logic Representation Learning} % section 3.2 
% \subsection{Logic Representation Learning and Reasoning}
% \section{Graph Representation \todo{Learning} and Reasoning}
\label{sec:representation_and_learning}
Given a logical question and its constructed graphs, we now build a logic-based model that is end-to-end trained for logic representation learning.
% and reasoning. 
The model takes the question and graphs as input, encodes the input sequence, conducts {edge evolving and} graph reasoning to produce logic representations, then fuses the fundamental encodings for downstream prediction. 
The reasoning module is a plugin module to a general encoder and leverages the contextual features. 
% leverages fundamental encoder and as a 
% encoded token embeddings hence acts as a plug-in method and is encoder-agnostic. 
Hence the overall model only needs a few rounds of fine-tuning for feature updates. Figure~\ref{fig:representation_learning} demonstrates the learning pipeline. 
% the reasoning process.

\subsubsection{The End-to-End Learning Pipeline} % section 3.2.1
\label{sec:pipeline}

\textbf{Text Inputs. }
For a logical question, the input sequences $S^c$, $c\in C$ are formulated as described in Section~\ref{sec:task_definition}.
Each $S^c$ is further truncated into tokens $S^c = (s^c_1, s^c_2, ..., s^c_L)$ where $L$ denotes the number of tokens.

\textbf{Graph Inputs. }
Each $S^c$ has a corresponding logic graph $\mathcal{G}^c$.
% For each $S^c$, the logic graph is $\mathcal{G}^c$.
% there are $C$ logic graphs for each option. 
% For each graph $\mathcal{G}^c$, 
The nodes correspond to elementary discourse units (EDUs) in $S^c$, which are recorded by $\mathrm{D}^c(l) = n$, a position mapping from token position $l$ to segment position $n$. 
$n\leqslant N$, $l\leqslant L$ with $L$ tokens and $N$ EDUs in total.
The edges are of three types, and the model takes their adjacency matrices $\{A^{c,E}, A^{c,I}, A^{c,S}\}$. 

\textbf{Token Encoding.}
The $S^c$, $c\in C$ 
% The input sequences $\{S^c\}_{c\in C}$ 
are individually fed into a shared encoder $\mathrm{E}$ and obtain the token embeddings: $\mathrm{E}(S^c) = (\mathbf{t}^c_1, \mathbf{t}^c_2, ..., \mathbf{t}^c_L)$, where $\mathbf{t}^c_*\in\mathbb{R}^b$ and $b$ is the dimension of a token embedding.

% edge-reasoning in section 3.2.1 
\textbf{{Logic Edge Reasoning. }}
{Given the adjacency matrices $\{A^{c,E}, A^{c,I}, A^{c,S}\}$, a module softly selects the edge types, then perform matrix multiplication to propagate new edges. The soft propagated edges are then converted into adjacency matrices $\{A^{c,(h)}\}_{h\in H}$, and $H$ is the maximum hops of graph reasoning. The set of adjacency matrices are then updated with the propagated edges $\bar{A} = \{A^{c,E}, A^{c,I}, A^{c,S}\} \bigcup \{A^{c,(h)}\}_{h\in H}$. As a result, the evolved graph $\bar{\mathcal{G}}^c$ contains the multi-hop inference edges derived from hybrid logical relations.   
% The hybrid mutli-hop edges join the constructed adjacency matrices $\{A^{c,E}, A^{c,I}, A^{c,S}\}$ for graph reasoning. 
The parameters in the soft edge selection are updated via end-to-end training.
% The soft edge selection is learned from the downstream graph reasoning and answer prediction. 
}

\textbf{Logic Graph Reasoning. }
% \textbf{Logic Representation Learning.}
Given the token embeddings $(\mathbf{t}^c_1, \mathbf{t}^c_2, ..., \mathbf{t}^c_L)$ and the graph inputs $\mathrm{D}^c(l) = n$, $\{A^{c,E}, A^{c,I}, A^{c,S}\}$ {$\bigcup \{A^{c,(h)}\}_{h\in H}$}, logic representations are learned via graph reasoning. Node embeddings are initialized by merging the token embeddings according to $\mathrm{D}^c(\cdot)$, then are updated via multi-step message propagation through the adjacency matrices. Afterword, the updated node embeddings are assigned to each token by $\mathrm{D}^c(\cdot)$ again as the learned logic representation for each token.
% \todo{This reasoning module leverage the constructed logic graph and introduce logical structural bias during training.}

\textbf{Feature Fusion. }
For each token, the learned high-level logic representation and the fundamental contextual embedding are fused. Furthermore, the token embeddings are pooled for downstream prediction. For each option $c$, the model obtains a pooled embedding $\hat{\textbf{p}}^c$.
% The logical embeddings are incorporated into the original token embeddings.  

\textbf{Option Ranking. }
Each option embedding $\hat{\textbf{p}}^c$ is fed into a linear layer to get a ranking score. Furthermore, the probabilities for selecting the options are obtained by a softmax function:
\begin{gather}
    \hat{p}^c = \mathbf{W}\hat{\mathbf{p}}^c + \mathbf{b}, \\
    p^c = \frac{e^{\hat{p}^c}}{\sum\limits_{c\in C}e^{\hat{p}^c}}.
\end{gather}
% \textbf{Prediction.}
% \subsection{Downstream Prediction} 
% \lm{It is not clear that your graph representation is general to be applied to any backbone model. You should disentangle your model in this subsection such that your model is a kind of graph representation plus a backbone model.}
% After graph reasoning, we obtain the output feature $\hat{\mathbf{p}}$ for each input, just as in any backbone models. 
% \subsubsection{Multiple Choice QA} 
% In a multiple-choice question, for each candidate option, we build independent logic graph. The model then learns logic representations for each candidate. 
% For a question with $C$ candidate options, the model outputs $C$ corresponding scores $\{p_c\}$, where $c\in C$. 
% \begin{equation}
%     p = \mathbf{W}_2\hat{\mathbf{p}} + b_2.
% \end{equation}

\textbf{Overall Objective Function. }
Given single question input \texttt{(passage, question, options)} or \texttt{(dialogue context, candidate responses)},
the model is end-to-end trained by cross-entropy loss with option labels $y^c$:
\begin{equation}
    \mathcal{L} = -\sum\limits_{c\in C} y^c \text{log}(p^c).
\end{equation}

\subsubsection{Logic Edge Reasoning}
% \subsubsection{Learnable Hybrid Edge Inference}  % section 3.2.2
% \subsubsection{Meta-Path Guided Logical Relation Reasoning}
% {In this section, we introduce the meta-path generation given a semi-constructed} logical graph $\mathcal{G}$. 
{
% This module produces hybrid edges for the multi-hop logical relation. 
% The logic edge reasoning mechanism elicits hybrid edges by softly selecting the multiple edges and performing meta-path generation, motivated by \cite{yun2019graph}. 
The \newmodule is demonstrated in Algorithm~\ref{alg:edge_reasoning}. 
Given a logic graph with three edge types, we concatenate their corresponding adjacency matrices with an identity matrix 
% Given the multiple types of edges $\{A^{c, E}, A^{c, I}, A^{c, S}\}$ in the graph $\mathcal{G}^{c}$, it first initializes the soft edge selection by concatenating the adjacency matrices along with an identity matrix, 
$\bar{A}^{(0)} = [A^{E}; A^{I}; A^{S}; I]$. 
% the \newmodule generates hybrid edges $\bar A^{c}$ and updates the adjacency matrix set for the graph $\{A^{c, E}, A^{c, I}, A^{c, S}\} \cup \bar A^P{c}$. 
% {It contains three small operations to achieve the goal: edge-selection, edge-increment, and edge-elicitation. } 
% For initialization, the three adjacency matrices along with an identity matrix $I^{c}$ are concatenated into $\hat{A}^{c, (0)}$. 
The soft edge selection weighted sum the adjacency matrices $\bar A^{(0)}$, 
% it softly selects the edges from $\bar{A}^{c, (0)}$ 
and outputs the soft selected edges $\Gamma^{(0)}$:
% For each node pair in the graph, the edge selection weights are computed to select over current edge types:
}
\begin{equation} % eq 6
	% \alpha_i=\sigma(\mathbf{W}^{\alpha}(\mathbf{v}^{(k-1)}_i)+\mathbf{b}^{\alpha}),
    \Gamma^{(0)} = \bar{A}^{(0)} \cdot \text{softmax}(\mathcal{W}^{(0)}).
\end{equation}
{where $\mathcal{W}^{(0)}\in\mathbb{R}^{N\times N}$ is a weight matrix initialized with normal distribution.

% During the edge reasoning, 
Then the edge reasoning is performed in an iterative manner and updates the final edge set $\bar A$. 
During the process, another soft edge selection is performed and yields $\hat\Gamma$. 
Then given the $\hat\Gamma$ and the soft edge matrix from the last iteration $\Gamma^{(i-1)}$, an edge propagation is performed by matrix multiplication between them, and produces the $i$-hop soft edge matrix:}
\begin{equation} % eq 7
    \Gamma^{(i)} = \Gamma^{(i-1)}\hat\Gamma.
\end{equation}

{The resulting $\Gamma^{(i)}$ is converted into a new adjacency matrix $\bar A^{(i)}$ if the soft edge element exceeds a threshold $\delta$, which is added to the final edge set $\bar A$. 
%
%
%
% Given the soft propagated matrix $\hat A^{(i-1)}$ from the last iteration, which indicates $i$-th hop edges, the soft edge matrix in the current step is directly assigned with: $\Gamma^{(i)} = \hat A^{(i-1)}$. 
%
% Meanwhile, a new soft edge matrix $\Gamma^{c, (0)}$ is computed from $\bar{A}^{c, (0)}$ as in Equation (6), which indicates 1-hop relations. 
%
% Then edge propagation is performed by matrix multiplication between the two soft edge matrices, $\hat{A}^{(i)} = \Gamma^{(i)}\Gamma^{(0)}$.
%
% The resulting $\hat{A}^{(i)}$ is converted into an adjacency matrix with a threshold and outputs to the final edge set $\bar A$. 
}

% {
% in the current $i$-th step are computed by the weight matrix to obtain softly selected edges $\Gamma^{, (i-1)}$, then is multiplicated with the initial selected edges $\Gamma^{, (0)}$ and obtain hybrid edges in current step $i$. A threshold $\delta$ is used to turn the soft matrix back into one-hot adjacency matrix $\bar A^{c, (i)}$, which are then added to the output adjacency matrices set $\bar A$. }

{
% After the iterations, t
To further increase the diversity of hybrid edges, the edge reasoning process is repeated for $d $ times and also updates $\bar A$. 
% the edge reasoning is conducted multiple times on the original graph $\mathcal{G}$, duplicated graph $\mathcal{G}$, and multiple masked graphs. All the produced hybrid edges are included as $\mathcal{E}^H$.
%
The logic graph is then updated with the hybrid edges $\mathcal{G} = (\mathcal{V}, \mathcal{E}\cup\mathcal{E}^H)$, where $\mathcal{E}^H$ is the edge set corresponds to $\bar A$.
% $\bar{A}$ is the set of adjacency matrices of $\mathcal{E}^H$. 
}

\subsubsection{Logic Graph Reasoning} % section 3.2.3
% \subsubsection{Logic Representation Learning}
\label{sec:learning}
% \vspace{-2mm}
This section illustrates the detailed logic representation learning process. This process is conducted via graph reasoning by a graph neural network. It consists of node initialization, graph reasoning, and logic embedding assignment for each token.

\textbf{Node Initialization. }
The graph nodes are EDUs, therefore they are initialized 
% The graph nodes are initialized 
with token embeddings to leverage contextual information.
Given token embeddings $(\mathbf{t}_1, \mathbf{t}_2, ..., \mathbf{t}_L)$ and the logical unit delimitations $\mathrm{D}(l) = n$, 
% where $\mathrm{D}$ is a mapping from token position $l$ to EDU position $n$, 
% $l\leqslant L$, $n\leqslant N$, $N$ is the number of EDUs in this sequence,
the node embedding corresponding to the $n$-th EDU (denoted as $U_n$) is then calculated via 
% Second, merging the token embeddings within spans via 
$\mathbf{v}^{(0)}_n = \mathrm{M}(\bigwedge_{\mathrm{D}(l)\in{U_n}} \mathbf{t}_l)$, where $\mathrm{D}(l)\in{U_n}$ denotes the tokens in the $n$-th EDU and $\mathrm{M}$ is the merging function.
% $U_n$ is the set of token indices belonging to $n$-th EDU and 
% $\bigwedge_{l\in{S_n}} \mathbf{t}_l$ denotes the token embeddings within the span $U_n$. 
Specifically, we use a trivial $\mathrm{M}$, which is sum pooling the token embeddings: $\mathbf{v}^{(0)}_n=\sum\limits_{\mathrm{D}(l)\in{U_n}} {\mathbf{t}_l}$.

% alg er
\begin{algorithm}[t!]
	% \SetAlgoLined
	\SetAlgoNoLine
	\caption{Logic Edge Reasoning}
	\label{alg:edge_reasoning}
	{\small{
			\KwIn{A logic graph $\mathcal{G}$ with explicit-connective edges $A^E$, % \in\mathbb{R}^{N\times N}
			implicit-connective edges $A^I$, % \in\mathbb{R}^{N\times N}
			and variable edges $A^S$, % \in\mathbb{R}^{N\times N}
			identity matrix $I$, % \in\mathbb{R}^{N\times N}
			edge-extraction threshold $\delta$,
			the max hop $H$
			}
			% \KwOut{The set of multi-hop logical edges $\bar A$ }	
            \KwOut{The set of hybrid logical edges $\bar A$ }	
			\BlankLine
			// Initialization
			\BlankLine
			$\bar A^{(0)}$ $\leftarrow$ $[A^E; A^I; A^S; I]$, $\enspace$
			$\mathcal{W}^{(0)}$ $\leftarrow$ $\mathcal{N}(0, 1)$ \\
			% $\Gamma^{(0)} \leftarrow \sum_{} (\hat A^{(0)} \cdot \text{softmax}(\mathcal{W}))$ \\
			$\Gamma^{(0)} \leftarrow $ \text{edgeSelection}($\bar A^{(0)}$, $\mathcal{W}^{(0)}$) \\	
                % $\hat{A}^{(0)} \leftarrow $ \text{edgeSelection}($\bar A^{(0)}$, $\mathcal{W}$) \\	
			$\bar A \leftarrow \emptyset$ \\			
			\BlankLine
			// Edge Reasoning 
			\BlankLine
			\For{$i=1:H$}
			{                
                    $\mathcal{W}^{(i)}$ $\leftarrow$ $\mathcal{N}(0, 1)$ \\
                    
				$\hat{\Gamma}$ $\leftarrow$ $\text{edgeSelection}(\bar A^{(0)}, \mathcal{W}^{(i)})$ \\
				
                    $\Gamma^{(i)}$ $\leftarrow$ $\text{edgePropagation}(\Gamma^{(i-1)}, \hat\Gamma)$ \\ 
		
				$\bar A^{(i)}$ $\leftarrow$ $\text{edgeExtraction}(\Gamma^{(i)}, \delta)$

				$\bar A$ $\leftarrow$ $\bar A \cup \{\bar A^{(i)}\}$ \\
			}			
			\Return $\bar A$
	}}	
\end{algorithm}
\setlength{\textfloatsep}{1pt}

\textbf{Graph Reasoning. }  % graph reasoning in section 3.2.3
% \subsubsection{Graph Reasoning} 
% % \lm{It is not reasoning but learning representation. Reasoning means that backbone model equipped with graph representation for the downstream tasks.}
% % \todo{(Graph reasoning usually means applying GNNs without evolving graph structures, which is the case in this paper. In contrast, graph learning usually learns the graph structures.)}
% Inspired by previous graph-based models that solve reasoning problems \cite{ran2019numnet,chen-etal-2020-question}, we also learn node representations to obtain higher-level features. However, we consider different graph construction and encoding.
Given a logic graph $\mathcal{G} = (\mathcal{V}, \mathcal{E})$, 
% where $\mathcal{E} = \mathcal{E}^E \cup \mathcal{E}^I \cup \mathcal{E}^S$,
{where $\mathcal{E} = \mathcal{E}^E \cup \mathcal{E}^I \cup \mathcal{E}^S \cup \mathcal{E}^H $},
for a node $v_i \in \mathcal{V}$, $\mathcal{N}_i = \{j|(v_j, v_i)\in\mathcal{E}\}$ indicates its neighbors.
{The node embeddings are updated via: the explicit-connective edges $\mathcal{E}^E$, the implicit-connective edges $\mathcal{E}^I$, the variable edges $\mathcal{E}^S$, and the hybrid edges $\mathcal{E}^H$. The corresponding adjacency matrices are $A^E$, $A^I$, $A^S$ and $\bar A$. }
% The node embeddings are updated via three types of edges: the explicit-connective edges $\mathcal{E}^E$, the implicit-connective edges $\mathcal{E}^I$, and the variable edges $\mathcal{E}^S$, whose adjacency matrices are correspondingly $A^E$, $A^I$ and $A^S$.
% the graph reasoning considers three types of adjacency: explicit discourse edges, implicit discourse edges, substitution edges, denoted as $r^E$, $r^I$, $r^S$ respectively. 
% $\mathbf{W}^{r}$ is an adjacency matrix of single edge type. 

For stability, we first normalize the variable matrix 
$A^S = 
\begin{pmatrix}
        0_{u} & B^S_{u,v} \\
        B^S_{v,u} & 0_{v}
\end{pmatrix}$
with:
\begin{gather} % eq 4
    \hat{B}^S_{u,v} = D^{-1}_{u,v}B^S_{u,v}, \quad
    \hat{B}^S_{v,u} = D^{-1}_{v,u}B^S_{v,u}
    % \vspace{-0.1cm}
\end{gather}
% $\hat{B}^S_{u,v} = D^{-1}_{u,v}B^S_{u,v}$, $\hat{B}^S_{v,u} = D^{-1}_{v,u}B^S_{v,u}$, 
\noindent where $D^{-1}_{u,v}$ is the degree matrix of $B^S_{u,v}$ and similar to $D^{-1}_{v,u}$. 
% \todo{(Question: where to put this para?)}

Then node features are updated via multiple graph learning layers to obtain multi-hop logic representations.
% Then multiple steps of feature update are conducted to obtain multi-hop logic representations.
% The learning process is recursively conducted to obtain multi-hop logic representations.
% In the learning process, 
For node $v_i$, its initial node embedding is $\mathbf{v}^{(0)}_i$.
Given node embedding $\mathbf{v}^{(k-1)}_i$ from the $(k-1)$-th layer,
% the model first calculates weight for each node 
a node weight is first calculated via linear transformation with a sigmoid function $\sigma$:
\vspace{-0.2cm}
\begin{equation} % eq 5        
	\alpha_i=\sigma(\mathbf{W}^{\alpha}(\mathbf{v}^{(k-1)}_i)+\mathbf{b}^{\alpha}),
\end{equation}
\vspace{-0.2cm}
% \noindent then the representation for logical unit $v_i$ is updated via 
\noindent then message propagation is conducted by simultaneously considering three relation types and taking information from the neighbors $\mathbf{v}_j \in \mathcal{N}_i$:
\begin{gather} % eq 6, 7
	\tilde{\mathbf{v}}^{(k-1)}_j = \mathbf{W}^\gamma\mathbf{v}^{(k-1)}_j + \mathbf{b}^\gamma, \\	
	\tilde{\mathbf{v}}^{(k-1)}_i = \frac{1}{|\mathcal{N}_i|}(\sum_{j\in\mathcal{N}_i}
		\sum_{\mathsf{E}\in\{E,I,S\}}
		\alpha_j{A}^{\mathsf{E}}_{ji}\tilde{\mathbf{v}}^{(k-1)}_j).
\end{gather}

% The logic representation update is finished by merging the node embedding itself:
The node embedding for the $k$-th layer is finished by joining the embeddings:
\begin{equation}
	\mathbf{v}^{(k)}_i = \text{ReLU}(\mathbf{W}^\eta\mathbf{v}^{(k-1)}_i + \tilde{\mathbf{v}}^{(k-1)}_i + \mathbf{b}^\eta),
\end{equation}
\noindent where $\mathbf{W}^\eta$ and $\mathbf{b}^\eta$ are weight and bias respectively.

\textbf{{Global Graph Representation. }} 
% {
% Self-attention is performed among the updated node embeddings $\{\mathbf{v}_i\}_{i\in N}$ to obtain a global graph representation $\mathbf{v}_G$. And we simply set $\alpha^{G}_i$ to 1 in this case. 
% % The global node representation is then fused with the global contextual representation $\mathbf{t}^\lambda_0$ to illustrate the logical consistency between context and candidates. 
% }
{The updated node embeddings $\{\mathbf{v}_i\}_{i\in N}$ are fed into a dot-product self-attention layer \cite{vaswani2017attention} and obtains $\{\mathbf{v}^{G}_i\}_{i \in N}$, which are then weighted summed into the global graph representation. The weights $\alpha^{G}_i$ are simply set to 1 in this case. 
} 
\begin{equation}
    \mathbf{v}_G = \sum_{i\in N} \alpha^{G}_{i}\mathbf{v}^{G}_i.
\end{equation}

% \begin{gather}
%     % \alpha^{G}_{ij} = \text{softmax}(\frac{1}{\sqrt{}}\cdot\mathbf{W}^G\mathbf{v}_i\cdot\mathbf{W}^G\mathbf{v}_j), \\
%     % \mathbf{v}^G_i = \sum_{j\in\mathcal{N}_i}\alpha^{G}_{ij}\mathbf{W}^G\mathbf{v}_j, \\

%     % \mathbf{V}^G = \text{softmax}(\frac{\mathbf{V}^{G}\mathbf{V}^{G}^{\top}}{\sqrt{d^V}})\mathbf{V}^{G}, \\
%     \mathbf{v}_G = \sum_{i\in N} \alpha^{G}_{i}\mathbf{v}^{G}_i, 
% \end{gather}

\textbf{Token-wise Logical Embeddings. } % section 3.2.3
% \textbf{Logical Embedding Assignment.}
% \textbf{Logical embedding for each token. } 
The updated node embeddings are assigned to each token. For each $l\in L$, based on $D(l) = n$, we have:
% Since the logical units are gathers of the tokens in sequence, we distribute the logic representations back to the tokens, so that each token has their corresponding logical embeddings:
\begin{equation}
    \mathbf{t}^\lambda_l = \mathbf{v}_n.
\end{equation}

\subsubsection{Feature Fusion} % section 3.2.4
\label{sec:fusion}
% \subsection{Hierarchical Feature Fusion}
% \textbf{\todo{(Question: No idea where to put this section.)}}\xd{here is good}
% \subsubsection{Hierarchical Feature Fusion}
% \subsection{Hierarchical Downstream Prediction}
% \todo{Downstream prediction is demonstrated in Figure~\ref{fig:downstream_prediction}. Question: need this figure?}

After logic representation learning, each token, now has an original token embedding, and a logical embedding.
{The start-token embedding pairs to the global graph representation $\mathbf{v}_G$, representing the correspondence between the text and the structure. 
% representing the logical consistency between the passage and the graph. 
}
% {The global graph representation $\mathbf{v}_G$ and the global contextual representation  }
The embeddings are fused with a hierarchical fusion, followed by pooling.

\textbf{Hierarchical Fusion. }
% Since the logical units are gathers of the tokens in sequence, we distribute the logic representations back to the tokens, so that each token has their corresponding logical embeddings. (moved)
% To use the logic embeddings and obtain rich information for downstream tasks, the model takes hierarchical features for rich information by incorporating both high-level logic representations and low-level token embeddings. 
For each token $s^c_l \in S^c$, $l\leqslant L$, the fundamental token embeddings $\mathbf{t}_l$ and the high-level logic embeddings $\mathbf{t}^\lambda_l$ are added up, followed by a layer normalization \cite{ba2016layer}:
% Then the enhanced token embeddings are summation of the logical embedding and the initial token embedding with layer normalization \cite{ba2016layer}.

\begin{gather}
% \mathbf{t}^\lambda_l = \mathbf{v}_n, \\
\hat{\mathbf{t}}_l = \text{LayerNorm}(\mathbf{t}^\lambda_l + \mathbf{t}_l).
\end{gather}
% \noindent where $l\in S_n$.

% To further obtain logic-aware embeddings over the context, t
The resulting token embedding sequence 
% $(\hat{\mathbf{t}}_1, \hat{\mathbf{t}}_2, ..., \hat{\mathbf{t}}_L)$ 
$(\hat{\mathbf{t}}_0, \hat{\mathbf{t}}_1, \hat{\mathbf{t}}_2, ..., \hat{\mathbf{t}}_L)$ 
are further fed into a bidirectional GRU \cite{cho2014properties} with residual structure \cite{he2016deep} and layer normalization:  

\begin{gather}
	% \hat{\mathbf{T}} = \text{Bi-GRU}(\hat{\mathbf{T}})
	% \mathbf{e}_l = GRU(\hat{\mathbf{t}}_l, \overrightarrow{\mathbf{h}}) || GRU(\hat{\mathbf{t}}_l, \overleftarrow{\mathbf{h}}).
	\bar{\mathbf{t}}_l = \text{Bi-GRU}(\hat{\mathbf{t}}_l), \\
	\mathbf{e}_l = \text{LayerNorm}(\hat{\mathbf{t}}_l + \bar{\mathbf{t}}_l).
\end{gather}

\textbf{Segment-wise Pooling. }
% \textbf{Sequential fusion}.
% The obtained contextual enhanced token embeddings 
The hierarchically fused embeddings $(\mathbf{e}_1, \mathbf{e}_2, ..., \mathbf{e}_L)$ are separated into three segments: 
% the first-token embedding, the passage-question embeddings and the option embeddings, 
the first-token segment $\mathbf{e}_1$, the passage segment $\{\mathbf{e}^p_*\} = (\mathbf{e}_2, ..., \mathbf{e}_M)$, and question-option segment $\{\mathbf{e}^o_*\} = (\mathbf{e}_{M+1}, ..., \mathbf{e}_L)$, $1 < M < L$.
% which respectively represent the global information, the shared information, and the unshared information of the sequence,
% respectively denoted as $\mathbf{e}_0$, 
% % $\mathbf{E}_p = \{\mathbf{e}_p\}$, 
% $\{\mathbf{e}^{p}_*\}$,
% % $\mathbf{E}_p = \{\mathbf{e}_o\}$. 
% $\{\mathbf{e}^{o}_*\}$.
The passage embeddings and the question-option embeddings are further merged into two single embeddings $\mathbf{e}^{p}$ and $\mathbf{e}^{o}$ via segment-wise attention pooling, respectively:

\begin{gather}
	% \mathbf{e}^{(single)}_p = \text{SoftMax}(\mathbf{W}\mathbf{E}_p)
	% \mathbf{e}^{(single)}_p = \frac{e^{\mathbf{e}_p}}{}
	\alpha_{p} = \frac{e^{\mathbf{e}^{p}_m}}{\sum_{m\in [2, M]} e^{\{\mathbf{e}^{p}_m\}}}, 
	    \quad \mathbf{e}^{p} = \sum_{m\in [2, M]} \alpha_{p}\mathbf{e}^{p}_m, \\
	\alpha_{o} = \frac{e^{\mathbf{e}^{o}_m}}{\sum_{m\in [M+1, L]} e^{\{\mathbf{e}^{o}_m\}}}, 
	    \quad \mathbf{e}^{o} = \sum_{m\in [M+1, L]} \alpha_{o}\mathbf{e}^{o}_m. 
\end{gather}

At last, the three segment-wise embeddings are integrated via concatenation and a single-layer perceptron with normalization:
% The final score for a single token sequence and the corresponding graph is obtained by concatenation of the three single embeddings and a two-layer perceptron:

\begin{gather}
	% \mathbf{e} = \mathbf{e}_0 || \mathbf{e}^{p}_{merge} || \mathbf{e}^{o}_{merge}, \\
	\mathbf{e} = [\mathbf{e}_1; \mathbf{e}^{p}; \mathbf{e}^{o}], \\
	\hat{\mathbf{p}} = \text{LayerNorm}(\text{GeLU}(\mathbf{W}^\sigma\mathbf{e} + \mathbf{b}^\sigma)).
% 	p = \mathbf{W}_2\hat{\mathbf{p}} + b_2.
\end{gather}

\section{Experiment} % section 4
\label{sec:exp}
To validate the logic graph construction and the representation learning, we conduct experiments on three textual logical reasoning datasets, including logical reasoning QA and multi-turn dialogue reasoning. We analyze and discuss graph construction and model components in representation learning. 
Besides, we conduct a generalization test among the datasets via zero-shot learning.
% To explore the generality, we present the zero-shot learning results among the datasets.
% We evaluate the effectiveness of the graph construction. Besides, we explore the generality of the a priori graph structures with logical information. 

\subsection{Datasets} % section 4.1
ReClor \cite{yu2020reclor} is a multiple-choice QA dataset with 6,138 logical reasoning questions modified from standardized tests such as GMAT and LSAT. The questions are split into train/dev/test sets with 4,638/500/1,000 questions respectively. 
ReClor contains 17 question types, including questions about logical components (such as ``Necessary Assumptions'', ``Sufficient Assumptions''), logical relations (such as ``Strenghthen'', ``Weaken''), reasoning evaluation (such as ``Evaluation'', ``Technique'') and so forth.
The passages contain a mass of complex sentences with uncommon words.
The training set and the development set are available. The test set is hold-out and split into an EASY subset and a HARD subset according to the performance of the BERT-base model \cite{devlin2019bert}. The test results are obtained by submitting the test predictions to the leaderboard. The evaluation metric is accuracy. 

LogiQA \cite{liu2020logiqa} is also a multiple-choice QA dataset with logical reasoning questions. It consists of 8,678 questions collected from the National Civil Servants Examinations of China and manually translated into English by professionals. 
LogiQA contains 5 question types. It shares some of the reasoning types with ReClor, for example, ``Sufficient Conditional Reasoning''. The texts are less lexically complex than that in ReClor.
The dataset is randomly split into train/dev/test sets with 7,376 / 651 / 651 samples respectively. 
% Both datasets contain multiple logical reasoning types. The evaluation metric is accuracy over the split sets. 

% table 2: reclor main result.
\begin{table}[!t]
    % \footnotesize
    \caption{
    Experimental results on ReClor dataset.
    Accuracies (\%) are reported.
    Test-E and Test-H represent the EASY and HARD set of ReClor testing, respectively.
    % \todo{DAGN+ denotes the fine-grained version of DAGN.}
    % * means that the results are obtained with augmented data. 
    % * With data augmentation. 
    }
    \label{tab:main_reclor}
    \centering
    \begin{threeparttable}
    \begin{tabular}{lcccc}% {|l|c|c|c|c|}
        % \hline
        \toprule
        \multirow{2}*{\textbf{Method}} & \multicolumn{4}{c}{\textbf{ReClor}} \\ % & \multicolumn{2}{|c}{LogiQA} \\
        \cmidrule{2-5}
         & \textbf{Dev} & \textbf{Test} & \textbf{Test-E} & \textbf{Test-H} \\ % & Dev & Test \\
        \midrule % \hline
        % \multirow{8}*{Baselines} & 
        Chance & 25.00 & 25.00 & 25.00 & 25.00 \\ % & 25.00 & 25.00 \\
        Human & - & 63.00 & 57.10 & 67.20 \\
        Ceiling Performance & - & 100.00 & 100.00 & 100.00 \\
        \midrule % \hline
         \textbf{\textit{Semantic Matching}} \\        
         FastText \cite{joulin2017bag} & 32.00 & 30.80 & 40.20 & 23.40 \\ % & - & - \\
         Bi-LSTM                       & 27.80 & 27.00 & 26.40 & 27.50 \\ % & - & - \\
        \midrule % \hline
        % \multirow{7}*{\shortstack{Pre-trained\\Models}} 
        % copy from the reclor paper
        \textbf{\textit{Transformer-based PLMs}} \\
         GPT \cite{radford2018improving}                & 47.60 & 45.40 & 73.00 & 23.80 \\ % & - & - \\
         GPT-2 \cite{radford2019language}               & 52.60 & 47.20 & 73.00 & 27.00 \\ % & - & - \\
        % copy from DAGN paper
         BERT-Large-MC \cite{devlin2019bert}               & 53.80 & 49.80 & 72.00 & 32.30 \\ % & 34.10 & 31.03 \\
         RoBERTa-Large-MC \cite{liu2019roberta,cui2019pre} & 62.60 & 55.60 & 75.50 & 40.00 \\ % & 35.02 & 35.33 \\
        \midrule
        \textbf{\textit{Graph Models}} \\
        Focal Reasoner \hl{(\robertalarge)} \cite{ouyang2021fact}                     & 66.80 & 58.90 & 77.05 & 44.64 \\
        \rowcolor{mypink} {DAGNs \hl{(\robertalarge)}}                       & {66.80} & {61.00} & {79.09} & {46.79} \\        
        \midrule
        \textbf{\textit{Data-Augmented Methods}} \\
        LReasoner \hl{(\robertalarge)} \cite{wang2021logic}                          & 66.20 & 62.40 & 81.40 & 47.50 \\
        LReasoner \hl{(\robertalarge)}$^{\dag}$                                      & 64.70 & 58.30 & 77.60 & 43.10 \\
        MERIt \hl{(\robertalarge)} \cite{jiao-etal-2022-merit}                       & 67.80 & 60.70 & 79.60 & 45.99 \\
        \rowcolor{mypink} {DAGNs + LReasoner} \hl{(\robertalarge)}           & {69.00} & {61.90} & {79.55} & {48.04} \\        
        \rowcolor{mypink} {DAGNs + MERIt} \hl{(\robertalarge)}               & {68.40} & {62.40} & {80.45} & {48.21} \\
        \bottomrule
    \end{tabular}
    \begin{tablenotes}
        \item[$\dag$] {This result is reproduced and reported by MERIt \cite{jiao-etal-2022-merit}}. 
    \end{tablenotes}
    \end{threeparttable}
    \vspace{0.5cm}
\end{table}

% table 3: logiqa main result.
\begin{table}[!t]
    % \footnotesize
    \caption{
    Experimental results on LogiQA dataset.
    Accuracies (\%) are reported.
    % \todo{DAGN+ denotes the fine-grained version of DAGN.}
    }
    \label{tab:main_logiqa}
    \centering
    \begin{threeparttable}
    \begin{tabular}{lcc}%{|l|c|c|%c|c|c|c}
        \toprule % \hline
        \multirow{2}*{\textbf{Method}} & \multicolumn{2}{c}{\textbf{LogiQA}} \\
        \cmidrule{2-3}
         & \textbf{Dev} & \textbf{Test} \\
        \midrule % \hline
        % \multirow{8}*{Baselines} & 
        Chance & 25.00 & 25.00 \\
        Human & - & 86.00 \\
        Ceiling & - & 95.00 \\
        \midrule % \hline
         % copy from the logiqa paper.
        %  \textbf{\textit{Rule-Based Methods}} \\
         \textbf{\textit{Lexical Matching}} \\
         % LogiQA: Rule-based
         Word Matching \cite{yih2013question}       & 27.49 & 28.37 \\
         Sliding Window \cite{richardson-etal-2013-mctest} & 23.58 & 22.51 \\
        \midrule
         \textbf{\textit{Deep QA Systems}} \\
         % LogiQA: Deep Learning
         Stanford Attentive Reader \cite{chen2016thorough} & 29.65 & 28.76 \\
         Gated-Attention Reader \cite{dhingra2017gated}    & 28.30 & 28.98 \\
         Co-Matching Network \cite{dhingra2017gated}       & 33.90 & 31.10 \\
        \midrule % \hline
        \textbf{\textit{Transformer-based PLMs}} \\
        % copy from DAGN paper
         BERT-Large-MC \cite{devlin2019bert}               & 34.10 & 31.03 \\
         RoBERTa-Large-MC \cite{liu2019roberta,cui2019pre} & 35.02 & 35.33 \\        
        \midrule
        \textbf{\textit{Graph Models}} \\
        Focal Reasoner \hl{(\robertalarge)} \cite{ouyang2021fact}                & 41.01 & 40.25 \\
        \rowcolor{mypink} {DAGNs \hl{(\robertalarge)}}                        & {{39.63}} & {{42.09}} \\
        \midrule
        \textbf{\textit{Data-Augmented Methods}} \\
        LReasoner \hl{(\robertalarge)}$^{\ddag}$ \cite{wang2021logic}            & 36.10 & 38.86 \\
        MERIt \hl{(\robertalarge)} \cite{jiao-etal-2022-merit}                   & 42.40 & 41.50 \\
        \rowcolor{mypink} {DAGNs + LReasoner \hl{(\robertalarge)}}            & {40.86} & {42.24} \\
        % \rowcolor{mypink} {DAGNs + MERIt}               & {43.01*} & {40.86*} \\
        \bottomrule % \hline
    \end{tabular}
    \begin{tablenotes}
    \item[$\ddag$] {We applied the official code on the LogiQA data.} 
        % \red{Question: LReasoner results currently have three different sources. Do we mention this here?}
    \end{tablenotes}
    \end{threeparttable}
    \vspace{0.5cm}
\end{table}

MuTual \cite{cui2020mutual} is a multi-turn dialogue reasoning dataset 
% with a response selection task.  
that evaluates logical reasoning in retrieval-based dialogue systems. 
% that requires predicting the subsequent response in an ongoing dialogue. 
% The task is 
The response selection task has 
% The response is chosen from 
four candidate responses for each dialogue, all relevant to the dialogue context, but only one is logically correct. The distracting answers are highly lexically overlapped with the context; hence it is challenging to solve text matching solely.
%
% The dataset has a modified MuTual$^\text{plus}$ version that, for each question, one of the candidate responses is replaced by a safe response (e.g., ``Could you repeat that?''). 
The modified version MuTual$^\text{plus}$ includes a safe response (e.g., ``Could you repeat that?'') among the candidates and is more challenging in logical reasoning. 
% this task is challenging to text matching approaches. 
The evaluation metrics include recall at position 1 (R@1), recall at position 2 (R@2), and Mean Reciprocal Rank (MRR) in 4 candidate responses.
% The dataset has a trivial MuTual version and a 
Since the passages are dialogues between two speakers, this dataset has more verbal and informal texts than ReClor and LogiQA.
The dataset is randomly split into training, development, and test sets with an 8:1:1 ratio.

% table 4: mutual main results.
\begin{table*}[!t]
    \caption{
    Experimental results on the MuTual development set.
    Recalls (R@1, R@2) and Mean Reciprocal Rank (MRR) are reported. 
    % \todo{DAGN+ denotes the fine-grained version of DAGN.}
    % \xd{should explain why DAGN+ is worse than DAGN in experiments}
    }
    \label{tab:main_mutual}
    \centering
    \begin{tabular}{lcccccc} % {|l|c|c|c|c|c|c|}
        \toprule % \hline
        \multirow{2}*{\textbf{Method}} & \multicolumn{3}{c}{\textbf{MuTual}} & \multicolumn{3}{c}{\textbf{MuTual$^\text{plus}$}} \\
        \cmidrule{2-7}
         & \textbf{R@1} & \textbf{R@2} & \textbf{MRR} & \textbf{R@1} & \textbf{R@2} & \textbf{MRR} \\
        \midrule % \hline        
        % copy from the MuTual paper
        Chance                                  & 25.00 & 50.00 & 60.40 & 25.00 & 50.00 & 60.40 \\
        % \midrule % \hline
        TF-IDF                                  & 27.60 & 54.10 & 54.10 & 28.30 & 53.00 & 76.30 \\
        % \midrule
        % \textbf{\textit{Dialogue Systems}} \\
        Dual LSTM \cite{lowe-etal-2015-ubuntu}  & 26.60 & 52.80 & 53.80 & - & - & - \\
        SMN \cite{wu-etal-2017-sequential}      & 27.40 & 52.40 & 57.50 & 26.40 & 52.40 & 57.80 \\
        DAM \cite{zhou-etal-2018-multi}         & 23.90 & 46.30 & 57.50 & 26.10 & 52.00 & 64.50 \\
        \midrule % \hline
        \textbf{\textit{Transformer-based PLMs}} \\
        BERT-Base \cite{devlin2019bert}         & 65.70 & 86.70 & 80.30 & 51.40 & 78.70 & 71.50 \\
        RoBERTa-Base \cite{liu2019roberta}      & 69.50 & 87.80 & 82.40 & 62.20 & 85.30 & 78.20 \\
        GPT-2 \cite{radford2019language}        & 33.50 & 59.50 & 58.60 & 30.50 & 56.50 & 56.20 \\
        GPT-2-FT \cite{radford2019language}     & 39.80 & 64.60 & 62.80 & 22.60 & 57.70 & 52.80 \\
        % \midrule % \hline
        BERT-Base-MC \cite{devlin2019bert}           & 66.10 & 87.10 & 80.60 & 58.60 & 79.10 & 75.10 \\
        RoBERTa-Base-MC \cite{liu2019roberta}        & 69.30 & 88.70 & 82.50 & 62.10 & 83.00 & 77.80 \\
        % \midrule % \hline
        RoBERTa-Large-MC \cite{liu2019roberta}       & 85.10 & 94.47 & 91.63 & 73.25 & 91.76 & 85.11 \\
        % \midrule % \hline
        % \textbf{\textit{Open Methods}} \\
        % Focal Reasoner \cite{ouyang2021fact}         & 73.40 & 90.30 & 84.90 & 63.70 & 86.10 & 79.10 \\
        \midrule
        \textbf{\textit{Dialogue Systems}} \\
        {Focal Reasoner (RoBERTa$_\text{Base}$) \cite{ouyang2021fact}}       & {73.40} & {90.30} & {84.90} & {63.70} & {86.10} & {79.10} \\
        {MDFN (\hl{\robertalarge}) \cite{Liu_Zhang_Zhao_Zhou_Zhou_2021}}  & {84.50} & {95.30} & {91.40} & {       -       } & {       -       } & {       -       } \\
        {MDFN (\hl{\electralarge}) \cite{Liu_Zhang_Zhao_Zhou_Zhou_2021}}  & {92.30} & {97.90} & {95.80} & {       -       } & {       -       } & {       -       } \\
        % {GRN (ALBERT) \cite{Liu_Feng_Wang_Song_Ren_Zhang_2021}} & {91.50} & {98.30} & {95.40} & {84.10} & {95.70} & {91.30} \\
        \midrule
        \textbf{\textit{Ours}} \\
        % ours        
        \rowcolor{mypink} {DAGNs \hl{(\robertalarge)}} & {86.79} & {96.50} & {92.73} & {78.22} & {92.55} & {88.14} \\
        \rowcolor{mypink} {DAGNs \hl{(\electralarge)}} & {92.55} & {98.19} & {95.97} & {82.73} & {95.26} & {90.51} \\
        % \midrule
        % \textbf{\textit{Data-Augmented Methods}} \\
        % {GRN (ALBERT) \cite{Liu_Feng_Wang_Song_Ren_Zhang_2021}} & {91.50} & {98.30} & {95.40} & {84.10} & {95.70} & {91.30} \\
        \bottomrule % \hline
    \end{tabular}
\end{table*}

\subsection{Implementation Details} % section 4.2

\textbf{Logic Graph Construction. }
For multiple-choice QA (ReClor and LogiQA), each question contains a passage, a question, and several candidate options.
Similarly, each sample contains a dialogue context and multiple candidate responses in the dialogue reasoning dataset.
Therefore, considering the different contexts of the candidates, we construct logic graphs for each candidate by pairing every candidate with the passage and the question.

\noindent\textbf{Graph Reasoning. }
% {The edge reasoning is performed 2 hops. The edge-extraction threshold is set to 0.25. The edge repetition $d$ is set to 2. 
% }
% And the GNN iteration step is 2.
% % The graph layer number is 2.
% The hidden sizes in GRU and perceptron are also set to 1,024.
%
%
For ReClor and LogiQA, we set the maximum length of the input token sequence to 256. 
% The encoder we use is the RoBERTa-Large model \cite{liu2019roberta} with 24 hidden layers whose hidden size is 1,024.
% For multiple-choice QA, 
The input format is ``\texttt{<s> passage </s> question || option </s>}'', where \texttt{<s>} and \texttt{</s>} are the special tokens for RoBERTa \cite{liu2019roberta} model, and \texttt{||} denotes concatenation, following previous works \cite{yu2020reclor,liu2020logiqa}. 
And the number of stacked GNN layers is 2 for ReClor and 3 for LogiQA.
% The graph layer number is 2.
The model is optimized with AdamW \cite{DBLP:conf/iclr/LoshchilovH19} with the learning rates 1e-5 for graph reasoning and 5e-6 for parameters. The epsilon is set to 1e-6. A linear scheduler is used and the warmup steps are set to 4,000. 

For MuTual, the maximum input length is set to 320. For the dialogue context sequence, we insert a separator token (``\texttt{</s>}'' for RoBERTa and ``\texttt{[SEP]}'' for ELECTRA) between each adjacent utterance pair, following \cite{Liu_Zhang_Zhao_Zhou_Zhou_2021}. 
And the GNN iteration step is 1.
The model is optimized with AdamW \cite{DBLP:conf/iclr/LoshchilovH19} with a learning rate of 4e-6 and an epsilon of 1e-8. A linear scheduler is used and the warmup proportion is set to 1\%.

For all datasets, {the edge reasoning is performed in 2 hops. The edge-extraction threshold is set to 0.25. The edge repetition $d$ is set to 2. }
The hidden sizes in GRU and perceptron are also set to 1,024.
The weight decay is set to 0.01 for all.
The overall dropout rate is 10\%.
The model is trained for 30 epochs with a batch size of 16 on one Nvidia Tesla V100 GPU.

\subsection{Results in Supervised Scenarios}  % section 4.3
% \subsection{Comparison with Baselines in Supervised Scenarios} % section 4.3 
% \subsection{Comparison with Current Methods}
\subsubsection{ReClor Dataset} % section 4.3.1
% baselines 
\noindent\textbf{Compared Methods. }
FastText \cite{joulin2017bag} and Bi-LSTM learns semantics matching. FastText learns n-gram features for text classification, 
whereas Bi-LSTM learns contextual features with recurrent network architecture. 
Transformer-based pre-trained language models (PLMs) learn contextual embeddings from large-scale corpora. 
{We also compare with the state-of-the-art Focal Reasoner \cite{ouyang2021fact}, LReasoner \cite{wang2021logic}, and MERIt \cite{jiao-etal-2022-merit}.
The Focal Reasoner is a graph-based model that builds ad-hoc graphs with entity-based nodes and coreference edges.
The LReasoner trains the PLMs with a contrastive learning framework, and the negative samples are constructed by pre-defined logical expressions. 
MERIt performs domain-specific pre-training also in a contrastive learning manner, where the augmented data is constructed via graph meta-paths. 
To conduct fair comparisons with LReasoner and MERIt, we train DAGNs by including the augmented negative data. The resulting models are denoted as ``DAGNs + LReasoner'' and ``DAGNs + MERIt'', respectively. 
For ``DAGNs + LReasoner'', logic graphs for the negative samples are constructed in the same manner. Then the model is fine-tuned with the contrastive learning objective function follows \cite{wang2021logic}.
For ``DAGNs + MERIt'', logic graphs for the negative instances are constructed as usual. The model is fine-tuned with the pre-trained checkpoints from \cite{jiao-etal-2022-merit}.
}
\hl{The compared Focal Reasoner, LReasoner, MERIt, and the proposed DAGNs all use \robertalarge\ as the backbone PLM for a fair comparison. }

% results and analysis. table 2.
\noindent\textbf{Results. }
Table~\ref{tab:main_reclor} demonstrates the results on the ReClor dataset. 
{The DAGNs \hl{(\robertalarge)} outperform Focal Reasoner \hl{(\robertalarge)} in both ReClor and LogiQA, demonstrating the effectiveness of the logic graph-constrained learning. Moreover, the results of ``DAGNs + LReasoner \hl{(\robertalarge)}'' and ``DAGNs + MERIt \hl{(\robertalarge)}'' also outperform their counterparts. This indicates that the structural constraints are still beneficial regardless of training schemes.
Further, compared to the PLM counterpart \robertalarge, DAGNs \hl{(\robertalarge)} show significant improvements. This indicates that the logic graphs provide useful information beyond the contextual embeddings learned from the plain texts, which is beneficial to reasoning.
Moreover, the improvements on the test-HARD set are significant. DAGNs \hl{(\robertalarge)} achieve 46.79\%, which is comparable to the strong LReasoner \hl{(\robertalarge)} and MERIt \hl{(\robertalarge)} with augmented data. ``DAGNs + LReasoner \hl{(\robertalarge)}'' and ``DAGNs + MERIt \hl{(\robertalarge)}'' also show great improvements over LReasoner \hl{(\robertalarge)} and MERIt \hl{(\robertalarge)}, respectively. 
The overall observations indicate the effectiveness of DAGNs and the structural logic representations are beneficial for challenging reasoning questions.
}

\subsubsection{LogiQA Dataset} % section 4.3.2
\noindent\textbf{Compared Methods. }
The word matching \cite{yih2013question} and sliding window \cite{richardson-etal-2013-mctest} perform lexical matching between the passage-question pair and candidate answers. 
Deep QA systems, including Stanford Attentive Reader \cite{chen2016thorough}, Gated-Attention Reader \cite{dhingra2017gated}, and Co-Matching Network \cite{wang2018co} calculate semantic similarity or use fine-grained attention mechanisms to match the context and the candidate answers. The performances are around chance, which indicates that the lexical or semantic matching is insufficient for catching the logic behind the texts. 
% The Co-Matching Network performs slightly better, and the results are closer to PLMs. 
Transformer-based pre-trained language models (PLMs) perform better than lexical or semantic matching, but the results are still inferior. It is indicated that the powerful contextual embeddings partially help the logical reasoning QA, but the inferiority of the lack of logical structure is obvious.
{We also compare with the state-of-the-art methods Focal Reasoner \cite{ouyang2021fact}, LReasoner \cite{wang2021logic}, and MERIt \cite{jiao-etal-2022-merit}. }
\hl{Similar as in the ReClor dataset, the compared Focal Reasoner, LReasoner, MERIt, and the proposed DAGNs all use \robertalarge\ as the backbone PLM for a fair comparison.}

% table 5: zero-shot transfer - 1
\begin{table}[!t]
    \caption{
        Zero-shot transfer between ReClor and LogiQA compared with supervised learning results. 
        ``RoBERTa-L" means ``RoBERTa-Large".        
        ``DAGNs-CT'' indicates full training on the target dataset after zero-shot transfer. 
    }
    \label{tab:zeroshot_1}
    \centering
    \begin{tabular}{lcccccc} %{|l|c|c|c|c|c|c|}
        \toprule % \hline
        \multirow{2}*{\textbf{Method}} & \multicolumn{4}{c}{\textbf{LogiQA $\to$ ReClor}} & \multicolumn{2}{c}{\textbf{ReClor $\to$ LogiQA}} \\
        \cmidrule{2-7}
         & \textbf{Dev} & \textbf{Test} & \textbf{Test-E} & \textbf{Test-H} & \textbf{Dev} & \textbf{Test} \\
        \midrule % \hline
        RoBERTa-L                         & 41.40 & 38.30 & 41.82 & 35.54 & 35.79 & 37.94 \\
        \rowcolor{mypink} {DAGNs} & {{44.20}} & {{41.90}} & {{46.59}} & {{38.21}} & {41.47} & {39.94} \\
        \rowcolor{mypink} {DAGNs-CT} & {60.60} & {55.40} & {77.73} & {37.86} & {41.63} & {43.78} \\
        \midrule % \hline
        \multirow{2}*{\textbf{Method}} & \multicolumn{4}{c}{\textbf{ReClor}} & \multicolumn{2}{c}{\textbf{LogiQA}} \\
        \cmidrule{2-7}
         & \textbf{Dev} & \textbf{Test} & \textbf{Test-E} & \textbf{Test-H} & \textbf{Dev} & \textbf{Test} \\
        \midrule % \hline
        RoBERTa-L                         & 62.60 & 55.60 & 75.50 & 40.00 & 35.02 & 35.33 \\        
        % \rowcolor{mypink} {DAGNs}   & {65.00} & {60.50} & {79.55} & {45.54} & {{39.63}} & {{42.09}} \\
        \rowcolor{mypink} {DAGNs}   & \hl{66.80} & \hl{61.00} & \hl{79.09} & \hl{46.79} & {{39.63}} & {{42.09}} \\
        \bottomrule % \hline
    \end{tabular}
\end{table}

% table 6: zero-shot transfer - 2
\begin{table}[!t]
    \caption{
        Zero-shot transfer from ReClor to MuTual and LogiQA to Mutual compared with supervised learning results.
        ``DAGNs-CT'' indicates full training on the target dataset after zero-shot transfer. 
    }
    \label{tab:zeroshot_2}
    \centering
    \begin{tabular}{lcccccc} %{|l|c|c|c|c|c|c|}
        \toprule % \hline
        \multirow{2}*{\textbf{Method}} & \multicolumn{3}{c}{\textbf{ReClor $\to$ MuTual}} & \multicolumn{3}{c}{\textbf{ReClor $\to$ MuTual$^\text{plus}$}} \\
        \cmidrule{2-7}
         & \textbf{R@1} & \textbf{R@2} & \textbf{MRR} & \textbf{R@1} & \textbf{R@2} & \textbf{MRR} \\
        \midrule % \hline
        RoBERTa-Large                     & 41.31 & 69.64 & 64.61 & 37.25 & 63.43 & 61.21 \\
        % \rowcolor{mypink} DAGN            & 50.23 & 74.83 & 71.79 & 40.29 & 70.77 & 65.24 \\
        % \rowcolor{mypink} DAGNs           & 50.45 & 74.15 & 71.96 & 35.33 & 69.30 & 62.94 \\
        \rowcolor{mypink} {DAGNs} & {48.53} & {74.60} & {70.56} & {45.37} & {71.90} & {67.67} \\
        \rowcolor{mypink} {DAGNs-CT}      & {80.70} & {92.21} & {90.53} & {73.14} & {90.07} & {85.46} \\
        \midrule % \hline
        \multirow{2}*{\textbf{Method}} & \multicolumn{3}{c}{\textbf{LogiQA $\to$ MuTual}} & \multicolumn{3}{c}{\textbf{LogiQA $\to$ MuTual$^\text{plus}$}} \\
        \cmidrule{2-7}
         & \textbf{R@1} & \textbf{R@2} & \textbf{MRR} & \textbf{R@1} & \textbf{R@2} & \textbf{MRR} \\
        \midrule % \hline
        RoBERTa-Large                     & 25.96 & 51.58 & 58.00 & 21.44 & 48.53 & 55.68 \\
        % \rowcolor{mypink} DAGN            & 54.06 & 78.56 & 74.50 & 45.15 & 72.80 & 68.14 \\
        % \rowcolor{mypink} DAGNs           & 52.71 & 77.88 & 73.77 & 38.60 & 71.67 & 65.04 \\
        \rowcolor{mypink} {DAGNs} & {{58.24}} & {{81.38}} & {{76.94}} & {{48.42}} & {{75.73}} & {{70.14}} \\
        \rowcolor{mypink} {DAGNs-CT}      & {83.63} & {93.91} & {91.91} & {74.15} & {90.97} & {85.98} \\
        \midrule % \hline
        \multirow{2}*{\textbf{Method}} & \multicolumn{3}{c}{\textbf{MuTual}} & \multicolumn{3}{c}{\textbf{MuTual$^\text{plus}$}} \\
        \cmidrule{2-7}
         & \textbf{R@1} & \textbf{R@2} & \textbf{MRR} & \textbf{R@1} & \textbf{R@2} & \textbf{MRR} \\
        \midrule % \hline
        RoBERTa-Large                     & 85.10 & 94.47 & 91.63 & 73.25 & 91.76 & 85.11 \\
        % \rowcolor{mypink} DAGN            & 86.57 & 95.03 & 93.96 & 77.65 & 92.10 & 88.11 \\
        % \rowcolor{mypink} DAGNs           & 84.88 & 94.70 & 93.14 & 76.30 & 91.87 & 87.94 \\
        % \rowcolor{mypink} {DAGNs} & {86.23} & {94.24} & {93.34} & {78.22} & {92.55} & {88.14} \\
        \rowcolor{mypink} {DAGNs} & \hl{86.79} & \hl{96.50} & \hl{92.73} & {78.22} & {92.55} & {88.14} \\
        \bottomrule % \hline
    \end{tabular}
    \vspace{0.5cm}
\end{table}

% results and analysis. table 3.
\noindent\textbf{Results. }
Table~\ref{tab:main_logiqa} shows the results on the LogiQA dataset.
{
DAGNs \hl{(\robertalarge)} also outperform the Focal Reasoner \hl{(\robertalarge)} on the LogiQA test set. Moreover, The DAGNs \hl{(\robertalarge)} also outperform the data-augmented LReasoner \hl{(\robertalarge)} and MERIt \hl{(\robertalarge)} on the test set. 
Furthermore, using the training paradigm in LReasoner, 
``DAGNs + LReasoner \hl{(\robertalarge)}'' also show superiority over LReasoner \hl{(\robertalarge)} and MERIt \hl{(\robertalarge)}. 
The results demonstrate that this method is generally effective for logical reasoning questions, regardless of training paradigms. 
The logic graph constraint provides beneficial guidance to representation learning and is superior to augmented plain texts. 
}

% table 7: ablation study - 2
\begin{table}[!t]
    \caption{
    Ablation of graph representation and structure on ReClor.
    % \todo{DAGN+ denotes the fine-grained version of DAGN.}
    }
    \label{tab:ablation_2}
    \centering
    \begin{tabular}{lcccc} %{|l|c|c|c|c|}
        \toprule % \hline
        \multirow{2}*{\textbf{Method}} & \multicolumn{4}{c}{\textbf{ReClor}} \\
        \cmidrule{2-5}
         & \textbf{Dev} & \textbf{Test} & \textbf{Test-E} & \textbf{Test-H} \\ 
        \midrule % \hline
            % \rowcolor{mypink} {DAGNs}               & {65.00} & {\textbf{60.50}} & {\textbf{79.55}} & {\textbf{45.54}} \\
            \rowcolor{mypink} {DAGNs}               & {{66.80}} & {{61.00}} & {{79.09}} & {{46.79}} \\
        \midrule % \hline
            \textit{\textbf{Graph representation}} & & & & \\
            \quad random node embeddings                    & 60.60 & 56.40 & 76.36 & 40.71 \\
        \midrule % \hline
            \textit{\textbf{Graph structure}} & & & & \\
            % \quad {{rule-based edges}}                   & {\textbf{67.40}} & {59.50} & {78.41} & {44.64} \\
            \quad {homogeneous variable edges}           & {63.60} & {59.30} & {77.73} & {44.82} \\
            \quad fully-connected edge linking              & 63.00 & 56.10 & 74.32 & 41.79 \\
            \quad random edge linking                       & 61.00 & 55.90 & 74.09 & 41.61 \\            
            \quad single edge types                         & 62.80 & 57.70 & 75.00 & 44.11 \\
            \quad clause nodes                              & 63.40 & 56.60 & 75.23 & 41.96 \\
            \quad sentence nodes                            & 60.40 & 57.30 & 74.32 & 43.93 \\
        \bottomrule % \hline
    \end{tabular}
\end{table}

% table 8: ablation study - 1
\begin{table}[!t]
    \caption{
    Ablation of model components (ReClor dev and test accuracy (\%)). 
    % MP indicates the \red{meta-path guided relation reasoning module}. 
    % MS indicates the edge-masking. 
    ER indicates the \newmoduleNoSpace.
    GB indicates the global graph representation. 
    NT indicates binary node types.
    VE indicates variable edges.
    % \todo{DAGN+ denotes the fine-grained version of DAGN.}
    }
    \label{tab:ablation_1}
    \centering
    \begin{tabular}{lcccc} %{|l|c|c|c|c|}
        \toprule % \hline
        \multirow{2}*{\textbf{Method}} & \multicolumn{4}{c}{\textbf{ReClor}} \\
        \cmidrule{2-5}
         & \textbf{Dev} & \textbf{Test} & \textbf{Test-E} & \textbf{Test-H} \\ 
        \midrule % \hline        
        \rowcolor{mypink} {DAGNs}    & {{66.80}} & {{61.00}} & {{79.09}} & {{46.79}} \\
        % \quad {w/o EM}               & {67.60} & {60.90} & {76.14} & {48.93} \\
        \quad {w/o GB}               & {62.00} & {60.10} & {77.73} & {46.25} \\
        \quad {w/o ER}               & {66.80} & {59.80} & {78.64} & {45.00} \\
        \quad {w/o ER, GB}           & {{67.40}} & {59.50} & {78.41} & {44.64} \\
        \quad w/o NT                                    & 63.40 & 58.70 & 77.05 & 44.29 \\
        \quad w/o NT, VE. \cite{huang-etal-2021-dagn}   & 65.20 & 58.20 & 76.14 & 44.11 \\
        \quad w/o graph reasoning                       & 55.20 & 52.00 & 74.77 & 34.11 \\
        \bottomrule
    \end{tabular}
    \vspace{0.5cm}
\end{table}

% % fig 5: ablation study - 2
\begin{figure*}[t!]
    \centering
    % \includegraphics[width=0.5\textwidth]{figs/line_num_gnn_layers.png}
    % \caption{Ablation of the number of GNN layers on the ReClor dataset.}
    % \includegraphics[width=\textwidth]{figs/mp_diff_er.png}
    \includegraphics[width=\textwidth]{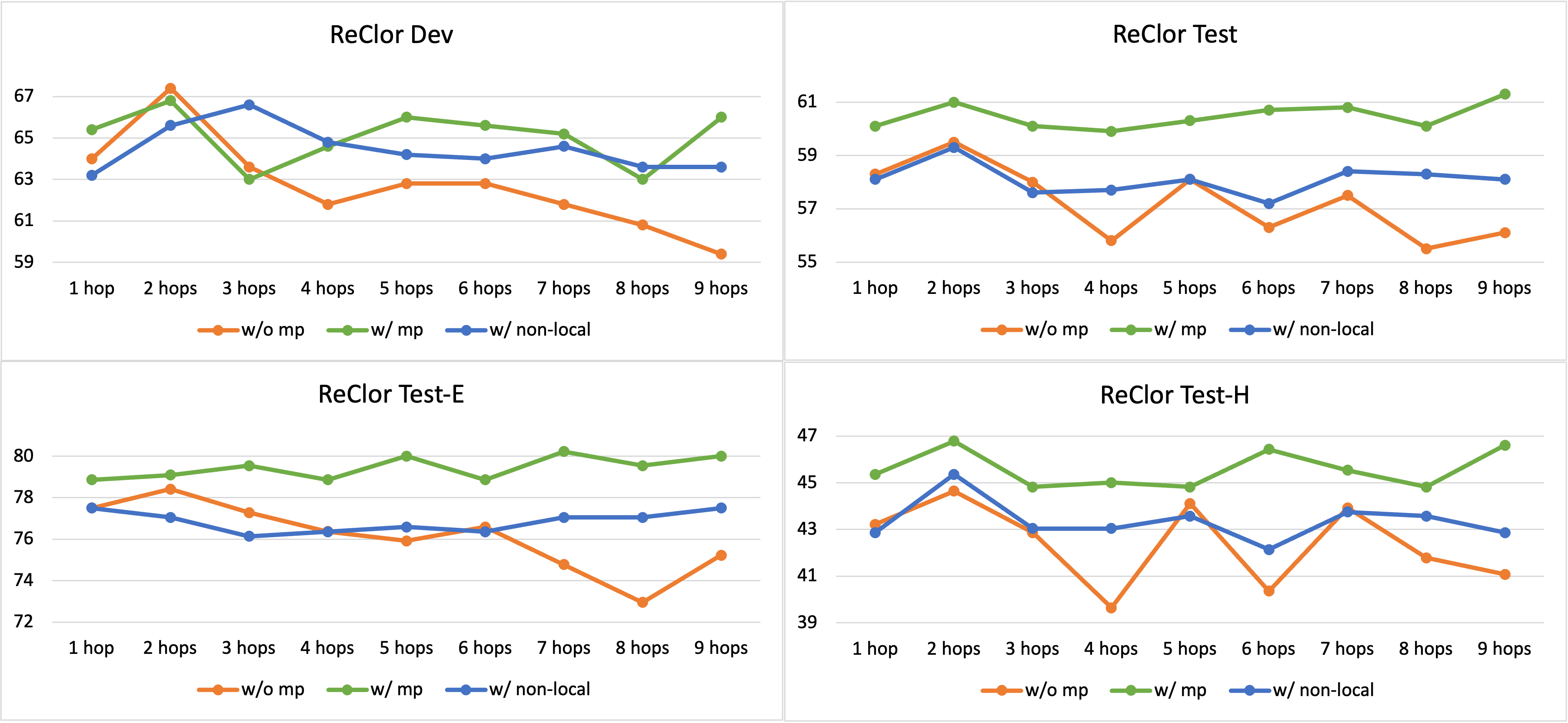}
    \caption{    
    \boxyl{Performance comparison among DAGNs, DAGNs with non-local GNNs, and DAGNs without the \newmodule over }
    \boxyl{multiple GNN layers. }
    }
    \label{fig:ablation_n_gnn_layers}
\end{figure*}

\subsubsection{MuTual Dataset} % section 4.3.3
% dataset.
% The MuTual dataset is different from the ReClor or the LogiQA dataset with more verbalized and informal texts in dialogue. It requires choosing logical correct response from four relative ones according to the dialogue context. 
\noindent\textbf{Compared Methods. }
% baseline methods.
The TF-IDF, Dual LSTM \cite{lowe-etal-2015-ubuntu}, SMN \cite{wu-etal-2017-sequential}, and DAM \cite{zhou-etal-2018-multi} conduct semantic text matching between dialogue context and candidate responses by using similarity of feature attention. According to the recall at positions 1 and 2, these methods select the correct responses by chance. The MRRs are also all lower than chance. This is not surprising considering the high lexical overlap between the context and the negative responses. 
For pre-trained LMs, GPT \cite{radford2018improving}, and GPT-2 \cite{radford2019language} perform as poorly as the text-matching methods, indicating that the generative models are inferior in reasoning. 
BERT \cite{devlin2019bert} and RoBERTa \cite{liu2019roberta} show better performances, especially the RoBERTa-Large model. 
{We also compare with Focal Reasoner \cite{ouyang2021fact} and MDFN \cite{Liu_Zhang_Zhao_Zhou_Zhou_2021}.
Focal Reasoner is a graph-based model with entity-based nodes and coreference relations.
MDFN uses multiple attention masks to decouple the contextual representations in utterance-aware and speaker-aware manners, then fuse the representation with a gate. 
}
\hl{We follow MDFN to use \robertalarge\ and \electralarge\ as the backbone PLMs for a fair comparison.}

% results and analysis. table 4.
\noindent\textbf{Results. }
Table~\ref{tab:main_mutual} shows the compared results on the MuTual datasets. 
{
DAGNs surpass the compared methods, including graph-based model and attention mask-based decoupling-fusion network. The results demonstrate that our proposed method is effective for less formal text such as multi-turn dialogue.
% the effectiveness of this method for less formal texts. 
% Both RoBERTa and ELECTRA 
% DAGNs outperform Focal Reasoner with entity-based graphs, which indicates that DAGNs learn more beneficial logic representations.
% fine-grained discourse-aware graphs provide useful information for logical reasoning. 
% the effectiveness of the fine-grained discourse-aware logic graph. 
% DAGNs also beat MDFN, 
%
% DAGNs with RoBERTa surpass MDFN on R@1 and MRR, and R@2 is also comparable. The results also significantly outperform the PLM baseline. Given that MuTual is a dialogue-understanding dataset with casual text, the results indicate that the logical structure is still beneficial.
} 

% DAGNs further improve their performances. The DAGN reaches 84.88 of R@1, 94.70 of R@2, and 93.14 of MRR on MuTual, and reaches 76.30 of R@1, 91.87 of R@2, and 87.94 of MRR on MuTual$^\text{plus}$. 
% %
% It is worth noting that on MuTual$^{plus}$, DAGNs is 3.05 in R@1 and 2.83 in MRR of improvements over the RoBERTa-Large-MC. Thus the provided logic structure information and learned logic representation even help the model detect those save responses, which is intuitive since the save responses are logically consistent with the dialogue context. 

\subsection{Results in Zero-shot Scenarios}  % section 4.4
% \subsection{Comparison with Baselines in Zero-shot Scenarios} % section 4.4
% \subsection{Logic Graph Structure Generalization Test}
We conduct zero-shot transfer experiments among the three datasets to see whether the constructed logic graph structure helps the models with unseen logical reasoning questions.
Considering the similarity between ReClor and LogiQA and the distinguishment of MuTual, we first train the models on LogiQA, then conduct direct testing on the ReClor development set and test set in a zero-shot manner, and vice versa. We then train the models on ReClor or LogiQA, respectively, then evaluate the MuTual development set in a zero-shot way. 
{For further comparison, we conduct continue full training on the target datasets.}
The results are demonstrated in Table~\ref{tab:zeroshot_1} and Table~\ref{tab:zeroshot_2}.

\subsubsection{Zero-shot Transfer between ReClor and LogiQA} % section 4.4.1
Comparing the results in the zero-shot setting and that in the supervised learning setting, it is surprising that the pre-trained LM and our DAGNs both show generality to some extent. 
% logiqa -> reclor v.s. reclor
Transferring from LogiQA to ReClor, RoBERTa-Large reaches 38.30\% in the test set, which is only 17.3 points behind that in the supervised learning setting. 
DAGNs (LogiQA $\to$ ReClor) achieve {41.90\%} in the test set compared to 59.50\% in the supervised learning setting. 
Interestingly, the generality in the EASY subset is harder. Both PLM and DAGNs accuracies are around 40\% in the zero-shot setting, but they achieve over 75\% in the supervised learning setting. But the transfer in the HARD subset does not lose much. The performances in the zero-shot setting are over 35\%, being comparable to the supervised-learning counterparts.
% relcor -> logiqa v.s. logiqa
Moreover, DAGNs (ReClor $\to$ LogiQA) achieve {41.47\%/39.94\%} on the development and test sets, comparable to {39.63\%/42.09\%} of the fine-tuning model.
% Surprisingly, the models perform better on the LogiQA dataset when trained on the ReClor dataset rather than the LogiQA dataset itself. 
% DAGNs v.s. PLM
The experimental results indicate that the generality of DAGNs is better than RoBERTa-Large, both transferring from LogiQA to ReClor and from ReClor to LogiQA.
It is indicated that the DAGNs improve the generality with the logic graphs and logic representations. 

{Moreover, after fine-tuning the zero-shot models on the target data, DAGNs-CT (ReClor $\to$ LogiQA) reaches 43.78\% on the test set, DAGNs-CT (LogiQA $\to$ ReClor) reaches 55.40\% on the test set, and over 30\% on the test-EASY set, while the performance on test-HARD is only 0.35\% inferior, which is still comparable. It is indicated that the transfer does not harm the performance given that the source and target data are different in reasoning types and data distribution. }

\subsubsection{Zero-shot Transfer to MuTual} % section 4.4.2
% To further explore the generality between different data, we conduct zero-shot transfer experiments from ReClor to MuTual and from LogiQA to MuTual respectively. 
% The MuTual data is more verbalized and informal since they are daily dialogue contents.
% Instead of predicting the logical relations close to first-order logic or the logical roles of the sentences in the context, the logical correctness of the response to the context is related to reasoning types such as attitude reasoning, algebraic reasoning, intention prediction, situation reasoning, multi-fact reasoning, which is more casual.

% between zero-shot setting and supervised learning setting. 
% The results indicate that all models show some generality to the MuTual dataset. 
The RoBERTa-Large struggles with the transfer, especially from LogiQA. RoBERTa-Large (LogiQA $\to$ MuTual/MuTual$^\text{plus}$) only achieves results around chance.
% The RoBERTa-Large generalizes fine from ReClor to MuTual/MuTual$^\text{plus}$. 
% But the performances of RoBERTa-Large (LogiQA $\to$ MuTual/MuTual$^\text{plus}$) are only around chance.
This may be due to that the MuTual dataset shares less familiarity with the ReClor or LogiQA dataset, and the ReClor dataset is more challenging with more complex sentences and logical structures, so learning from ReClor makes solving the MuTual dataset easier. In contrast, the LogiQA provides less beneficial structural information for solving MuTual.
% On the other hand, the LogiQA dataset is less challenging and with less complex logic graphs, so the models learn little from it to solve the MuTual dataset.

% in zero-shot setting.
% In the ReClor $\to$ MuTual setting, DAGNs 
{DAGNs (ReClor $\to$ MuTual/MuTual$^\text{plus}$) outperform RoBERTa-Large (ReCLor $\to$ MuTual/MuTual$^\text{plus}$). Similar results are observed between DAGNs (LogiQA $\to$ MuTual/MuTual$^\text{plus}$) and RoBERTa-Large (LogiQA $\to$ MuTual/MuTual$^\text{plus}$).
The improvements of transferring to MuTual$^\text{plus}$ are more significant than transferring to MuTual, which relives the struggle of RoBERTa-Large. 
The observations are coherent with that in the ReClor $\to$ LogiQA setting. The results demonstrate that given MuTual/MuTual$^\text{plus}$ are significantly different from ReClor/LogiQA in data distribution and reasoning types, DAGNs show superiority in logical reasoning transfer. }

{Further fine-tuning on the MuTual/MuTual$^\text{plus}$ results in significant performance growth. R@1 of DAGNs-CT (ReClor/LogiQA $\to$ MuTual) are 80.70\% and 83.63\%, of DAGNs-CT (ReClor/LogiQA $\to$ MuTual$^\text{plus}$) are 73.14\% and 74.15\%, respectively, which are comparable to their fine-tuning counterparts. The result improvements further demonstrate that DAGNs learn beneficial and general logic representations. 
}

% the observation in ReClor $\to$ LogiQA, and the improvements of DAGNs are more significant.
% of the transfer between ReClor and LogiQA. Besides, the performance boost from RoBERTa-Large to DAGNs is more significant in MuTual than Mutual$^\text{plus}$.

% In the LogiQA $\to$ MuTual setting, the performance gain from RoBERTa-Large to DAGNs is even more apparent. 
% The observation between MuTual and MuTual$^\text{plus}$ is similar to that of ReClor $\to$ MuTual setting. 

% fig 6: cons, question types
\begin{figure}[t!]
    \centering
    \includegraphics[width=0.48\textwidth]{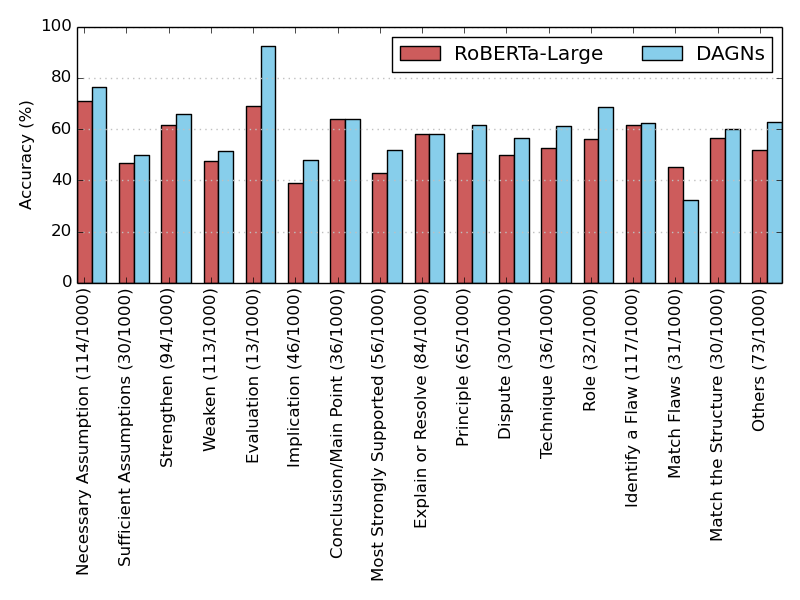}
    \caption{
    Performance comparison on question types in the ReClor test set. The numbers in parenthesis mean the number of samples in each question type over the test set scale.
    }    
    \label{fig:question_type}
    \vspace{0.5cm}
\end{figure}

\subsection{Ablation Study} % section 4.5
We conduct an ablation study further to explore the benefits of each part of our model. We take a close look at the model components, the importance of the graph components, and the effect of GNN layer stacks.

\subsubsection{Importance of Graph Components} % section 4.5.1
\label{sec:exp_graph_components}
We further validate each graph component.
Since the logic graph structure is significant to logical reasoning, we carefully modify the components of the logic graph and observe the performances.
The results are shown in Table~\ref{tab:ablation_2}.

We first vary the graph representation. The node embeddings in DAGNs are initialized with the EDU embeddings merged from the contextual token embeddings. We modify the pre-trained and merged EDU features with randomly initialized embeddings. The development set accuracy drops to 60.60\% and the test set to 56.40\%. It is worth noting that the accuracy in the HARD subset falls from 44.64\% to 40.71\%. It is a significant descent and demonstrates that the node features initialized from contextual embeddings are beneficial to logic graph reasoning.
% The EASY subset's accuracy slightly declines from the DAGN+ and is on par with DAGN. 

We then vary the graph structure by modifying the edges and the nodes. We make two changes to the edges: (1) modifying the edge linking and (2) modifying the edge type. 
For edge linking, 
{we first add variable edges within the context nodes and the candidate nodes (homogeneous variable edges), respectively. The performances drop to 63.60\% on the development set, and 59.30\% / 77.73\% / 44.82\% on the test / test-EASY / test-HARD sets, respectively. The results indicate that the homogeneous variable edges are redundant to the logic graphs. 
A possible reason is that the discourse connective edges within the context nodes and the candidate nodes are dense to some extent, so the homogeneous variable edges do not provide further information for the node feature update. 
}
{Then, }we ignore discourse relations and connect every pair of nodes, turning the graph fully connected. The resulting development accuracy drops to 63.00\%, and test accuracy drops to 56.10\%. 
Moreover, we remove all the edges from the logic graphs and randomly assign edges among the nodes with a Bernoulli distribution. The development set accuracy drops to 61.00\%, and the test set the precision to 55.90\%. 
The performances indicate that the fully-connected edge linking has unnecessary connections, while the random edge linking misses some linkings with helpful information. It reveals that in the logic graph we built, edges link EDUs in reasonable manners. 

For uncovering the contribution of edge types, instead of the differentiation of explicit discourse relations and implicit ones, all edges are regarded as a single type. With a single edge type, the model reaches 62.80\% on the development set and 57.70\% on the test set, which is 4.6\% and 1.8\% inferior to the entire model. 
% But the HARD subset accuracy is on par with that of DAGNs.
Therefore the two discourse-related edge types provide some helpful information to the model.
% , but lacking the types does not harm the performance much.

The nodes in the logic graphs act as reasoning units and are critical to logic representation learning. In substitution for EDUs, we use clauses or sentences as graph nodes. To obtain clause nodes, we remove “Explicit” connectives during discourse unit delimitation so that delimiters are only punctuation marks. For sentence nodes, we further reduce the delimiter library to solely period (“.”). The development and test accuracies drop to 63.40\% and 56.60\% with the modified graphs with clause nodes. When replaced with coarser sentence nodes, the performance drops to 60.40\% and 57.30\%. This indicates that clause or sentence nodes carry less discourse information and act poorly as logical reasoning units.

\subsubsection{Model Components} % section 4.5.2
To see the benefits of each component in the representation learning, 
we carefully remove them from the model, and the results on ReClor are shown in Table~\ref{tab:ablation_1}.
{
% We first remove the meta-path guided relation reasoning module. \red{xxx} 
% We first remove the edge-masking, and the test results slightly drop. 
%
We first remove the \newmoduleNoSpace, and the results drop to 66.80\%/59.80\%/78.64\%/45.00\% on the development/test/test-EASY/test-HARD sets, which indicates the effectiveness of the \newmoduleNoSpace.
Then, we remove the global node representation, from both full DAGNs and the DAGNs without \newmoduleNoSpace. Further performance drops are observed. It is indicated that the global graph representation catches some logical consistency between the context and candidates. 
}
% We first get rid of the passage-question/candidate option types. 
We then reduce the node types to only one type. As a result, the logic graphs only have a single node type. The dev accuracy drops dramatically from 67.40\% to 63.40\%. The test accuracy is slightly inferior, from 59.50\% to 58.70\%.
% , since the HARD subset drops merely 0.35\%\textbf{} and the EASY subset drops 1.36\%. 
Then, the variable edges are further removed from the model. The test accuracy further declines to 58.20\%.
% , and 76.14\% for the EASY 44.11\% for the HARD subset. 
Therefore, the logic graphs have reasonable structures for the logical reasoning QA task.

We further remove the whole graph reasoning operation. As a result, the hierarchical fusion is removed.
% This model solely contains an extra prediction module than the baseline.
% The performance on the ReClor dev set is between the baseline model and DAGN. Therefore, despite the prediction module beneﬁts the accuracy, t
The performance drops dramatically. It is indicated that the lack of graph reasoning leads to the absence of logic-aware features and degenerates the performance. It demonstrates the necessity of logical structures.
% discourse-based structure in logical reasoning.

% % fig 7
% % vis fig 1
% \begin{figure*}[ht]
%     \centering    
%     \includegraphics[width=\textwidth]{figs_vis/vis_edge_val_0.png}
%     \caption{
%         \boxyl{Visualization of learned hybrid edges from DAGNs.
%         In this case, the correct answer is option A.     
%         The DAGNs give the correct answer. }
%     }
%     \label{fig:vis_edge_1}
% \end{figure*}

% \vspace{7mm}
% % fig 8
% % vis fig 2
% \begin{figure*}[t!]
%     \centering    
%     \includegraphics[width=\textwidth]{figs_vis/vis_edge_val_351.png}
%     \caption{
%         \boxyl{Visualization of learned hybrid edges from DAGNs.
%         In this case, the correct answer is option C.     
%         The DAGNs give the correct answer. }
%     }
%     \label{fig:vis_edge_2}
% \end{figure*}

% fig 7
% vis fig 3
\begin{figure*}[t!]
    \centering    
    \includegraphics[width=\textwidth]{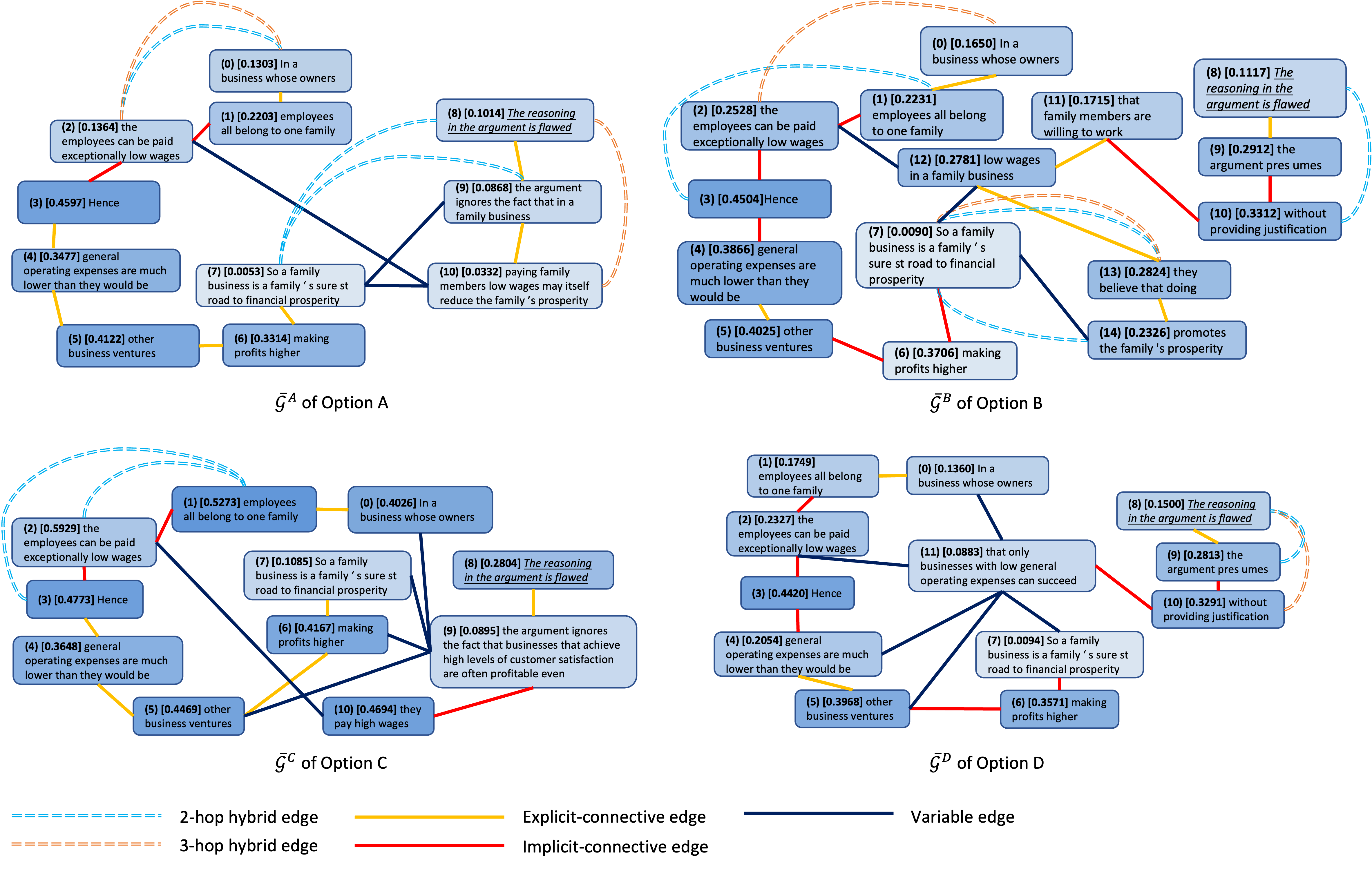}
    \caption{
    	\boxyl{Visualization of DAGNs with the learned hybrid edges and node weights. In this case, the correct answer is option A. } \\
    	\boxyl{The DAGNs give the correct answer.}    
    }
    \label{fig:vis_edge_1}
\end{figure*}

% fig 8
% vis fig 3
\begin{figure*}[t!]
    \setlength{\abovecaptionskip}{-1mm}
    \setlength{\belowcaptionskip}{-1mm}
    \centering    
    \includegraphics[width=\textwidth]{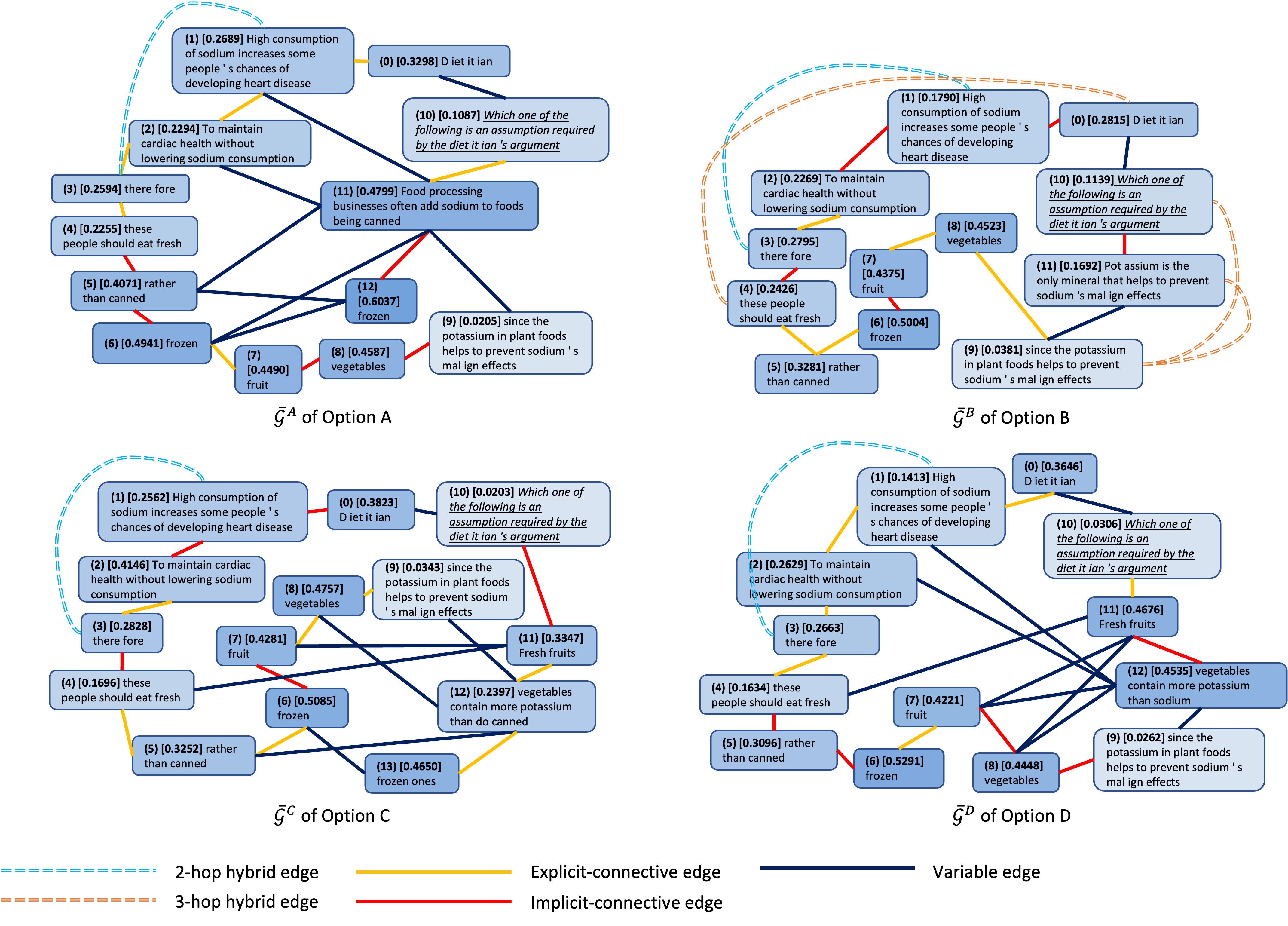}
    \caption{
    	\boxyl{Visualization of DAGNs with the learned hybrid edges and node weights. In this case, the correct answer is option C. } \\
        \boxyl{The DAGNs give the correct answer. }
    }
    \label{fig:vis_edge_2}
\end{figure*}

% fig 9
\begin{figure*}
% \captionsetup[subfigure]{labelformat=empty}
    % \setlength{\abovecaptionskip}{-1mm}
    \setlength{\belowcaptionskip}{-1mm}
	\centering     %%% not \center	
	\subfigure[\vspace{-3mm}DAGNs]{
            \includegraphics[width=.85\textwidth]{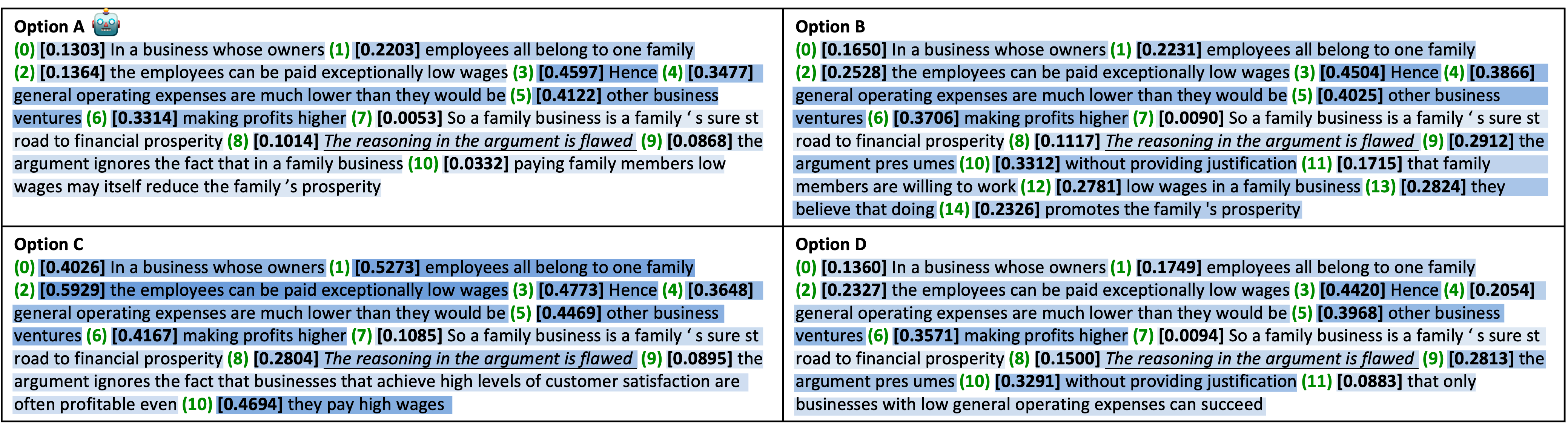}}
	% \vspace{-3mm}
	\subfigure[\vspace{-3mm}Zero-shot transfer (source: LogiQA)]{
            \includegraphics[width=.85\textwidth]{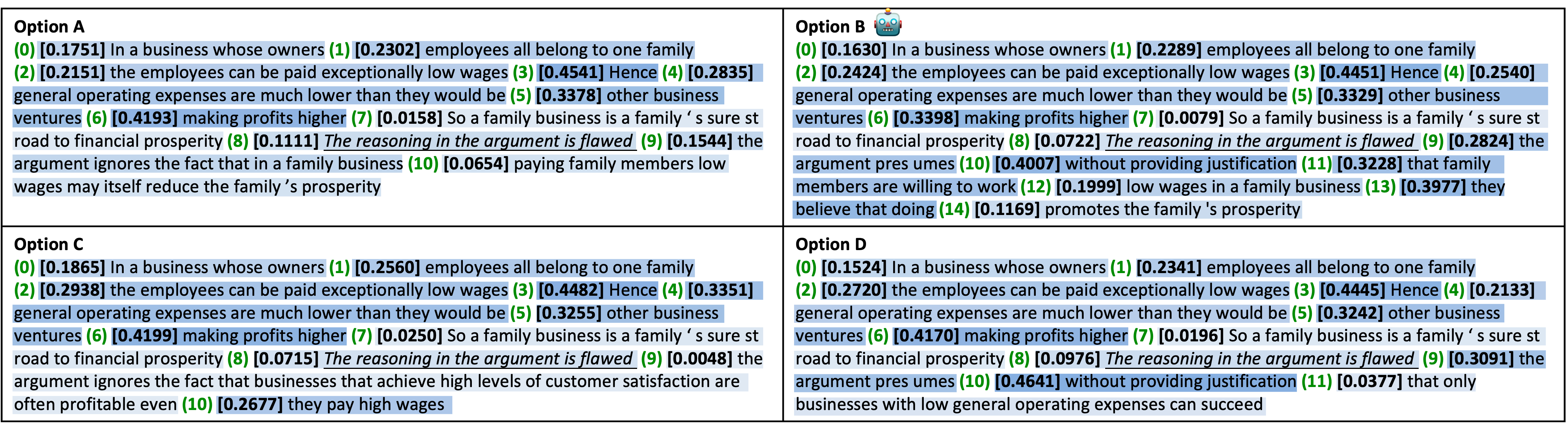}}
	% \vspace{-3mm}
	\subfigure[\vspace{-3mm}W/o edge-reasoning]{
            \includegraphics[width=.85\textwidth]{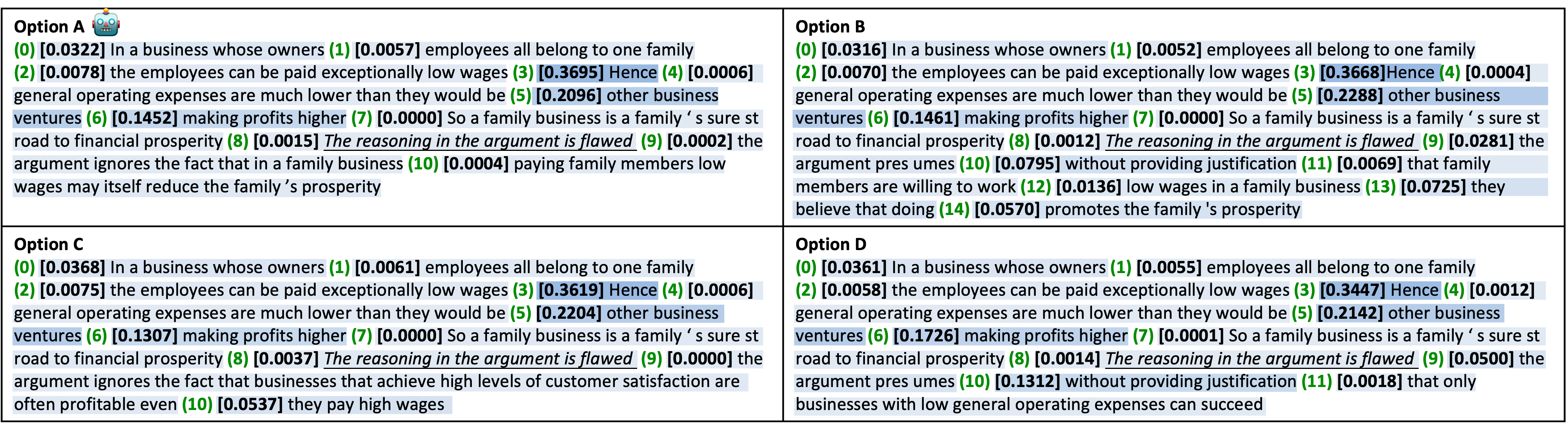}}
	% \vspace{-3mm}
	\subfigure[\vspace{-3mm}Fully-connected edge linking]{
            \includegraphics[width=.85\textwidth]{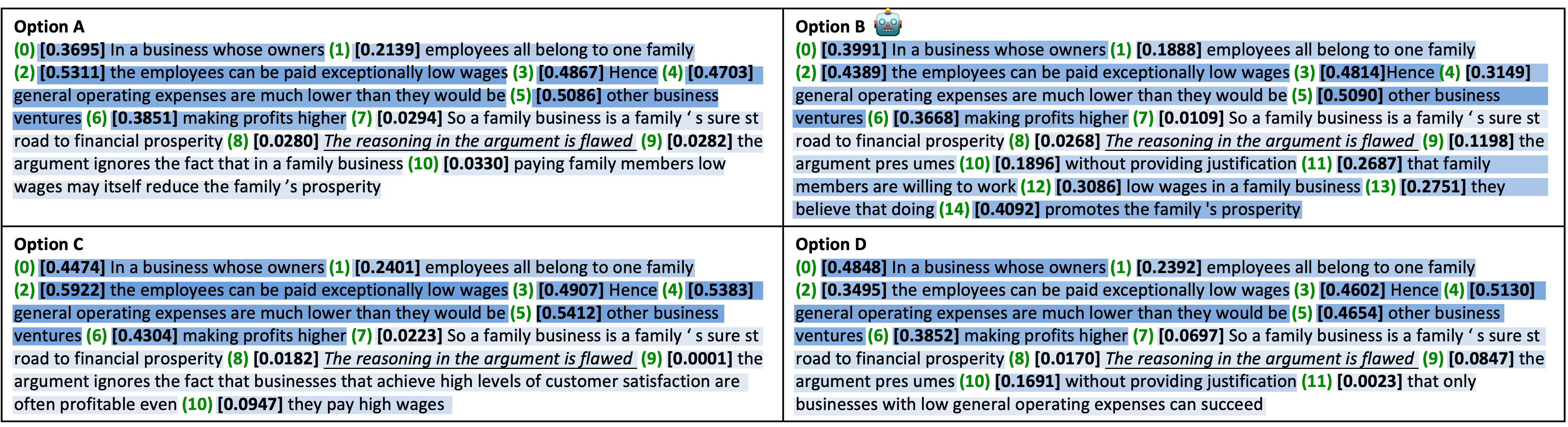}}
	% \vspace{-3mm}
	\subfigure[\vspace{-3mm}Sentence node]{
            \includegraphics[width=.85\textwidth]{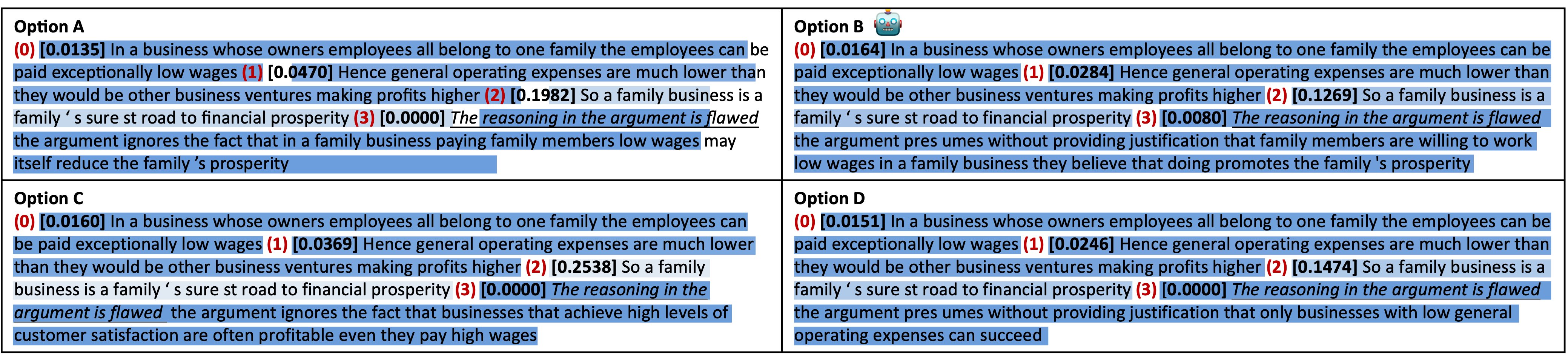}}	
	\vspace{-2mm}
	\caption{
	    \boxyl{Visualization of node weights learned from five models: DAGNs, DAGNs without the \newmoduleNoSpace, DAGNs with }
        \boxyl{fully-connected edge linking, DAGNs zero-shot transferred from LogiQA, and DAGNs with sentence nodes. In this case, the correct answer is }
        \boxyl{option A. In the passage, the EDU indices (*) in green are node delimitations from the full logic graph, and the indices in red are from the sentence } 
        \boxyl{nodes. The DAGNs, DAGNs w/o edge-reasoning give the correct answer. }
	}
	\label{fig:vis_dagns_g1}
\end{figure*}

% fig 10
\begin{figure*}
% \captionsetup[subfigure]{labelformat=empty}
    % \setlength{\abovecaptionskip}{-1mm}
    \setlength{\belowcaptionskip}{-1mm}
	\centering     %%% not \center	
	\subfigure[\vspace{-3mm}DAGNs]{
            \includegraphics[width=.88\textwidth]{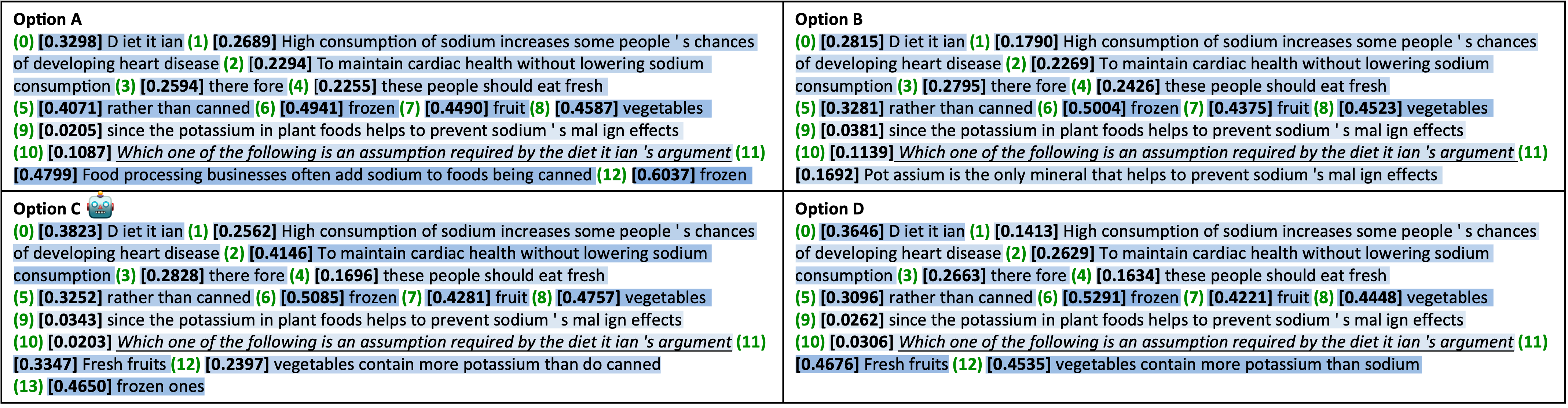}}
	% \vspace{-3mm}
	\subfigure[\vspace{-3mm}Zero-shot transfer (source: LogiQA)]{
            \includegraphics[width=.88\textwidth]{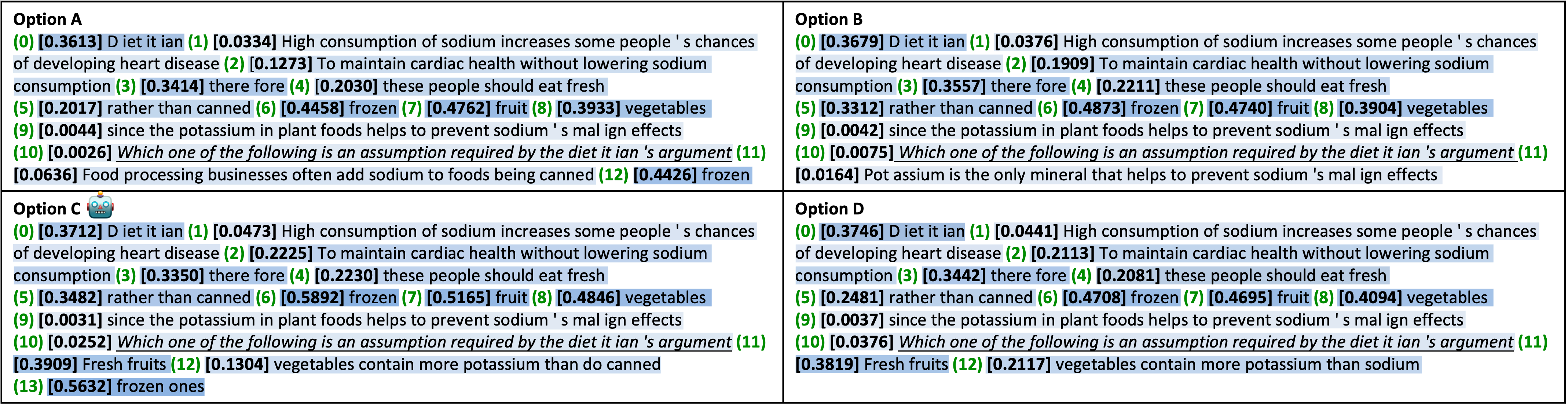}}
	% \vspace{-3mm}
	\subfigure[\vspace{-3mm}W/o edge-reasoning]{
            \includegraphics[width=.88\textwidth]{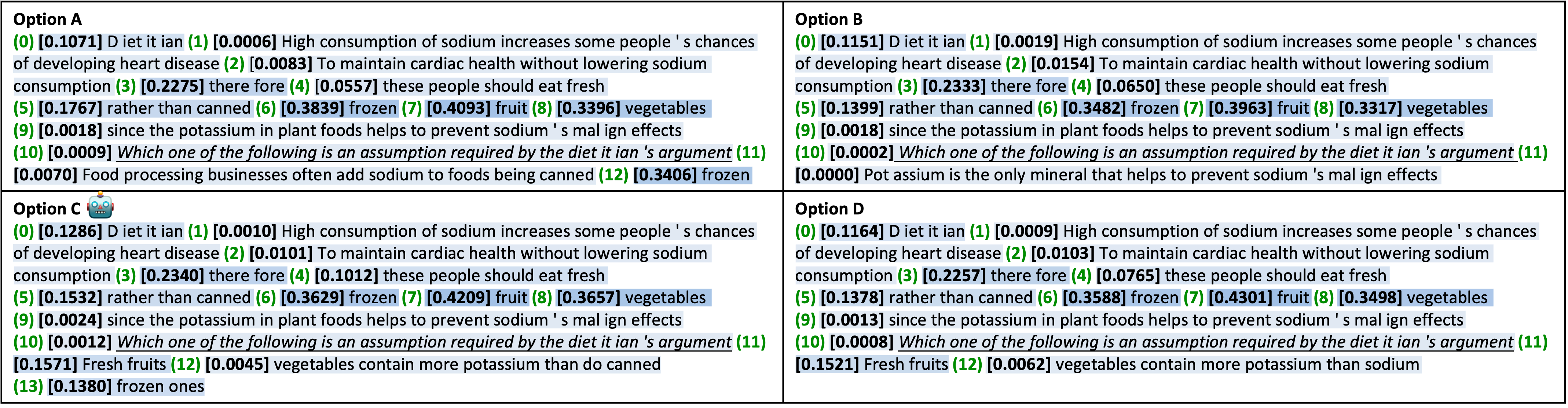}}
	% \vspace{-3mm}
	\subfigure[\vspace{-3mm}Fully-connected edge linking]{
            \includegraphics[width=.88\textwidth]{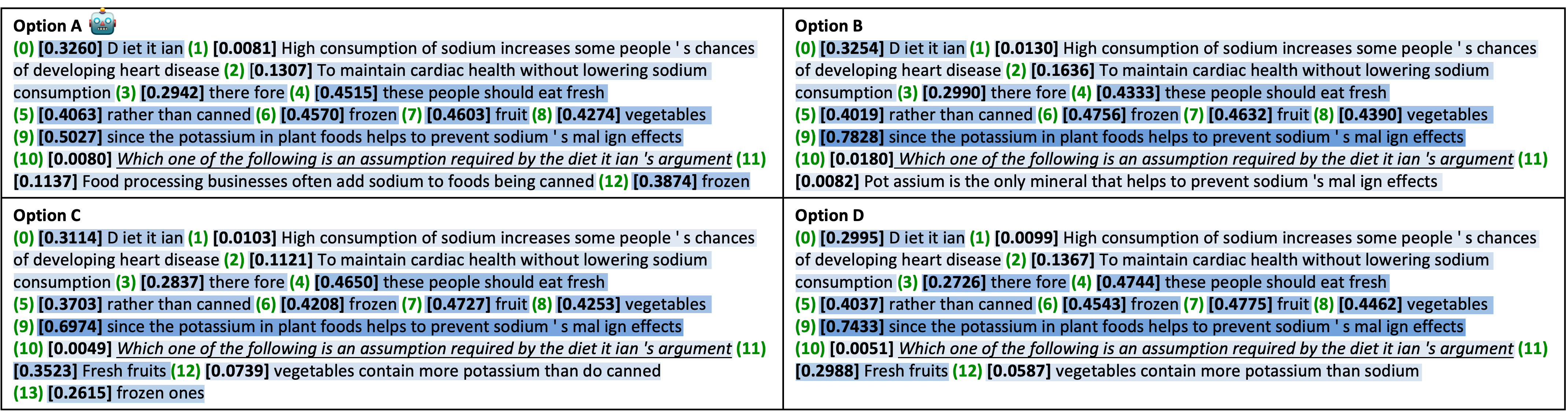}}
	% \vspace{-3mm}
	\subfigure[\vspace{-3mm}Sentence node]{
            \includegraphics[width=.88\textwidth]{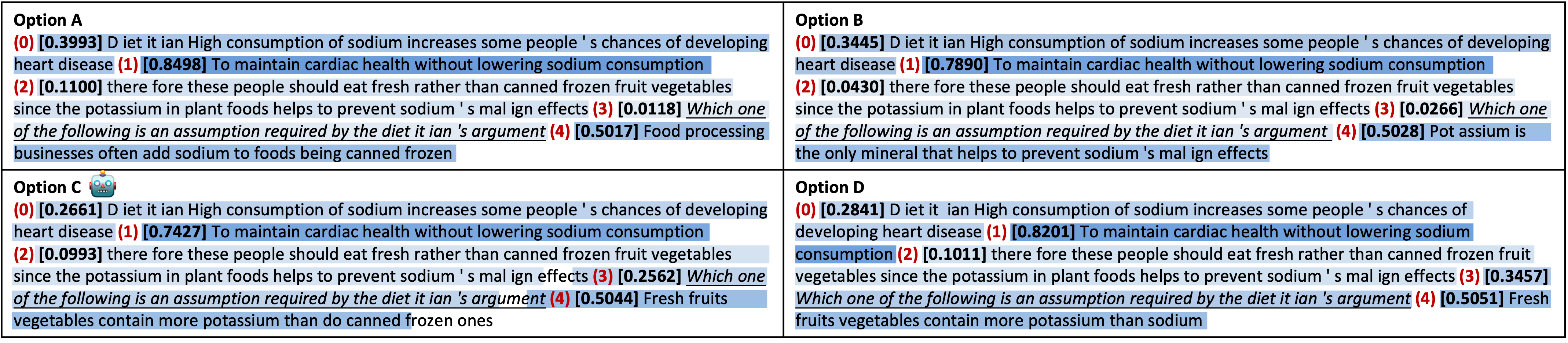}}	
	\vspace{-2mm}
	\caption{
	    \boxyl{Visualization of node weights learned from five models: DAGNs, DAGNs without the \newmoduleNoSpace, DAGNs with }
        \boxyl{fully-connected edge linking, DAGNs zero-shot transferred from LogiQA, and DAGNs with sentence nodes. In this case, the correct answer is }
        \boxyl{option A. In the passage, the EDU indices (*) in green are node delimitations from the full logic graph, and the indices in red are from the sentence } 
        \boxyl{nodes. The DAGNs, DAGNs w/o edge-reasoning give the correct answer. }
	}
	\label{fig:vis_dagns_g2}
\end{figure*}

% % fig 9
% % vis fig 3
% \begin{figure*}[t!]
%     \centering    
%     \includegraphics[width=\textwidth]{figs_vis/vis_node_val_0.png}
%     \caption{
%         \boxyl{Visualization of node weights learned from five models: DAGNs, DAGNs without the \newmoduleNoSpace, DAGNs with }
%         \boxyl{fully-connected edge linking, DAGNs zero-shot transferred from LogiQA, and DAGNs with sentence nodes. In this case, the correct answer is }
%         \boxyl{option A. In the passage, the EDU indices (*) in green are node delimitations from the full logic graph, and the indices in red are from the sentence } 
%         \boxyl{nodes. The DAGNs, DAGNs w/o edge-reasoning give the correct answer. }
%     }
%     \label{fig:vis_dagns_g1}
% \end{figure*}

% % fig 10
% % vis fig 4
% \begin{figure*}[t!]
%     \centering    
%     \includegraphics[width=\textwidth]{figs_vis/vis_node_val_351.png}
%     \caption{
%         \boxyl{Visualization of node weights learned from five models: DAGNs, DAGNs without the \newmoduleNoSpace, DAGNs with }
%         \boxyl{fully-connected edge linking, DAGNs zero-shot transferred from LogiQA and DAGNs with sentence nodes. In this case, the correct answer is} 
%         \boxyl{option C. 
%         In the passage, the EDU (*) in green are node delimitations from the full logic graph, and the indices in red are from the sentence} 
%         \boxyl{nodes. The DAGNs with fully-connected edge linking give a wrong answer. }
%     }
%     \label{fig:vis_dagns_g2}
% \end{figure*}

% \subsubsection{Effect of Multiple GNN Iterations}  % section 4.5.3
\subsubsection{Effect of GNN Layer Stacks} % section 4.5.3
% {TODO: ablation study. compare DAGNs w/ and w/o the mp module performs different on GNN layer stack. }
% We conduct experiments to see the effect of different numbers of GNN layers. 

% Since the logic representations are learned by a graph neural network with node feature aggregation, the number of steps that the node features are updated relates to the interaction among the EDU nodes and to what extent the logical reasoning units see each other. 
We change the number of the stacked GNN layers in our model to see how the graph reasoning steps affect the performances. 
{We compare the performances between the full DAGNs and the model without the \newmoduleNoSpace. } 
% \hl{
\hl{We run both the DAGNs with and without the \newmodule in this setting. }
\hl{Moreover, to compare the \newmodule with the non-local graph neural networks \cite{liu2021non,yu2020representative} for solving the over-smoothing problem \cite{li2018deeper,huang2020tackling} over the GNN layer stacks, we also compare with the DAGNs with  non-local GNNs \cite{liu2021non} as a replacement of the \newmoduleNoSpace.}
% }
The results are demonstrated in Figure~\ref{fig:ablation_n_gnn_layers}.

{Overall, the results show that the full DAGNs with the \newmodule perform steadily over multiple GNN layers, while the model without the \newmodule shows fluctuation and deterioration when the GNN iteration grows.}
{Specifically, the DAGNs without \newmodule reach peak performances with around two-step aggregation, after which decreases due to the general over-smoothing problem. In contrast, performance gains are observed in the full DAGNs, especially on the test and test-HARD sets. This indicates that shallow aggregation is insufficient for complex logical reasoning tasks, while the learnable \newmodule greatly relieves the over-smoothing problem in graph reasoning so that the model achieves deeper multi-hop reasoning as required. }
\hl{
One of the reasons is that the soft edge propagation in the edge-reasoning mechanism reasons new edges, which provides shortcuts for meaningful multi-hop relations and then accelerates the effective node feature update. As a result, the graph model takes fewer iterations to learn the multi-hop relations. 
}

\hl{Moreover, the DAGNs with non-local GNNs also show performance stability over multiple GNN layers, but are generally inferior to the DAGNs with the \newmoduleNoSpace.}
\hl{The results indicate that the edge-reasoning mechanism is superior in logical reasoning problems. 
The reason can be that the ``attention-guided sorting'' in the non-local GNNs pulls the distant nodes together according to a randomly initialized calibration vector, which is less informative than the edge propagation in the edge-reasoning mechanism to understand the logical relations.}

\subsection{Question Types} % section 4.6
The ReClor dataset contains multiple question types corresponding to diverse logical reasoning capabilities. We evaluate models in each question type, and the results are demonstrated in Figure~\ref{fig:question_type}.

Generally speaking, DAGNs perform better on most types of problems.
In question types such as ``Evaluation'', ``Technique'', and ``Most Strongly Supported'' that have high demands for knowledge of logical structures, the performance boosts over baseline models are significant. Therefore, the logic graphs are helpful to identify 
logical roles, such as the conclusion.
Moreover, the questions of ``Weakening'' and ``Implication'' are extremely challenging, and DAGNs also achieve improvements. It is indicated that the constructed logic graphs provide weakening relation and entailment relation information.
Other types of questions in which DAGNs perform well are ``Strengthen'', ``Conclusion/Main Point'', ``Explain or Resolve'', ``Principle'' and so forth.

In three question types, ``Match Flaws'', ``Identify a Flaw'', and ``Necessary Assumption'', DAGNs perform inferior to the RoBERTa-Large, especially in the challenging ``Match Flaws'' questions. Therefore, although the logic graphs and representation learning are beneficial overall, they do not cover each logical reasoning type. The ``Match Flaws'' questions also require awareness of logical structures and paring the structures in the passage and the options. Since the texts are logically flawed, the logic graphs directly constructed from the texts are not logically sound. Hence the learned logic representations are less desirable. 
% We also leave this to our future work.

\subsection{Visualization}  % section 4.7
\hl{To further investigate the interpretability of the model, we visualize the generated hybrid edges and the learned node weights, respectively. }

\hl{We first visualize the hybrid edges generated by the \newmoduleNoSpace, and two cases are shown in {Figures~\ref{fig:vis_edge_1} and \ref{fig:vis_edge_2}}. In option A in {Figure~\ref{fig:vis_edge_1}}, the \newmodule generates a 3-hop hybrid edge between node (8) and node (10), which bridges the question node and the key statement in the candidate. Moreover, the \newmodule learns the 2-hop edges (node (7), node (8)) and (node (7), node (9)). As node (7) is the conclusion, the \newmodule builds the hybrid edges for node (7) to help understand the argument and find the flaw. In contrast, in option C in this case, the key connections are lacking. }
\hl{Similarly, in the case of {Figure~\ref{fig:vis_edge_2}}, in option B, the model connects node (0) and node (4), which are the speaker and his/her opinion. The learned edges between node (9) and node (10), and between node (9) and node (11), builds connections between the speaker's opinion, the question, and the assumption in the candidate answer, which help find the inconsistency between (9) and (11). In option C, the graph has dense connections, especially between the context nodes and candidate nodes. The generated hybrid edges are relatively few. A reason is that the constructed edges are sufficient for identifying the logical consistency.}

\hl{Moreover, we visualize the graph node weights from multiple model variants, and two cases are presented in Figures~\ref{fig:vis_dagns_g1} and \ref{fig:vis_dagns_g2} here. The node weights are the $\alpha_i$ in Expression (7) demonstrated in Section 3.2.3 in the manuscript. The five model variants are (a) the full model DAGNs, (b) the DAGNs learned from LogiQA and then perform on the ReClor in a zero-shot manner, (c) the DAGNs without the \newmoduleNoSpace, (d) The DAGNs with fully-connected edge linking, and (e) the DAGNs with sentence nodes. 
{In Figure~\ref{fig:vis_dagns_g1}}, we observe that models (a) and (b) show better discrimination among the options as well as connections between the passage and the option. }
\hl{Interestingly, the zero-shot model (b) shows meaningful node attendance as the full-training model. 
Model (c) shows that without the learnable \newmoduleNoSpace, the model still being able to attend to the significant node such as node (3) that indicates an entailment, but the discrimination among the options is weaker. 
Model (d) shows that the model gives almost even attention to the sentences in the passage, nodes (2) and (5) have the highest weights, showing that the model is more interested in real entities and events, but it is less aware of the conclusion or premise. 
Model (e) with sentence nodes still attends to the conclusion sentence, but the coarse-grained delimitation does not provide sufficient information for telling the correct answer. }
\hl{Similar observations are found in Figure~\ref{fig:vis_dagns_g2}. The model (d) with fully-connected edge linking fails the question with vague discrimination among the nodes. }

\section{Related Works} % section 5
\label{sec:related_works}

\subsection{Textual Reasoning} % section 5.1
% \subsection{Reasoning QA} % section 5.1 
% \subsection{Language Reasoning}
% \subsection{Reasoning in Natural Language Understanding}
% \subsection{Textual Logical Reasoning \todo{(Logical Reasoning QA)}}
\label{sec:related_works_1}
% previous textual reasoning 
% There exist multiple textual reasoning tasks, being embedded in QA or dialogue datasets, the representative ones of which 
% Many existing QA tasks require reasoning, including 
% Language reasoning, especially reasoning for language comprehension, have been studied and evolved.

% related works start
{Textual reasoning tasks such as reasoning QA \cite{berant2013semantic,yang2018hotpotqa,talmor-etal-2019-commonsenseqa,dua2019drop}, Fact-Checking \cite{thorne2018fever}, and natural language inference (NLI) \cite{bowman2015large,williams2017broad} validate systems' reasoning with multiple schemes and granularity. Knowledge-based QA \cite{berant2013semantic,bao-etal-2016-constraint,yih-etal-2016-value,talmor-berant-2018-web,lopez2013evaluating,trivedi2017lc,dubey2019lc} provides large-scale knowledge bases \cite{freebase:datadumps,lehmann2015dbpedia} for question answering. Multi-hop QA \cite{yang2018hotpotqa,welbl2018constructing} requires models to reason over multiple documents and find supporting facts for the question. Commonsense reasoning QA \cite{talmor-etal-2019-commonsenseqa,huang2019cosmos,tandon2019wiqa} requires reasoning out the unstated world knowledge behind it. Moreover, Fact-Checking \cite{thorne2018fever} needs the models to retrieve supporting evidence for the given claims, while NLI \cite{bowman2015large,williams2017broad} requires the models to tell the inference relations between the given sentence pairs.}

{The previous QA \cite{yang2018hotpotqa,welbl2018constructing,talmor-etal-2019-commonsenseqa} and Fact-Checking tasks \cite{thorne2018fever} require models to retrieve supporting knowledge from a large set of documents. The models focus on effective knowledge retrieval and semantic matching. For example, HGN \cite{fang2020hierarchical} constructs hierarchical graphs to aggregate clues from the different granularity of evidence such as paragraph selection and supporting fact extraction. GEAR \cite{zhou-etal-2019-gear} constructs fully-connected evidence graphs with evidence-claim pairs as nodes for claim verification. Such models do not uncover text structures or simulate the reasoning processes with a given document. 
In contrast, solving logical reasoning questions requires the models to first reconstruct the structural reasoning process behind the text, and identify the logical components and relations, 
after which they can answer the questions about conclusion, assumption, argumentation strength, and logical fallacy. 
To achieve this, DAGNs use discourse-aware graphs to identify the logical components and use the variable edges to simulate the patterns. As a result, the graph reasoning under the structural constraints focuses on logic feature updates. Moreover, the \newmodule adapts the logical relations during training for more general logic representations.
%
% To achieve this, DAGNs use discourse-aware graphs to identify the logical components and relations. The plain texts are delimited into elementary discourse units as the logical components, which are then joined by discourse-connective edges to represent the logical relations and variable edges to represent the logical substitution relations. 
% The graph construction resorts to linguistic clues which are in accordance with the pragmatics \cite{toulmin2003uses,Freeman2011ArgumentSR} as discussed in Section 2.2.1, hence is general to textual logical reasoning problems. Moreover, the \newmodule in the DAGNs relieves the hand-craft construction and further improves the model generality. 
}

{On the other hand, 
previous tasks focus on the awareness of commonsense and world knowledge. 
For example, KagNet \cite{lin-etal-2019-kagnet} and MHGRN \cite{feng-etal-2020-scalable} encode subgraphs from ConceptNet \cite{speer2017conceptnet} and learn entity-based relational paths to answer commonsense questions. K-Adapter \cite{DBLP:conf/acl/WangTDWHJCJZ21} injects knowledge into pre-trained models. 
For solving NLI \cite{bowman2015large}, DRCN \cite{kim2019semantic} aggregates the semantics, while SemBERT \cite{zhang2020semantics} and SGNet \cite{zhang2020sg} learn semantics under different linguistic constraints. 
In contrast, logical reasoning QA focus on inference patterns rather than knowledge. 
% For example, the context in a logical reasoning QA can be reduced to 
For example in Figure~\ref{fig:intuition}, the correct inference pattern is the law of contraposition: ``if A implies B, then not-B implies not-A, and vice versa,'' which is the key to the question. 
The law is true regardless of the details in A and B. 
% Such laws are tautology, which means that it is true regardless of the content of A and B. 
Such knowledge-inference disentanglement provides generality to unseen reasoning data.
To this end, DAGNs is a pilot study for modeling the inference structure rather than focusing on knowledge. 
% It leverages the \newmodule to discover new logical connections given the constructed structure. As a result, it can learn general logic representations well on questions from different distributions.
}

{Moreover, the recent Focal Reasoner \cite{ouyang2021fact} for logical reasoning QA also performs graph reasoning. However, the constructed graphs extract entities and coreference relations following the previous QA models, which shows inferiority in capturing the logical relations between statements. 
Besides, LReasoner \cite{wang2021logic} trains the PLMs with a contrastive learning framework, and the negative samples are constructed by pre-defined logical expressions. The negative samples are derived by logical expressions. 
MERIt \cite{jiao-etal-2022-merit} performs domain-specific pre-training also in a contrastive learning manner, where the augmented data is constructed via graph meta-paths. }
% derives logical expressions and addresses question answering by training PLMs with contrastive learning aided by augmented data which is formed by the logical expressions. 
% It is suggested that pre-defined logic information is helpful.
%
However, the injected logic-biased data is in natural language format, and the model they use is plain PLM, which models the logical reasoning process implicitly.
% However, the logical information is in natural language format, and pure PLMs learn logical reasoning implicitly. 
It remains unclear how explicit logic formulation facilitates QA systems and what kind of logical structure is beneficial. 
Hence, in contrast, this paper focuses on logic-biased deep models that explicitly model the logical reasoning process and obtain the desired logic features. Our method also leverages PLMs but does not use augmented data. Hence this study is orthogonal to the previous \cite{wang2021logic}. 

\subsection{Discourse Applications} % section 5.2
\label{sec:related_works_3}
% \subsection{\el{Discourse Structures / Discourse-Aware Representations}}
% \subsection{\el{Sentence Representation}}
% Discourse parsing constructs the discourse structure of a text with defined discourse relations and EDUs.
% Discourse analysis \cite{ enwiki:1056527925} studies language use in passages. It is generally applied to multiple disciplines including linguistics and argumentation studies. It uncovers structures for texts, identifies roles and significant relations to understand the logic between lines.

Discourse information provides a high-level understanding of texts and hence is beneficial for many natural language tasks, for instance, text summarization \cite{cohan-etal-2018-discourse,joty-etal-2019-discourse,xu-etal-2020-discourse,feng2020dialogue}, neural machine translation \cite{voita2018context}, and coherent text generation \cite{bosselut2018discourse}. There are also discourse-based applications for reading comprehension. DISCERN \cite{gao2020discern} segments texts into EDUs and learns interactive EDU features. Mihaylov and Frank \cite{mihaylov-frank-2019-discourse} provide additional discourse-based annotations and encode them with discourse-aware self-attention models. 
However, such information is not yet considered in logical reasoning. Unlike previous works, this work builds discourse-aware logic graphs by first using discourse relations as graph edges that connect EDUs, then learning the discourse features via message passing with graph neural networks.

In natural language processing, the most influential theories of discourse structure are the Rhetorical Structure Theory (RST) \cite{mann1988rhetorical} and Lexicalized Tree-Adjoining Grammar for Discourse (DLTAG) \cite{webber2004d}. RST studies reconstructing tree-like structures for texts. The D-LTAG focuses on detecting discourse relations within local text units, and the units are disjoint sentences or two clauses in a sentence.
Inspired by the theories, several treebanks are constructed, and the most influential ones are RST-DT \cite{carlson2003building} and PDTB \cite{carlson2003building}. 
Models \cite{feng2014two,hayashi2016empirical,ji2014representation,kishimoto2020adapting,lei2018linguistic} are  trained on these treebanks to accomplish discourse parsing. And discourse parsing is also applied for downstream applications \cite{ji2017neural,liu2018learning,liu2019single,gao2020discern}.

However, current discourse parsers are primarily trained on small datasets via supervised learning, where the representative corpus is the 1 million-word Wall Street Journal (WSJ) Corpus. As a result, it is challenging for the parsers to transfer to unseen texts, especially in new topics or domains. Therefore, these parsers are not applicable for logical structure parsing.
In this paper, we customize rules to perform discourse segmentation and relation detection based on observations of the argument passages.

\section{Conclusion} % section 6
This paper explores a structure-based solution to textual logical reasoning that explicitly models the logical reasoning process. The challenges include: 
(1) Uncovering the inference structure from plain texts for effective structural constraints.
(2) Learning the inference processes rather than the knowledge for effective logical reasoning.
% (1) Reconstructing logical structures from the plain text that are reasonable and beneficial to downstream tasks, and such logical structures are beyond previous entity-based modeling;
% (2) Learning the structural logic features effectively for QA prediction and the model is efficiently trained.

To address the problems, we propose discourse-aware graph networks (DAGNs) with logic graph construction and logic representation learning. To construct beneficial logical structures, DAGNs get inspired by logic theories and convert plain text into logic graphs via several factors. The in-line discourse connectives indicate logical relations; hence are applied as text delimiters and split passages into clause-like logical units. Then the recurring topic-related terms are detected. The graph edges are two folds: the discourse connectives indicate logical relations, and the variable connection simulates logical expression derivation.

For learning the logic features, {DAGNs take the constructed graphs as input and perform soft edge selection and propagation to produce multi-hop hybrid relations. It then updates the node features via several steps of graph reasoning. } The graph network leverages contextual encoding and learns the logic representations, which are then fused for downstream prediction. 
% DAGNs perform graph reasoning over the constructed graphs for several steps to simulate the multi-hop logical reasoning process. 
% The graph network leverages contextual encoding; hence only needs a few rounds of fine-tuning and is widely applicable to general encoders. To apply the logic features, DAGNs fuse the features with the fundamental contextual embeddings and perform end-to-end training. 

Extensive experiments are conducted on two logical reasoning QA datasets and one multi-turn dialogue reasoning dataset. The results demonstrate the overall superiority of DAGNs. The constructed logic graph structure is reasonable, {and the \newmodule helps learn general logic representations and improves model stability. }
The zero-shot transfer results show that DAGNs perform remarkably well on unseen reasoning questions, which indicates that the learned logic representations are general in reasoning and beyond knowledge.

% \section*{Acknowledgments}

% The acknowledgments should go immediately before the references. Do not number the acknowledgments section.
% \textbf{Do not include this section when submitting your paper for review.}

% \bibliographystyle{acl_natbib}
% \bibliography{acl2021}

%\appendix

% \section{Conclusion}
% The conclusion goes here.

% or


\appendices


% % you can choose not to have a title for an appendix
% % if you want by leaving the argument blank
% \section{}

% % use section* for acknowledgment
% \ifCLASSOPTIONcompsoc
%   % The Computer Society usually uses the plural form
%   \section*{Acknowledgments}
% \else
%   % regular IEEE prefers the singular form
% \fi


% % Can use something like this to put references on a page
% % by themselves when using endfloat and the captionsoff option.
% \ifCLASSOPTIONcaptionsoff
%   \newpage
% \fi

% % trigger a \newpage just before the given reference
% % number - used to balance the columns on the last page
% % adjust value as needed - may need to be readjusted if
% % the document is modified later
% %\IEEEtriggeratref{8}
% % The "triggered" command can be changed if desired:
% %\IEEEtriggercmd{\enlargethispage{-5in}}


\bibliographystyle{IEEEtran}
\bibliography{r_baselines, r_datasets, r_discourse, r_gnns, r_others, r_train}

\begin{thebibliography}{10}
\providecommand{\url}[1]{#1}
\csname url@samestyle\endcsname
\providecommand{\newblock}{\relax}
\providecommand{\bibinfo}[2]{#2}
\providecommand{\BIBentrySTDinterwordspacing}{\spaceskip=0pt\relax}
\providecommand{\BIBentryALTinterwordstretchfactor}{4}
\providecommand{\BIBentryALTinterwordspacing}{\spaceskip=\fontdimen2\font plus
\BIBentryALTinterwordstretchfactor\fontdimen3\font minus
  \fontdimen4\font\relax}
\providecommand{\BIBforeignlanguage}[2]{{%
\expandafter\ifx\csname l@#1\endcsname\relax
\typeout{** WARNING: IEEEtran.bst: No hyphenation pattern has been}%
\typeout{** loaded for the language `#1'. Using the pattern for}%
\typeout{** the default language instead.}%
\else
\language=\csname l@#1\endcsname
\fi
#2}}
\providecommand{\BIBdecl}{\relax}
\BIBdecl

\bibitem{huang-etal-2021-dagn}
\BIBentryALTinterwordspacing
Y.~Huang, M.~Fang, Y.~Cao, L.~Wang, and X.~Liang, ``{DAGN}: Discourse-aware
  graph network for logical reasoning,'' in \emph{Proceedings of the 2021
  Conference of the North American Chapter of the Association for Computational
  Linguistics: Human Language Technologies}.\hskip 1em plus 0.5em minus
  0.4em\relax Online: Association for Computational Linguistics, Jun. 2021, pp.
  5848--5855. [Online]. Available:
  \url{https://www.aclweb.org/anthology/2021.naacl-main.467}
\BIBentrySTDinterwordspacing

\bibitem{yu2020reclor}
W.~{Yu}, Z.~{Jiang}, Y.~{Dong}, and J.~{Feng}, ``Reclor: A reading
  comprehension dataset requiring logical reasoning,'' in \emph{ICLR 2020 :
  Eighth International Conference on Learning Representations}, 2020.

\bibitem{liu2020logiqa}
J.~Liu, L.~Cui, H.~Liu, D.~Huang, Y.~Wang, and Y.~Zhang, ``Logiqa: A challenge
  dataset for machine reading comprehension with logical reasoning,''
  \emph{IJCAI 2020}, 2020.

\bibitem{cui2020mutual}
L.~Cui, Y.~Wu, S.~Liu, Y.~Zhang, and M.~Zhou, ``Mutual: A dataset for
  multi-turn dialogue reasoning,'' in \emph{Proceedings of the 58th Annual
  Meeting of the Association for Computational Linguistics}, 2020, pp.
  1406--1416.

\bibitem{chen2016thorough}
D.~Chen, J.~Bolton, and C.~D. Manning, ``A thorough examination of the
  cnn/daily mail reading comprehension task,'' in \emph{Proceedings of the 54th
  Annual Meeting of the Association for Computational Linguistics (Volume 1:
  Long Papers)}, 2016, pp. 2358--2367.

\bibitem{dhingra2017gated}
B.~Dhingra, H.~Liu, Z.~Yang, W.~Cohen, and R.~Salakhutdinov, ``Gated-attention
  readers for text comprehension,'' in \emph{Proceedings of the 55th Annual
  Meeting of the Association for Computational Linguistics (Volume 1: Long
  Papers)}, 2017, pp. 1832--1846.

\bibitem{wang2018co}
S.~Wang, M.~Yu, J.~Jiang, and S.~Chang, ``A co-matching model for multi-choice
  reading comprehension,'' in \emph{Proceedings of the 56th Annual Meeting of
  the Association for Computational Linguistics (Volume 2: Short Papers)},
  2018, pp. 746--751.

\bibitem{wu-etal-2017-sequential}
\BIBentryALTinterwordspacing
Y.~Wu, W.~Wu, C.~Xing, M.~Zhou, and Z.~Li, ``Sequential matching network: A new
  architecture for multi-turn response selection in retrieval-based chatbots,''
  in \emph{Proceedings of the 55th Annual Meeting of the Association for
  Computational Linguistics (Volume 1: Long Papers)}.\hskip 1em plus 0.5em
  minus 0.4em\relax Vancouver, Canada: Association for Computational
  Linguistics, Jul. 2017, pp. 496--505. [Online]. Available:
  \url{https://aclanthology.org/P17-1046}
\BIBentrySTDinterwordspacing

\bibitem{zhou-etal-2018-multi}
\BIBentryALTinterwordspacing
X.~Zhou, L.~Li, D.~Dong, Y.~Liu, Y.~Chen, W.~X. Zhao, D.~Yu, and H.~Wu,
  ``Multi-turn response selection for chatbots with deep attention matching
  network,'' in \emph{Proceedings of the 56th Annual Meeting of the Association
  for Computational Linguistics (Volume 1: Long Papers)}.\hskip 1em plus 0.5em
  minus 0.4em\relax Melbourne, Australia: Association for Computational
  Linguistics, Jul. 2018, pp. 1118--1127. [Online]. Available:
  \url{https://aclanthology.org/P18-1103}
\BIBentrySTDinterwordspacing

\bibitem{de2019question}
N.~De~Cao, W.~Aziz, and I.~Titov, ``Question answering by reasoning across
  documents with graph convolutional networks,'' in \emph{Proceedings of the
  2019 Conference of the North American Chapter of the Association for
  Computational Linguistics: Human Language Technologies, Volume 1 (Long and
  Short Papers)}, 2019, pp. 2306--2317.

\bibitem{qiu-etal-2019-dynamically}
L.~Qiu, Y.~Xiao, Y.~Qu, H.~Zhou, L.~Li, W.~Zhang, and Y.~Yu, ``Dynamically
  fused graph network for multi-hop reasoning,'' in \emph{Proceedings of the
  57th Annual Meeting of the Association for Computational Linguistics}.\hskip
  1em plus 0.5em minus 0.4em\relax Florence, Italy: Association for
  Computational Linguistics, Jul. 2019, pp. 6140--6150.

\bibitem{fang2020hierarchical}
Y.~Fang, S.~Sun, Z.~Gan, R.~Pillai, S.~Wang, and J.~Liu, ``Hierarchical graph
  network for multi-hop question answering,'' in \emph{Proceedings of the 2020
  Conference on Empirical Methods in Natural Language Processing (EMNLP)},
  2020, pp. 8823--8838.

\bibitem{zheng2020srlgrn}
C.~Zheng and P.~Kordjamshidi, ``Srlgrn: Semantic role labeling graph reasoning
  network,'' in \emph{Proceedings of the 2020 Conference on Empirical Methods
  in Natural Language Processing (EMNLP)}, 2020, pp. 8881--8891.

\bibitem{welbl2018constructing}
J.~Welbl, P.~Stenetorp, and S.~Riedel, ``Constructing datasets for multi-hop
  reading comprehension across documents,'' \emph{Transactions of the
  Association for Computational Linguistics}, vol.~6, pp. 287--302, 2018.

\bibitem{yang2018hotpotqa}
Z.~Yang, P.~Qi, S.~Zhang, Y.~Bengio, W.~Cohen, R.~Salakhutdinov, and C.~D.
  Manning, ``Hotpotqa: A dataset for diverse, explainable multi-hop question
  answering,'' in \emph{Proceedings of the 2018 Conference on Empirical Methods
  in Natural Language Processing}, 2018, pp. 2369--2380.

\bibitem{kipf2016semi}
T.~N. {Kipf} and M.~{Welling}, ``Semi-supervised classification with graph
  convolutional networks,'' in \emph{ICLR (Poster)}, 2016.

\bibitem{ran2019numnet}
Q.~Ran, Y.~Lin, P.~Li, J.~Zhou, and Z.~Liu, ``Numnet: Machine reading
  comprehension with numerical reasoning,'' in \emph{Proceedings of the 2019
  Conference on Empirical Methods in Natural Language Processing and the 9th
  International Joint Conference on Natural Language Processing
  (EMNLP-IJCNLP)}, 2019, pp. 2474--2484.

\bibitem{zhou-etal-2019-gear}
J.~Zhou, X.~Han, C.~Yang, Z.~Liu, L.~Wang, C.~Li, and M.~Sun, ``{GEAR}:
  Graph-based evidence aggregating and reasoning for fact verification,'' in
  \emph{Proceedings of the 57th Annual Meeting of the Association for
  Computational Linguistics}.\hskip 1em plus 0.5em minus 0.4em\relax Florence,
  Italy: Association for Computational Linguistics, Jul. 2019, pp. 892--901.

\bibitem{liu2020fine}
Z.~Liu, C.~Xiong, M.~Sun, and Z.~Liu, ``Fine-grained fact verification with
  kernel graph attention network,'' in \emph{Proceedings of the 58th Annual
  Meeting of the Association for Computational Linguistics}, 2020, pp.
  7342--7351.

\bibitem{kim2019semantic}
S.~Kim, I.~Kang, and N.~Kwak, ``Semantic sentence matching with
  densely-connected recurrent and co-attentive information,'' in
  \emph{Proceedings of the AAAI conference on artificial intelligence},
  vol.~33, no.~01, 2019, pp. 6586--6593.

\bibitem{zhang2020semantics}
Z.~Zhang, Y.~Wu, H.~Zhao, Z.~Li, S.~Zhang, X.~Zhou, and X.~Zhou,
  ``Semantics-aware bert for language understanding,'' in \emph{Proceedings of
  the AAAI Conference on Artificial Intelligence}, vol.~34, no.~05, 2020, pp.
  9628--9635.

\bibitem{zhang2020sg}
Z.~Zhang, Y.~Wu, J.~Zhou, S.~Duan, H.~Zhao, and R.~Wang, ``Sg-net: syntax
  guided transformer for language representation,'' \emph{IEEE Transactions on
  Pattern Analysis and Machine Intelligence}, 2020.

\bibitem{radford2018improving}
A.~Radford, K.~Narasimhan, T.~Salimans, and I.~Sutskever, ``Improving language
  understanding by generative pre-training,'' 2018.

\bibitem{devlin2019bert}
J.~Devlin, M.-W. Chang, K.~Lee, and K.~Toutanova, ``Bert: Pre-training of deep
  bidirectional transformers for language understanding,'' in \emph{Proceedings
  of the 2019 Conference of the North American Chapter of the Association for
  Computational Linguistics: Human Language Technologies, Volume 1 (Long and
  Short Papers)}, 2019, pp. 4171--4186.

\bibitem{liu2019roberta}
Y.~Liu, M.~Ott, N.~Goyal, J.~Du, M.~Joshi, D.~Chen, O.~Levy, M.~Lewis,
  L.~Zettlemoyer, and V.~Stoyanov, ``Roberta: A robustly optimized bert
  pretraining approach,'' \emph{arXiv preprint arXiv:1907.11692}, 2019.

\bibitem{yang2019xlnet}
\BIBentryALTinterwordspacing
Z.~Yang, Z.~Dai, Y.~Yang, J.~G. Carbonell, R.~Salakhutdinov, and Q.~V. Le,
  ``Xlnet: Generalized autoregressive pretraining for language understanding,''
  in \emph{Advances in Neural Information Processing Systems 32: Annual
  Conference on Neural Information Processing Systems 2019, NeurIPS 2019,
  December 8-14, 2019, Vancouver, BC, Canada}, H.~M. Wallach, H.~Larochelle,
  A.~Beygelzimer, F.~d'Alch{\'{e}}{-}Buc, E.~B. Fox, and R.~Garnett, Eds.,
  2019, pp. 5754--5764. [Online]. Available:
  \url{https://proceedings.neurips.cc/paper/2019/hash/dc6a7e655d7e5840e66733e9ee67cc69-Abstract.html}
\BIBentrySTDinterwordspacing

\bibitem{lan2019albert}
\BIBentryALTinterwordspacing
Z.~Lan, M.~Chen, S.~Goodman, K.~Gimpel, P.~Sharma, and R.~Soricut, ``{ALBERT:}
  {A} lite {BERT} for self-supervised learning of language representations,''
  in \emph{8th International Conference on Learning Representations, {ICLR}
  2020, Addis Ababa, Ethiopia, April 26-30, 2020}.\hskip 1em plus 0.5em minus
  0.4em\relax OpenReview.net, 2020. [Online]. Available:
  \url{https://openreview.net/forum?id=H1eA7AEtvS}
\BIBentrySTDinterwordspacing

\bibitem{wang2018glue}
A.~Wang, A.~Singh, J.~Michael, F.~Hill, O.~Levy, and S.~R. Bowman, ``Glue: A
  multi-task benchmark and analysis platform for natural language
  understanding,'' \emph{arXiv preprint arXiv:1804.07461}, 2018.

\bibitem{lai2017race}
G.~Lai, Q.~Xie, H.~Liu, Y.~Yang, and E.~Hovy, ``Race: Large-scale reading
  comprehension dataset from examinations,'' in \emph{Proceedings of the 2017
  Conference on Empirical Methods in Natural Language Processing}, 2017, pp.
  785--794.

\bibitem{rajpurkar2016squad}
P.~Rajpurkar, J.~Zhang, K.~Lopyrev, and P.~Liang, ``Squad: 100,000+ questions
  for machine comprehension of text,'' in \emph{Proceedings of the 2016
  Conference on Empirical Methods in Natural Language Processing}, 2016, pp.
  2383--2392.

\bibitem{reif2019visualizing}
E.~Reif, A.~Yuan, M.~Wattenberg, F.~B. Viegas, A.~Coenen, A.~Pearce, and
  B.~Kim, ``Visualizing and measuring the geometry of bert,'' \emph{Advances in
  Neural Information Processing Systems}, vol.~32, 2019.

\bibitem{ye-etal-2020-coreferential}
\BIBentryALTinterwordspacing
D.~Ye, Y.~Lin, J.~Du, Z.~Liu, P.~Li, M.~Sun, and Z.~Liu, ``{C}oreferential
  {R}easoning {L}earning for {L}anguage {R}epresentation,'' in
  \emph{Proceedings of the 2020 Conference on Empirical Methods in Natural
  Language Processing (EMNLP)}.\hskip 1em plus 0.5em minus 0.4em\relax Online:
  Association for Computational Linguistics, Nov. 2020, pp. 7170--7186.
  [Online]. Available: \url{https://aclanthology.org/2020.emnlp-main.582}
\BIBentrySTDinterwordspacing

\bibitem{clark2020transformers}
P.~Clark, O.~Tafjord, and K.~Richardson, ``Transformers as soft reasoners over
  language,'' \emph{arXiv preprint arXiv:2002.05867}, 2020.

\bibitem{prasad2008penn}
R.~Prasad, N.~Dinesh, A.~Lee, E.~Miltsakaki, L.~Robaldo, A.~K. Joshi, and B.~L.
  Webber, ``The penn discourse treebank 2.0.'' in \emph{LREC}.\hskip 1em plus
  0.5em minus 0.4em\relax Citeseer, 2008.

\bibitem{toulmin2003uses}
S.~E. Toulmin, \emph{The uses of argument}.\hskip 1em plus 0.5em minus
  0.4em\relax Cambridge university press, 2003.

\bibitem{Freeman2011ArgumentSR}
J.~B. Freeman, ``Argument structure: Representation and theory,'' in
  \emph{Argumentation Library}, 2011.

\bibitem{yun2019graph}
S.~Yun, M.~Jeong, R.~Kim, J.~Kang, and H.~J. Kim, ``Graph transformer
  networks,'' \emph{Advances in Neural Information Processing Systems},
  vol.~32, pp. 11\,983--11\,993, 2019.

\bibitem{walicki2016introduction}
M.~Walicki, \emph{Introduction To Mathematical Logic (Extended Edition)}.\hskip
  1em plus 0.5em minus 0.4em\relax World Scientific Publishing Company, 2016.

\bibitem{sep-argument}
C.~Dutilh~Novaes, ``{Argument and Argumentation},'' in \emph{The {Stanford}
  Encyclopedia of Philosophy}, {F}all 2021~ed., E.~N. Zalta, Ed.\hskip 1em plus
  0.5em minus 0.4em\relax Metaphysics Research Lab, Stanford University, 2021.

\bibitem{sep-logic-informal}
L.~Groarke, ``{Informal Logic},'' in \emph{The {Stanford} Encyclopedia of
  Philosophy}, {F}all 2021~ed., E.~N. Zalta, Ed.\hskip 1em plus 0.5em minus
  0.4em\relax Metaphysics Research Lab, Stanford University, 2021.

\bibitem{mann1988rhetorical}
W.~C. Mann and S.~A. Thompson, ``Rhetorical structure theory: Toward a
  functional theory of text organization,'' \emph{Text}, vol.~8, no.~3, pp.
  243--281, 1988.

\bibitem{feng2014two}
V.~W. Feng and G.~Hirst, ``Two-pass discourse segmentation with pairing and
  global features,'' \emph{arXiv preprint arXiv:1407.8215}, 2014.

\bibitem{li2018segbot}
J.~Li, A.~Sun, and S.~R. Joty, ``Segbot: A generic neural text segmentation
  model with pointer network.'' in \emph{IJCAI}, 2018, pp. 4166--4172.

\bibitem{vaswani2017attention}
A.~Vaswani, N.~Shazeer, N.~Parmar, J.~Uszkoreit, L.~Jones, A.~N. Gomez,
  {\L}.~Kaiser, and I.~Polosukhin, ``Attention is all you need,''
  \emph{Advances in neural information processing systems}, vol.~30, 2017.

\bibitem{ba2016layer}
J.~L. Ba, J.~R. Kiros, and G.~E. Hinton, ``Layer normalization,'' \emph{stat},
  vol. 1050, p.~21, 2016.

\bibitem{cho2014properties}
K.~Cho, B.~Van~Merri{\"e}nboer, D.~Bahdanau, and Y.~Bengio, ``On the properties
  of neural machine translation: Encoder-decoder approaches,'' \emph{arXiv
  preprint arXiv:1409.1259}, 2014.

\bibitem{he2016deep}
K.~He, X.~Zhang, S.~Ren, and J.~Sun, ``Deep residual learning for image
  recognition,'' in \emph{Proceedings of the IEEE conference on computer vision
  and pattern recognition}, 2016, pp. 770--778.

\bibitem{joulin2017bag}
A.~Joulin, {\'E}.~Grave, P.~Bojanowski, and T.~Mikolov, ``Bag of tricks for
  efficient text classification,'' in \emph{Proceedings of the 15th Conference
  of the European Chapter of the Association for Computational Linguistics:
  Volume 2, Short Papers}, 2017, pp. 427--431.

\bibitem{radford2019language}
A.~Radford, J.~Wu, R.~Child, D.~Luan, D.~Amodei, I.~Sutskever \emph{et~al.},
  ``Language models are unsupervised multitask learners,'' \emph{OpenAI blog},
  vol.~1, no.~8, p.~9, 2019.

\bibitem{cui2019pre}
Y.~Cui, W.~Che, T.~Liu, B.~Qin, Z.~Yang, S.~Wang, and G.~Hu, ``Pre-training
  with whole word masking for chinese bert,'' \emph{arXiv preprint
  arXiv:1906.08101}, 2019.

\bibitem{ouyang2021fact}
S.~Ouyang, Z.~Zhang, and H.~Zhao, ``Fact-driven logical reasoning,''
  \emph{arXiv preprint arXiv:2105.10334}, 2021.

\bibitem{wang2021logic}
S.~Wang, W.~Zhong, D.~Tang, Z.~Wei, Z.~Fan, D.~Jiang, M.~Zhou, and N.~Duan,
  ``Logic-driven context extension and data augmentation for logical reasoning
  of text,'' \emph{arXiv preprint arXiv:2105.03659}, 2021.

\bibitem{jiao-etal-2022-merit}
\BIBentryALTinterwordspacing
F.~Jiao, Y.~Guo, X.~Song, and L.~Nie, ``{MERI}t: {M}eta-{P}ath {G}uided
  {C}ontrastive {L}earning for {L}ogical {R}easoning,'' in \emph{Findings of
  the Association for Computational Linguistics: ACL 2022}.\hskip 1em plus
  0.5em minus 0.4em\relax Dublin, Ireland: Association for Computational
  Linguistics, May 2022, pp. 3496--3509. [Online]. Available:
  \url{https://aclanthology.org/2022.findings-acl.276}
\BIBentrySTDinterwordspacing

\bibitem{yih2013question}
S.~W.-t. Yih, M.-W. Chang, C.~Meek, and A.~Pastusiak, ``Question answering
  asing enhanced lexical semantic models,'' 2013.

\bibitem{richardson-etal-2013-mctest}
\BIBentryALTinterwordspacing
M.~Richardson, C.~J. Burges, and E.~Renshaw, ``{MCT}est: A challenge dataset
  for the open-domain machine comprehension of text,'' in \emph{Proceedings of
  the 2013 Conference on Empirical Methods in Natural Language
  Processing}.\hskip 1em plus 0.5em minus 0.4em\relax Seattle, Washington, USA:
  Association for Computational Linguistics, Oct. 2013, pp. 193--203. [Online].
  Available: \url{https://aclanthology.org/D13-1020}
\BIBentrySTDinterwordspacing

\bibitem{lowe-etal-2015-ubuntu}
\BIBentryALTinterwordspacing
R.~Lowe, N.~Pow, I.~Serban, and J.~Pineau, ``The {U}buntu dialogue corpus: A
  large dataset for research in unstructured multi-turn dialogue systems,'' in
  \emph{Proceedings of the 16th Annual Meeting of the Special Interest Group on
  Discourse and Dialogue}.\hskip 1em plus 0.5em minus 0.4em\relax Prague, Czech
  Republic: Association for Computational Linguistics, Sep. 2015, pp. 285--294.
  [Online]. Available: \url{https://aclanthology.org/W15-4640}
\BIBentrySTDinterwordspacing

\bibitem{Liu_Zhang_Zhao_Zhou_Zhou_2021}
\BIBentryALTinterwordspacing
L.~Liu, Z.~Zhang, H.~Zhao, X.~Zhou, and X.~Zhou, ``Filling the gap of
  utterance-aware and speaker-aware representation for multi-turn dialogue,''
  \emph{Proceedings of the AAAI Conference on Artificial Intelligence},
  vol.~35, no.~15, pp. 13\,406--13\,414, May 2021. [Online]. Available:
  \url{https://ojs.aaai.org/index.php/AAAI/article/view/17582}
\BIBentrySTDinterwordspacing

\bibitem{DBLP:conf/iclr/LoshchilovH19}
\BIBentryALTinterwordspacing
I.~Loshchilov and F.~Hutter, ``Decoupled weight decay regularization,'' in
  \emph{7th International Conference on Learning Representations, {ICLR} 2019,
  New Orleans, LA, USA, May 6-9, 2019}.\hskip 1em plus 0.5em minus 0.4em\relax
  OpenReview.net, 2019. [Online]. Available:
  \url{https://openreview.net/forum?id=Bkg6RiCqY7}
\BIBentrySTDinterwordspacing

\bibitem{liu2021non}
M.~Liu, Z.~Wang, and S.~Ji, ``Non-local graph neural networks,'' \emph{IEEE
  transactions on pattern analysis and machine intelligence}, vol.~44, no.~12,
  pp. 10\,270--10\,276, 2021.

\bibitem{yu2020representative}
C.~Yu, Y.~Liu, C.~Gao, C.~Shen, and N.~Sang, ``Representative graph neural
  network,'' in \emph{Computer Vision--ECCV 2020: 16th European Conference,
  Glasgow, UK, August 23--28, 2020, Proceedings, Part VII 16}.\hskip 1em plus
  0.5em minus 0.4em\relax Springer, 2020, pp. 379--396.

\bibitem{li2018deeper}
Q.~Li, Z.~Han, and X.-M. Wu, ``Deeper insights into graph convolutional
  networks for semi-supervised learning,'' in \emph{Thirty-Second AAAI
  conference on artificial intelligence}, 2018.

\bibitem{huang2020tackling}
W.~Huang, Y.~Rong, T.~Xu, F.~Sun, and J.~Huang, ``Tackling over-smoothing for
  general graph convolutional networks,'' \emph{arXiv e-prints}, pp.
  arXiv--2008, 2020.

\bibitem{berant2013semantic}
J.~Berant, A.~Chou, R.~Frostig, and P.~Liang, ``Semantic parsing on freebase
  from question-answer pairs,'' in \emph{Proceedings of the 2013 conference on
  empirical methods in natural language processing}, 2013, pp. 1533--1544.

\bibitem{talmor-etal-2019-commonsenseqa}
\BIBentryALTinterwordspacing
A.~Talmor, J.~Herzig, N.~Lourie, and J.~Berant, ``{C}ommonsense{QA}: A question
  answering challenge targeting commonsense knowledge,'' in \emph{Proceedings
  of the 2019 Conference of the North {A}merican Chapter of the Association for
  Computational Linguistics: Human Language Technologies, Volume 1 (Long and
  Short Papers)}.\hskip 1em plus 0.5em minus 0.4em\relax Minneapolis,
  Minnesota: Association for Computational Linguistics, Jun. 2019, pp.
  4149--4158. [Online]. Available: \url{https://aclanthology.org/N19-1421}
\BIBentrySTDinterwordspacing

\bibitem{dua2019drop}
D.~Dua, Y.~Wang, P.~Dasigi, G.~Stanovsky, S.~Singh, and M.~Gardner, ``Drop: A
  reading comprehension benchmark requiring discrete reasoning over
  paragraphs,'' in \emph{Proceedings of the 2019 Conference of the North
  American Chapter of the Association for Computational Linguistics: Human
  Language Technologies, Volume 1 (Long and Short Papers)}, 2019, pp.
  2368--2378.

\bibitem{thorne2018fever}
J.~Thorne, A.~Vlachos, C.~Christodoulopoulos, and A.~Mittal, ``Fever: a
  large-scale dataset for fact extraction and verification,'' in
  \emph{Proceedings of the 2018 Conference of the North American Chapter of the
  Association for Computational Linguistics: Human Language Technologies,
  Volume 1 (Long Papers)}, 2018, pp. 809--819.

\bibitem{bowman2015large}
S.~Bowman, G.~Angeli, C.~Potts, and C.~D. Manning, ``A large annotated corpus
  for learning natural language inference,'' in \emph{Proceedings of the 2015
  Conference on Empirical Methods in Natural Language Processing}, 2015, pp.
  632--642.

\bibitem{williams2017broad}
A.~Williams, N.~Nangia, and S.~R. Bowman, ``A broad-coverage challenge corpus
  for sentence understanding through inference,'' in \emph{NAACL-HLT}, 2018.

\bibitem{bao-etal-2016-constraint}
\BIBentryALTinterwordspacing
J.~Bao, N.~Duan, Z.~Yan, M.~Zhou, and T.~Zhao, ``Constraint-based question
  answering with knowledge graph,'' in \emph{Proceedings of {COLING} 2016, the
  26th International Conference on Computational Linguistics: Technical
  Papers}.\hskip 1em plus 0.5em minus 0.4em\relax Osaka, Japan: The COLING 2016
  Organizing Committee, Dec. 2016, pp. 2503--2514. [Online]. Available:
  \url{https://aclanthology.org/C16-1236}
\BIBentrySTDinterwordspacing

\bibitem{yih-etal-2016-value}
\BIBentryALTinterwordspacing
W.-t. Yih, M.~Richardson, C.~Meek, M.-W. Chang, and J.~Suh, ``The value of
  semantic parse labeling for knowledge base question answering,'' in
  \emph{Proceedings of the 54th Annual Meeting of the Association for
  Computational Linguistics (Volume 2: Short Papers)}.\hskip 1em plus 0.5em
  minus 0.4em\relax Berlin, Germany: Association for Computational Linguistics,
  Aug. 2016, pp. 201--206. [Online]. Available:
  \url{https://aclanthology.org/P16-2033}
\BIBentrySTDinterwordspacing

\bibitem{talmor-berant-2018-web}
\BIBentryALTinterwordspacing
A.~Talmor and J.~Berant, ``The web as a knowledge-base for answering complex
  questions,'' in \emph{Proceedings of the 2018 Conference of the North
  {A}merican Chapter of the Association for Computational Linguistics: Human
  Language Technologies, Volume 1 (Long Papers)}.\hskip 1em plus 0.5em minus
  0.4em\relax New Orleans, Louisiana: Association for Computational
  Linguistics, Jun. 2018, pp. 641--651. [Online]. Available:
  \url{https://aclanthology.org/N18-1059}
\BIBentrySTDinterwordspacing

\bibitem{lopez2013evaluating}
V.~Lopez, C.~Unger, P.~Cimiano, and E.~Motta, ``Evaluating question answering
  over linked data,'' \emph{Journal of Web Semantics}, vol.~21, pp. 3--13,
  2013.

\bibitem{trivedi2017lc}
P.~Trivedi, G.~Maheshwari, M.~Dubey, and J.~Lehmann, ``Lc-quad: A corpus for
  complex question answering over knowledge graphs,'' in \emph{International
  Semantic Web Conference}.\hskip 1em plus 0.5em minus 0.4em\relax Springer,
  2017, pp. 210--218.

\bibitem{dubey2019lc}
M.~Dubey, D.~Banerjee, A.~Abdelkawi, and J.~Lehmann, ``Lc-quad 2.0: A large
  dataset for complex question answering over wikidata and dbpedia,'' in
  \emph{International semantic web conference}.\hskip 1em plus 0.5em minus
  0.4em\relax Springer, 2019, pp. 69--78.

\bibitem{freebase:datadumps}
Google, ``Freebase data dumps,''
  \url{https://developers.google.com/freebase/data}, <year>.

\bibitem{lehmann2015dbpedia}
J.~Lehmann, R.~Isele, M.~Jakob, A.~Jentzsch, D.~Kontokostas, P.~N. Mendes,
  S.~Hellmann, M.~Morsey, P.~Van~Kleef, S.~Auer \emph{et~al.}, ``Dbpedia--a
  large-scale, multilingual knowledge base extracted from wikipedia,''
  \emph{Semantic web}, vol.~6, no.~2, pp. 167--195, 2015.

\bibitem{huang2019cosmos}
L.~Huang, R.~Le~Bras, C.~Bhagavatula, and Y.~Choi, ``Cosmos qa: Machine reading
  comprehension with contextual commonsense reasoning,'' in \emph{Proceedings
  of the 2019 Conference on Empirical Methods in Natural Language Processing
  and the 9th International Joint Conference on Natural Language Processing
  (EMNLP-IJCNLP)}, 2019, pp. 2391--2401.

\bibitem{tandon2019wiqa}
N.~Tandon, B.~Dalvi, K.~Sakaguchi, P.~Clark, and A.~Bosselut, ``Wiqa: A dataset
  for “what if...” reasoning over procedural text,'' in \emph{Proceedings
  of the 2019 Conference on Empirical Methods in Natural Language Processing
  and the 9th International Joint Conference on Natural Language Processing
  (EMNLP-IJCNLP)}, 2019, pp. 6078--6087.

\bibitem{lin-etal-2019-kagnet}
\BIBentryALTinterwordspacing
B.~Y. Lin, X.~Chen, J.~Chen, and X.~Ren, ``{K}ag{N}et: Knowledge-aware graph
  networks for commonsense reasoning,'' in \emph{Proceedings of the 2019
  Conference on Empirical Methods in Natural Language Processing and the 9th
  International Joint Conference on Natural Language Processing
  (EMNLP-IJCNLP)}.\hskip 1em plus 0.5em minus 0.4em\relax Hong Kong, China:
  Association for Computational Linguistics, Nov. 2019, pp. 2829--2839.
  [Online]. Available: \url{https://aclanthology.org/D19-1282}
\BIBentrySTDinterwordspacing

\bibitem{feng-etal-2020-scalable}
\BIBentryALTinterwordspacing
Y.~Feng, X.~Chen, B.~Y. Lin, P.~Wang, J.~Yan, and X.~Ren, ``Scalable multi-hop
  relational reasoning for knowledge-aware question answering,'' in
  \emph{Proceedings of the 2020 Conference on Empirical Methods in Natural
  Language Processing (EMNLP)}.\hskip 1em plus 0.5em minus 0.4em\relax Online:
  Association for Computational Linguistics, Nov. 2020, pp. 1295--1309.
  [Online]. Available: \url{https://aclanthology.org/2020.emnlp-main.99}
\BIBentrySTDinterwordspacing

\bibitem{speer2017conceptnet}
R.~Speer, J.~Chin, and C.~Havasi, ``Conceptnet 5.5: An open multilingual graph
  of general knowledge,'' in \emph{Thirty-first AAAI conference on artificial
  intelligence}, 2017.

\bibitem{DBLP:conf/acl/WangTDWHJCJZ21}
\BIBentryALTinterwordspacing
R.~Wang, D.~Tang, N.~Duan, Z.~Wei, X.~Huang, J.~Ji, G.~Cao, D.~Jiang, and
  M.~Zhou, ``K-adapter: Infusing knowledge into pre-trained models with
  adapters,'' in \emph{Findings of the Association for Computational
  Linguistics: {ACL/IJCNLP} 2021, Online Event, August 1-6, 2021}, ser.
  Findings of {ACL}, C.~Zong, F.~Xia, W.~Li, and R.~Navigli, Eds., vol.
  {ACL/IJCNLP} 2021.\hskip 1em plus 0.5em minus 0.4em\relax Association for
  Computational Linguistics, 2021, pp. 1405--1418. [Online]. Available:
  \url{https://doi.org/10.18653/v1/2021.findings-acl.121}
\BIBentrySTDinterwordspacing

\bibitem{cohan-etal-2018-discourse}
A.~Cohan, F.~Dernoncourt, D.~S. Kim, T.~Bui, S.~Kim, W.~Chang, and N.~Goharian,
  ``A discourse-aware attention model for abstractive summarization of long
  documents,'' in \emph{Proceedings of the 2018 Conference of the North
  {A}merican Chapter of the Association for Computational Linguistics: Human
  Language Technologies, Volume 2 (Short Papers)}.\hskip 1em plus 0.5em minus
  0.4em\relax New Orleans, Louisiana: Association for Computational
  Linguistics, Jun. 2018, pp. 615--621.

\bibitem{joty-etal-2019-discourse}
S.~Joty, G.~Carenini, R.~Ng, and G.~Murray, ``Discourse analysis and its
  applications,'' in \emph{Proceedings of the 57th Annual Meeting of the
  Association for Computational Linguistics: Tutorial Abstracts}.\hskip 1em
  plus 0.5em minus 0.4em\relax Florence, Italy: Association for Computational
  Linguistics, Jul. 2019, pp. 12--17.

\bibitem{xu-etal-2020-discourse}
J.~Xu, Z.~Gan, Y.~Cheng, and J.~Liu, ``Discourse-aware neural extractive text
  summarization,'' in \emph{Proceedings of the 58th Annual Meeting of the
  Association for Computational Linguistics}.\hskip 1em plus 0.5em minus
  0.4em\relax Online: Association for Computational Linguistics, Jul. 2020, pp.
  5021--5031.

\bibitem{feng2020dialogue}
X.~Feng, X.~Feng, B.~Qin, X.~Geng, and T.~Liu, ``Dialogue discourse-aware graph
  convolutional networks for abstractive meeting summarization,'' \emph{arXiv
  preprint arXiv:2012.03502}, 2020.

\bibitem{voita2018context}
E.~Voita, P.~Serdyukov, R.~Sennrich, and I.~Titov, ``Context-aware neural
  machine translation learns anaphora resolution,'' in \emph{Proceedings of the
  56th Annual Meeting of the Association for Computational Linguistics (Volume
  1: Long Papers)}, 2018, pp. 1264--1274.

\bibitem{bosselut2018discourse}
A.~Bosselut, A.~Celikyilmaz, X.~He, J.~Gao, P.-S. Huang, and Y.~Choi,
  ``Discourse-aware neural rewards for coherent text generation,'' in
  \emph{Proceedings of the 2018 Conference of the North American Chapter of the
  Association for Computational Linguistics: Human Language Technologies,
  Volume 1 (Long Papers)}, 2018, pp. 173--184.

\bibitem{gao2020discern}
Y.~Gao, C.-S. Wu, J.~Li, S.~Joty, S.~C. Hoi, C.~Xiong, I.~King, and M.~Lyu,
  ``Discern: Discourse-aware entailment reasoning network for conversational
  machine reading,'' in \emph{Proceedings of the 2020 Conference on Empirical
  Methods in Natural Language Processing (EMNLP)}, 2020, pp. 2439--2449.

\bibitem{mihaylov-frank-2019-discourse}
T.~Mihaylov and A.~Frank, ``Discourse-aware semantic self-attention for
  narrative reading comprehension,'' in \emph{Proceedings of the 2019
  Conference on Empirical Methods in Natural Language Processing and the 9th
  International Joint Conference on Natural Language Processing
  (EMNLP-IJCNLP)}.\hskip 1em plus 0.5em minus 0.4em\relax Hong Kong, China:
  Association for Computational Linguistics, Nov. 2019, pp. 2541--2552.

\bibitem{webber2004d}
B.~Webber, ``D-ltag: extending lexicalized tag to discourse,'' \emph{Cognitive
  Science}, vol.~28, no.~5, pp. 751--779, 2004.

\bibitem{carlson2003building}
L.~Carlson, D.~Marcu, and M.~E. Okurowski, ``Building a discourse-tagged corpus
  in the framework of rhetorical structure theory,'' in \emph{Current and new
  directions in discourse and dialogue}.\hskip 1em plus 0.5em minus 0.4em\relax
  Springer, 2003, pp. 85--112.

\bibitem{hayashi2016empirical}
K.~Hayashi, T.~Hirao, and M.~Nagata, ``Empirical comparison of dependency
  conversions for rst discourse trees,'' in \emph{Proceedings of the 17th
  annual meeting of the special interest group on discourse and dialogue},
  2016, pp. 128--136.

\bibitem{ji2014representation}
Y.~Ji and J.~Eisenstein, ``Representation learning for text-level discourse
  parsing,'' in \emph{Proceedings of the 52nd annual meeting of the association
  for computational linguistics (volume 1: Long papers)}, 2014, pp. 13--24.

\bibitem{kishimoto2020adapting}
Y.~Kishimoto, Y.~Murawaki, and S.~Kurohashi, ``Adapting bert to implicit
  discourse relation classification with a focus on discourse connectives,'' in
  \emph{Proceedings of The 12th Language Resources and Evaluation Conference},
  2020, pp. 1152--1158.

\bibitem{lei2018linguistic}
W.~Lei, Y.~Xiang, Y.~Wang, Q.~Zhong, M.~Liu, and M.-Y. Kan, ``Linguistic
  properties matter for implicit discourse relation recognition: Combining
  semantic interaction, topic continuity and attribution,'' in
  \emph{Proceedings of the AAAI Conference on Artificial Intelligence},
  vol.~32, no.~1, 2018.

\bibitem{ji2017neural}
Y.~{Ji} and N.~A. {Smith}, ``Neural discourse structure for text
  categorization.'' in \emph{Proceedings of the 55th Annual Meeting of the
  Association for Computational Linguistics (Volume 1: Long Papers)}, vol.~1,
  2017, pp. 996--1005.

\bibitem{liu2018learning}
Y.~{Liu} and M.~{Lapata}, ``Learning structured text representations,''
  \emph{Transactions of the Association for Computational Linguistics}, vol.~6,
  pp. 63--75, 2018.

\bibitem{liu2019single}
Y.~{Liu}, I.~{Titov}, and M.~{Lapata}, ``Single document summarization as tree
  induction.'' in \emph{Proceedings of the 2019 Conference of the North
  American Chapter of the Association for Computational Linguistics: Human
  Language Technologies, Volume 1 (Long and Short Papers)}, 2019, pp.
  1745--1755.

\end{thebibliography}
%
% % <OR> manually copy in the resultant .bbl file
% % set second argument of \begin to the number of references
% % (used to reserve space for the reference number labels box)
% % \begin{thebibliography}{1}

% % \bibitem{IEEEhowto:kopka}
% % H.~Kopka and P.~W. Daly, \emph{A Guide to \LaTeX}, 3rd~ed.\hskip 1em plus
% %   0.5em minus 0.4em\relax Harlow, England: Addison-Wesley, 1999.

% % \end{thebibliography}

\newpage
% % biography section
% % 
% % If you have an EPS/PDF photo (graphicx package needed) extra braces are
% % needed around the contents of the optional argument to biography to prevent
% % the LaTeX parser from getting confused when it sees the complicated
% % \includegraphics command within an optional argument. (You could create
% % your own custom macro containing the \includegraphics command to make things
% % simpler here.)
% %\begin{IEEEbiography}[{\includegraphics[width=1in,height=1.25in,clip,keepaspectratio]{mshell}}]{Michael Shell}
% % or if you just want to reserve a space for a photo:

\begin{IEEEbiography}[{\includegraphics[width=1in,height=1.25in,clip,keepaspectratio]{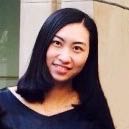}}]{Yinya Huang}
is currently a Ph.D. student in Computer Science at the School of Intelligent Systems Engineering at Sun Yat-Sen University, China. Her research interests include machine reasoning and natural language understanding.
\end{IEEEbiography}

\begin{IEEEbiography}[{\includegraphics[width=1in,height=1.25in,clip,keepaspectratio]{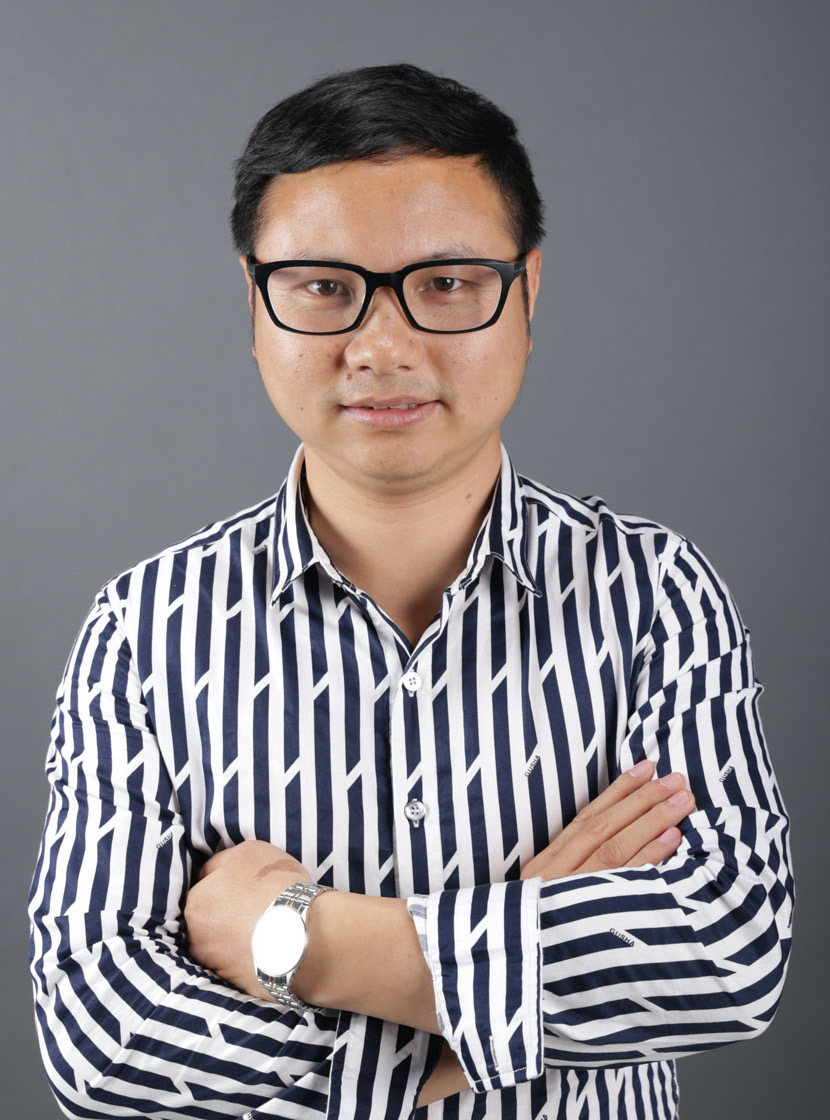}}]{Lemao Liu}
is a senior researcher at Natural Language Processing Center, Tencent AI Lab, China. Previously, he was with the National Institute of Information and Communications Technology (NICT), Japan. He received his Ph.D. degree from the Harbin Institute of Technology. His research interests include machine translation, syntactic parsing, and natural language understanding. He has published about 50 research papers in leading conferences and journals, such as ACL, EMNLP, NAACL, COLING, ICLR, AAAI, and JAIR. He received an outstanding paper award in ACL 2021 and the best demo award in CCL 2020. He served as a publication co-chair in EMNLP 2020 (Findings), and a senior program committee member in IJCAI 2021.
\end{IEEEbiography}

\begin{IEEEbiography}[{\includegraphics[width=1in,height=1.25in,clip,keepaspectratio]{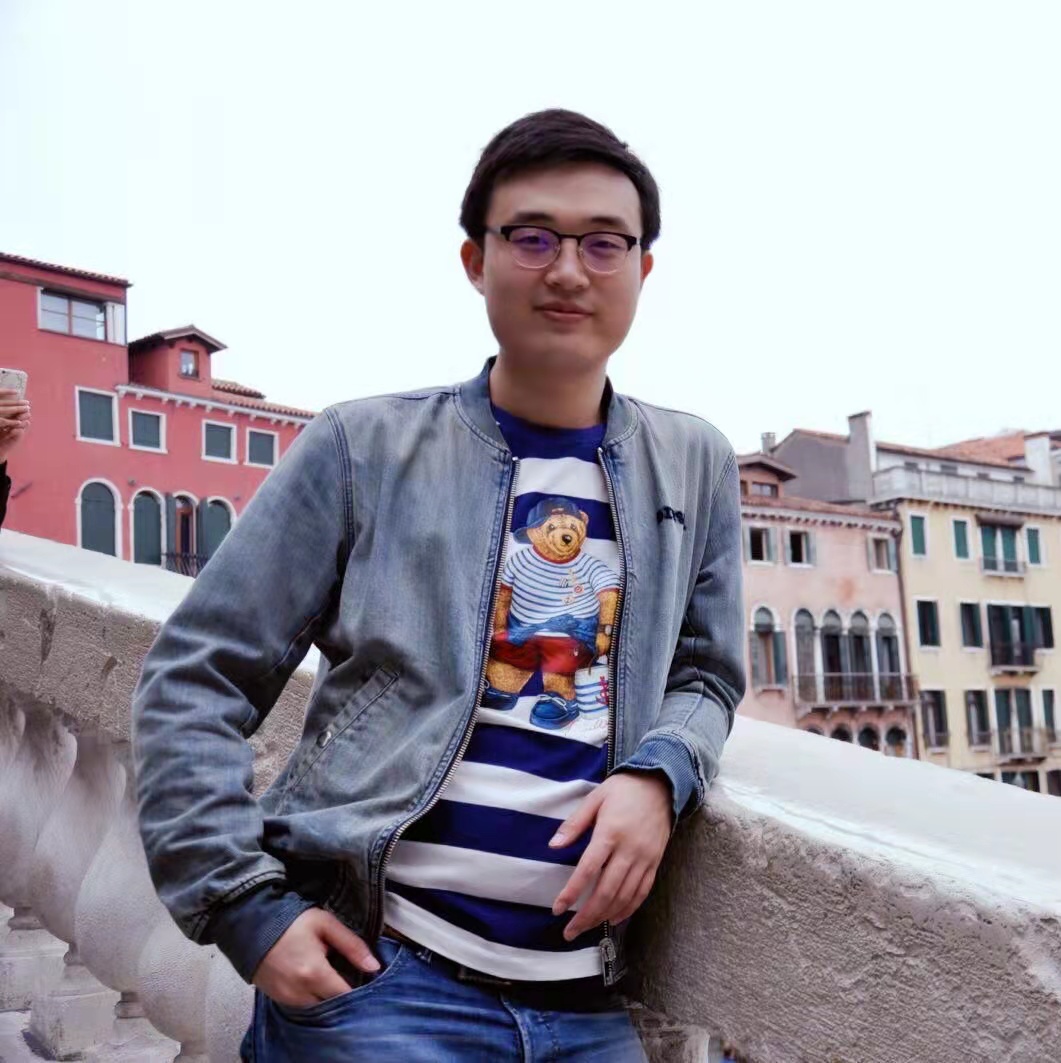}}]{Kun Xu}
is currently a principal research scientist at Huawei. He was a senior research scientist in Tencent AI Lab from 2018 to 2021. He received his Ph.D. degree from Peking University in 2016. He has published more than 40 papers in top conferences of NLP such as ACL, EMNLP, and NAACL. He was the recipient of the best paper award in NAACL in 2021. His research interests include question-answering and semantic parsing.
\end{IEEEbiography}

\begin{IEEEbiography}[{\includegraphics[width=1in,height=1.25in,clip,keepaspectratio]{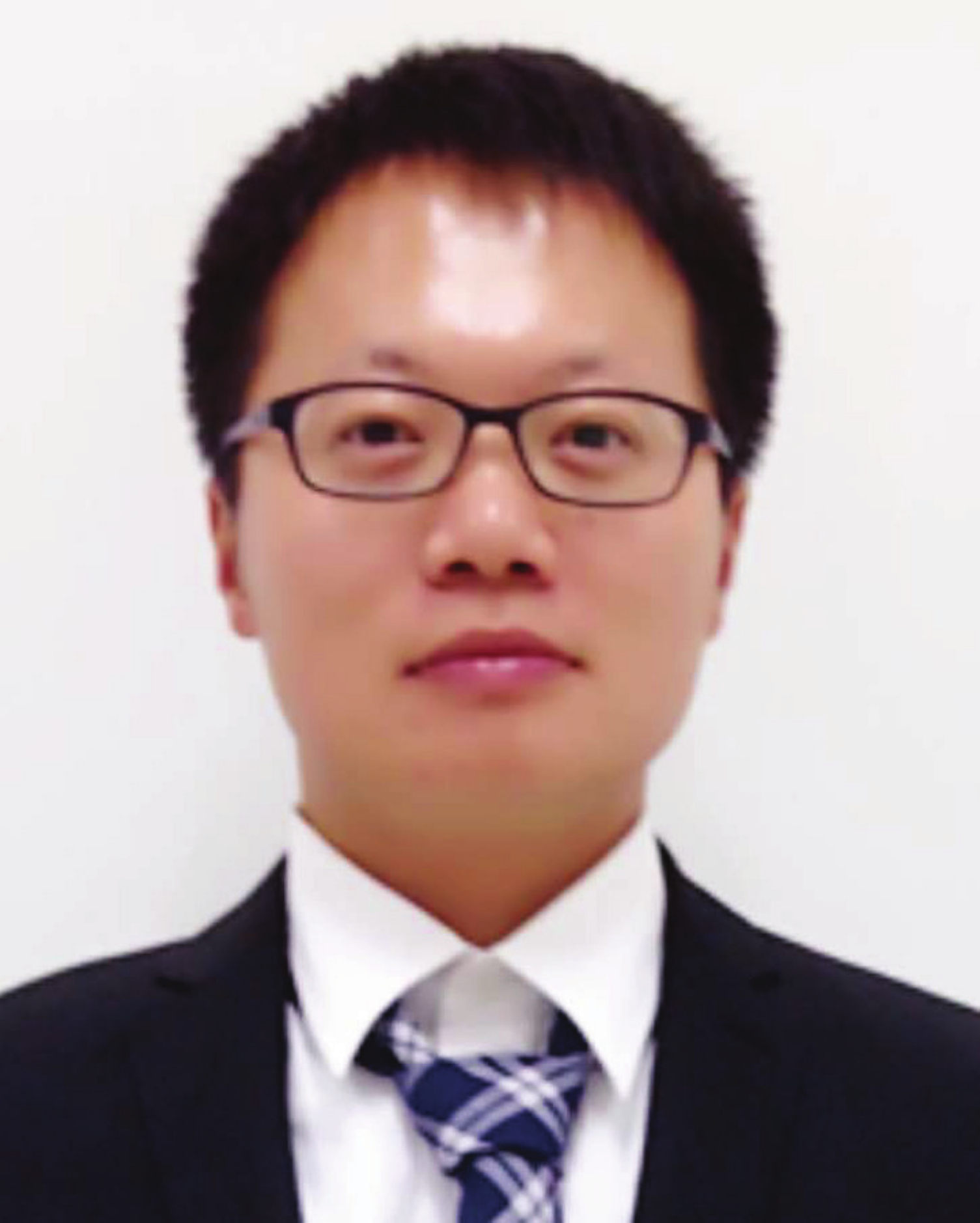}}]{Meng Fang}
is currently an Assistant Professor at the University of Liverpool. He received his Ph.D. degree from the University of Technology Sydney, Australia, in 2015. He was a postdoctoral research fellow in the School of Computing and Information Systems, the University of Melbourne. His current research interests include reinforcement learning, natural language processing, and machine learning.
\end{IEEEbiography}

\begin{IEEEbiography}[{\includegraphics[width=1in,height=1.25in,clip,keepaspectratio]{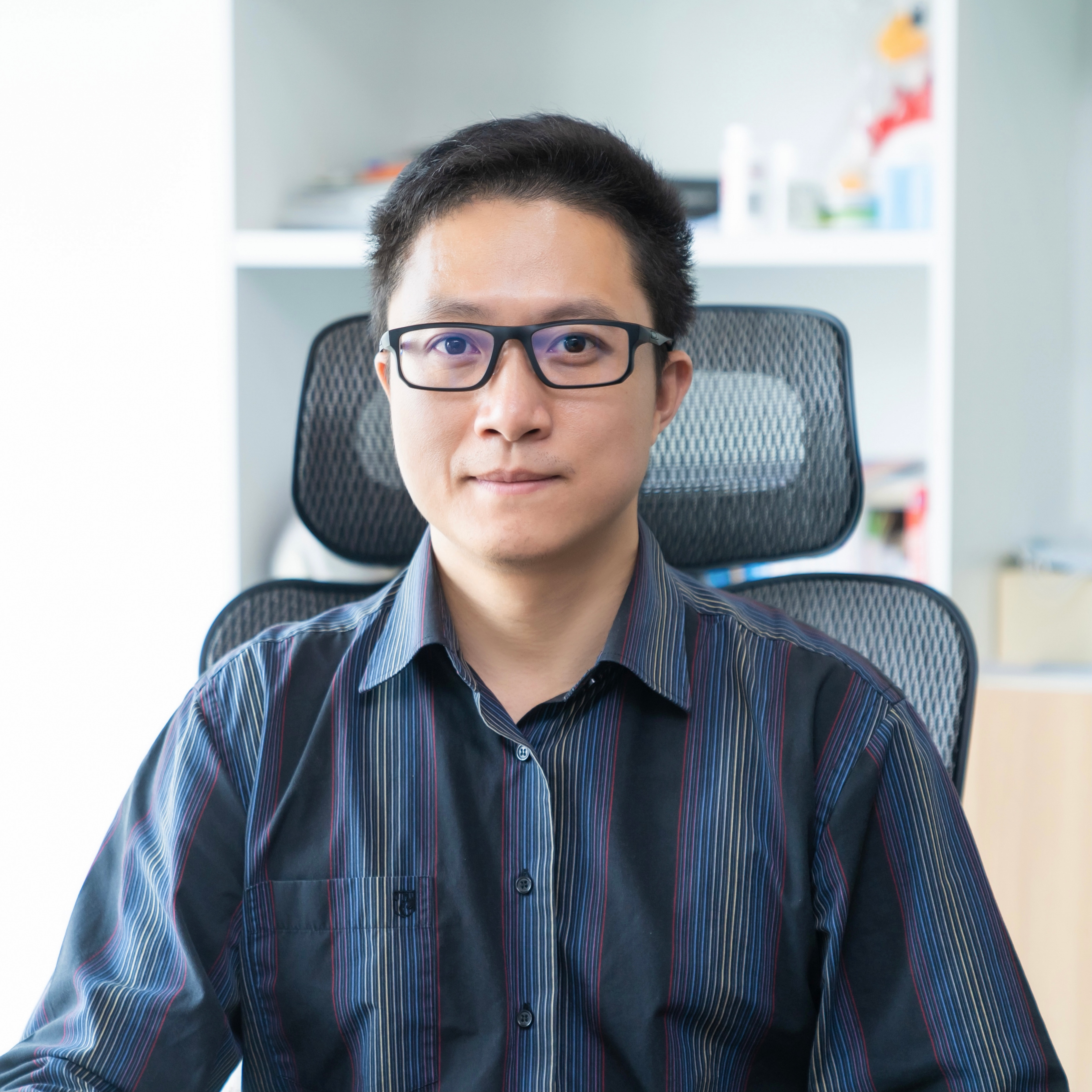}}]{Liang Lin}
 is a full professor of Computer Science at Sun Yat-sen University and CEO of DarkerMatter AI. He worked as the Executive Director of the SenseTime Group from 2016 to 2018, leading the R\&D teams in developing cutting-edge, deliverable solutions in computer vision, data analysis and mining, and intelligent robotic systems.  He has authored or co-authored more than 200 papers in leading academic journals and conferences. He is an associate editor of IEEE Trans.  Human-Machine Systems and IET Computer Vision, and he served as the area/session chair for numerous conferences such as CVPR, ICME, ICCV. He was the recipient of Annual Best Paper Award by Pattern Recognition (Elsevier) in 2018, Dimond Award for best paper in IEEE ICME in 2017, ACM NPAR Best Paper Runners-Up Award in 2010, Google Faculty Award in 2012, award for the best student paper in IEEE ICME in 2014, and Hong Kong Scholars Award in 2014. He is a Fellow of IET.
\end{IEEEbiography}

\begin{IEEEbiography}
[{\includegraphics[width=1in,height=1.25in,clip,keepaspectratio]{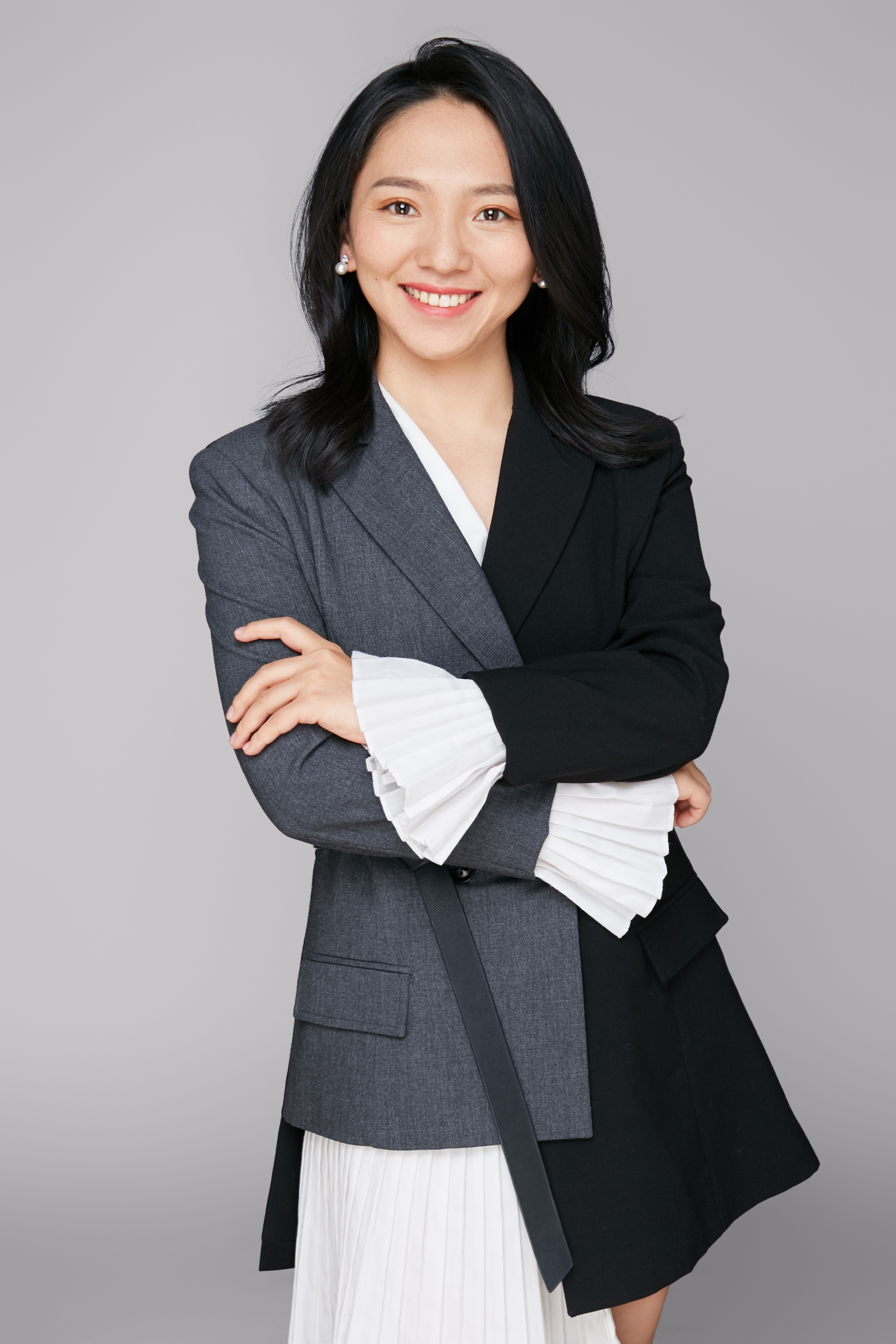}}]
{Xiaodan Liang}
 is currently an Associate Professor at Sun Yat-sen University. She was a postdoc researcher in the machine learning department at Carnegie Mellon University, working with Prof. Eric Xing, from 2016 to 2018. She received her Ph.D. degree from Sun Yat-sen University in 2016. She has published several cutting-edge projects on human-related analysis,
including human parsing, pedestrian detection, and instance segmentation, 2D/3D human pose estimation, and activity recognition.
\end{IEEEbiography}

% % \begin{IEEEbiographynophoto}{John Doe}
% % Biography text here.
% % \end{IEEEbiographynophoto}

% % insert where needed to balance the two columns on the last page with
% %\newpage

% % \begin{IEEEbiographynophoto}{Jane Doe}
% % Biography text here.
% % \end{IEEEbiographynophoto}

% % You can push biographies down or up by placing
% % a \vfill before or after them. The appropriate
% % use of \vfill depends on what kind of text is
% % on the last page and whether or not the columns
% % are being equalized.

% %\vfill

% % Can be used to pull up biographies so that the bottom of the last one
% % is flush with the other column.
% %\enlargethispage{-5in}

% % that's all folks
\end{document}